\documentclass[11pt]{article}

\usepackage[final]{latex/acl}

\usepackage{times}
\usepackage{latexsym}
\usepackage{amsmath}

\usepackage{amsmath,amsfonts,bm}

\def\eqref#1{equation~\ref{#1}}

\def\1{\bm{1}}

\DeclareMathAlphabet{\mathsfit}{\encodingdefault}{\sfdefault}{m}{sl}
\SetMathAlphabet{\mathsfit}{bold}{\encodingdefault}{\sfdefault}{bx}{n}

\usepackage{hyperref}
\usepackage{url}
\usepackage{enumitem}
\usepackage{booktabs}       
\usepackage{amsfonts}       
\usepackage{nicefrac}       
\usepackage{microtype}      
\usepackage{xcolor}         
\usepackage{graphicx}
\usepackage{multirow}
\usepackage{float}
\usepackage[table]{xcolor}
\usepackage{adjustbox}
\usepackage{tabularx} 
\usepackage{graphicx}   
\usepackage{booktabs}   
\usepackage[most]{tcolorbox}
\usepackage{siunitx} 
\usepackage{pifont}
\usepackage[utf8]{inputenc}
\usepackage{csquotes}
\usepackage[T1]{fontenc}
\usepackage[linesnumbered,ruled,vlined]{algorithm2e}
\usepackage[T1]{fontenc}

\usepackage{microtype}

\usepackage{inconsolata}

\usepackage{graphicx}
\usepackage{booktabs}
\usepackage{multirow}
\usepackage{xcolor} 
\usepackage[most]{tcolorbox}

\usepackage[most]{tcolorbox}
\tcbset{
  hugbw/.style={
    enhanced, breakable,
    colback=white, colframe=black!30, boxrule=0.4pt,
    colbacktitle=black!5, coltitle=black,
    fonttitle=\bfseries,
    attach boxed title to top left={yshift=-2pt, xshift=4pt},
    boxed title style={sharp corners, boxrule=0pt, 
      colframe=black!30, colback=black!5,
      left=6pt, right=6pt, top=2pt, bottom=2pt},
    sharp corners,
    left=7pt, right=7pt, top=6pt, bottom=6pt,
  }
}

\definecolor{huggreen}{RGB}{60,179,113} 

\usepackage{tikz}
\newcommand{\solidcircle}[1]{%
  \tikz[baseline=-0.8ex]{    
    \node[
      circle, fill=huggreen,
      inner sep=1pt,
      minimum size=0.6em,
      text=white,
      font=\sffamily\bfseries\footnotesize,
      text height=1.3ex, text depth=0pt
    ] {\scriptsize #1};
  }%
}

\title{\scalebox{0.98}[1]{HugAgent: A Human Simulation Benchmark for Individual-Level Reasoning}\vspace{0.6em}}

\author{%
\textbf{Chance Jiajie Li\textsuperscript{1}\thanks{\;Equal contribution. $^\dagger$Now at Google. Correspondence: \texttt{jiajie@mit.edu}.},
Zhenze Mo\textsuperscript{8}\footnotemark[1],
Yuhan Tang\textsuperscript{5}\footnotemark[1],
Ao Qu\textsuperscript{4},
Jiayi Wu\textsuperscript{9},} \\ [0.4ex]
\textbf{Kaiya Ivy Zhao\textsuperscript{2,3}, 
Yulu Gan\textsuperscript{2},
Jie Fan\textsuperscript{2,7,$\dagger$},
Jiangbo Yu\textsuperscript{10},
Hang Jiang\textsuperscript{1,8},} \\[0.4ex]
\textbf{ Paul Pu Liang\textsuperscript{1,2},
Jinhua Zhao\textsuperscript{4,5,6}, 
Luis Alonso\textsuperscript{1},
Kent Larson\textsuperscript{1}} \\[1.5ex]
\textsuperscript{1}MIT Media Lab \quad
\textsuperscript{2}MIT EECS \quad
\textsuperscript{3}MIT BCS \quad
\textsuperscript{4}MIT IDSS \quad
\textsuperscript{5}MIT CEE \quad
\textsuperscript{6}MIT DUSP \\
\textsuperscript{7}MIT Architecture \quad
\textsuperscript{8}Northeastern University \quad
\textsuperscript{9}Brown University \quad
\textsuperscript{10}McGill University%
}

\usepackage{capt-of}
\makeatletter
\AtBeginDocument{%
  \g@addto@macro\@maketitle{%
    \par\vspace{0.2em}%
    {\centering
     \includegraphics[width=1\textwidth]{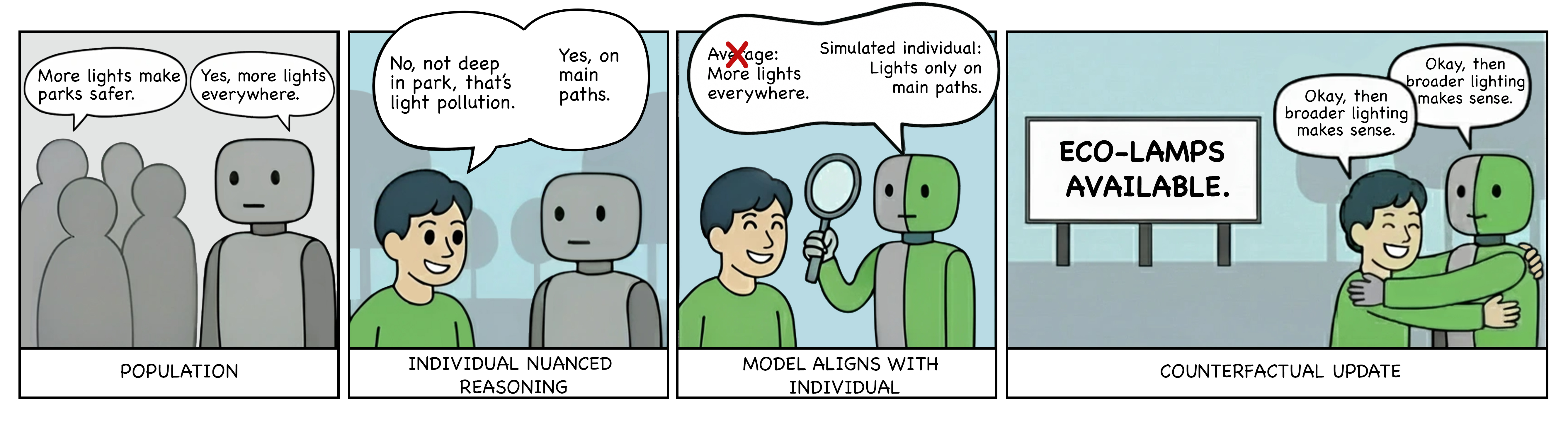}\par
     \captionof{figure}{Illustration of \textbf{HugAgent} operationalizing ``average-to-individual`` reasoning adaptation: a gray robot repeats population consensus, then observes an individual’s nuanced reasoning, gradually aligns with that individual (turning colorful), and finally adapts under counterfactual updates (e.g., with eco-lamps). This illustrates the shift from consensus mimicry to individualized reasoning.}%
     \label{fig:hugagent-average-to-individual}\par}%
    \vspace{1em}%
  }%
}
\makeatother

\begin{document}

\maketitle

\renewcommand{\thefootnote}{\fnsymbol{footnote}}
\maketitle
\renewcommand{\thefootnote}{\arabic{footnote}}
\setcounter{footnote}{0}

    

\begin{abstract}
  
    Simulating human reasoning in open-ended tasks has long been a central aspiration in AI and cognitive science. While large language models now approximate human responses at scale, they remain tuned to population-level consensus, often erasing the individuality of reasoning styles and belief trajectories. To advance the vision of more human-like reasoning in machines, we introduce \textbf{HugAgent} 
    (\textbf{\textsc{Hu}}man-\textbf{\textsc{G}}rounded \textbf{\textsc{Agent}} Benchmark), which rethinks human reasoning simulation along three dimensions: 
    (\textit{i}) from \textbf{averaged} to \textbf{individualized} reasoning, 
    (\textit{ii}) from \textbf{behavioral mimicry} to \textbf{cognitive alignment}, and 
    (\textit{iii}) from \textbf{vignette-based} to \textbf{open-ended}\footnotemark{} data. The benchmark evaluates whether a model can predict a \textit{specific person’s} \textbf{behavioral responses} and the underlying \textbf{reasoning dynamics
    } in out-of-distribution scenarios, given partial evidence of their prior views. 
    HugAgent combines structured questionnaires with semi-structured think-aloud interviews to collect ecologically valid belief states, belief updates, and reasoning traces from human participants. Our experiments reveal a clear asymmetry: models recover a person's belief state from their own context reasonably well, but struggle to predict belief updates under intervention. Cross-person and cross-domain controls trace this gap to associative matching within a topic rather than identity-consistent reasoning, suggesting that progress requires better-calibrated change detection, not simply more context. We scope the benchmark to self-reported belief reasoning in three policy domains: healthcare, surveillance, and zoning. The benchmark, along with its complete data collection pipeline and companion chatbot, is open-sourced as \href{https://github.com/jajamoa/HugAgent}{\emph{HugAgent}} and \href{https://github.com/jajamoa/trace-your-thinking}{\emph{TraceYourThinking}}.
    \end{abstract}

    \footnotetext{``Open-ended'' here describes how the data was elicited: unconstrained think-aloud protocols~\citep{ericsson1993protocol} rather than researcher-authored vignettes~\citep{martinez1999cognition}. The evaluation itself uses structured response formats, but the content being evaluated comes from participants' own unprompted reasoning.}

    \section{Introduction}
    
   \paragraph{Background.} Large language models (LLMs) are increasingly being used to simulate people: role-playing individuals, building digital twins, and generating synthetic (\enquote*{silicon}) samples to test social and policy ideas~\citep{park2023thousand, argyle2023out, xie2024can, jiang2022communitylmprobingpartisanworldviews}. These systems promise scalability and accessibility: instead of recruiting thousands of people, researchers and practitioners can use LLMs to approximate human perspectives at scale. Yet because LLMs are pretrained on population-level corpora, they tend to collapse into an ``average voice,'' capturing consensus patterns while erasing the individuality of personal histories, beliefs, and reasoning styles~\citep{wang2025large, santurkar2023whose, durmus2023towards}. 
    \begin{quote}
    \noindent\textbf{This paper asks a core question:}  
    \emph{can LLMs move from simulating the \textbf{average} to simulating the \textbf{individual}?}  
    \end{quote}
    In other words, can they predict how a specific person would think, believe, and reason in new scenarios, given evidence of their past views? 
    We formalize this broad challenge as \textbf{average-to-individual reasoning adaptation}, a measurable task that targets \emph{intra-individual fidelity} in human simulation (formally defined in Section~\ref{sec:problem}).
    
    \paragraph{Motivation.} Current benchmarks fail to capture this ability, across three key dimensions. 
    \solidcircle{1} \textbf{Intra-agent vs.\ inter-agent fidelity.} Existing pluralistic alignment benchmarks probe group dynamics and social influence~\citep{sorensen2024roadmap}, but neglect whether models can faithfully reproduce reasoning \emph{within} a single agent, which is crucial for identity-consistent modeling.
    \solidcircle{2} \textbf{Reasoning traces vs.\ behavioral outcomes.} Large-scale “digital twin” datasets such as Agent Bank \citep{park2023thousand} and Twin-2K-500~\citep{toubia2025twin2k500datasetbuildingdigital} primarily assess static behavioral outcomes, but not the evolving reasoning trajectories of a single individual, which are essential for credible social simulation~\citep{li2025positionsimulatingsocietyrequires}. 
    \solidcircle{3} \textbf{Open-ended vs.\ vignettes.} 
    Commonsense and social reasoning benchmarks (e.g., SocialIQA, ATOMIC) often reduce diverse answers to a single ground truth \citep{ sap2019atomicatlasmachinecommonsense, sap2019socialiqa}. 
    Opinion-oriented datasets likewise emphasize aggregate patterns over individual variation~\citep{argyle2023out, santurkar2023whose}. 
    Theory-of-Mind style tests typically rely on short vignettes with designer labels~\citep{wimmer1983beliefaboutbelifs, chen2024tombenchbenchmarkingtheorymind}, 
    limiting ecological validity and overlooks first-person reasoning traces as a richer gold standard~\citep{ying2025benchmark}.

    \paragraph{Methodology.} Motivated by these gaps, we introduce \textbf{HugAgent}, a benchmark targeting \emph{intra-agent fidelity} by operationalizing average-to-individual reasoning adaptation as a measurable task. For Dimension \solidcircle{1}, HugAgent shifts the granularity from \textit{inter-agent} to \textit{intra-agent} fidelity:  
    given a person’s profile and reasoning history, a model must predict both their current belief state and how it would evolve when presented with new counterfactual evidence. In Dimension \solidcircle{2}, HugAgent advances beyond static outcomes toward reasoning trajectories.  It collects \textit{first-person, out-loud self-reports} as gold-standard reasoning traces. These traces offer a deeper target for prediction than the choice outcomes or survey responses typically captured in lab experiments.  To address Dimension \solidcircle{3}, instead of relying on vignette-style benchmarks, HugAgent builds evaluation around open-ended contexts.  
    We curated real-world topics, beginning with \textit{socially and politically controversial issues that introduce inherent conflicts}.  
    Through sustained follow-up questions, the benchmark probes participants’ \textit{deliberate, System 2 style} reasoning~\citep{kahneman2011thinking, evans2013dual}, transforming the dataset from toy settings into complex, open-ended domains. For a broader discussion of prior work on personalization, social reasoning, and user modeling in LLMs, we refer readers to Appendix~\ref{app:related_work}. 
    Together, these questionnaires and interviews provide human-grounded labels for evaluating both belief states and belief updates.

        \paragraph{Contributions.}
        Our contributions span \emph{formulation}, \emph{evaluation}, \emph{diagnosis}, and \emph{infrastructure}.
        \paragraph{1. What does it mean to adapt from the average to the individual?}  
        We formalize \emph{average-to-individual reasoning adaptation} as a measurable task: predicting an individual’s beliefs and reasoning trajectory from partial self-reported data, rather than collapsing variation into an ``average'' label.
        \paragraph{2. How well do today’s models perform?}  
        We introduce \textbf{HugAgent}, a human-grounded benchmark for \emph{Belief State Inference} and \emph{Belief Dynamics Update}. Experiments with state-of-the-art LLMs establish initial baselines and reveal adaptation gaps. \url{https://github.com/jajamoa/HugAgent}

        \paragraph{3. Where do they fail, and what can improve?}
        Our experiments reveal a clear asymmetry: models recover a person’s belief state from their own context reasonably well, but struggle to predict belief updates under intervention. Cross-person and cross-domain controls trace this to associative matching within a topic, pointing to better-calibrated change detection rather than simply more context.
        \paragraph{4. How can such evaluation scale and persist?}  
        We release the entire pipeline as \textit{\textbf{open source}}, including a semi-structured interview chatbot that elicits fine-grained, “out-loud” reasoning data on arbitrary topics. This provides the community with previously lacking resources for capturing not only static answers but also the reasoning processes behind them, ensuring HugAgent is reproducible, extensible, and sustainable. \url{https://github.com/jajamoa/trace-your-thinking}

    By making ``average-to-individual'' reasoning adaptation measurable, \textbf{HugAgent} provides a concrete benchmark for studying individualized reasoning in human simulation.

    \section{Problem Setup and Theoretical Framing}
    \label{sec:problem}
    
    We operationalize individual reasoning through \emph{belief states} (snapshots) and \emph{belief dynamics} (updates under interventions). This framing allows measurable comparison while respecting the diversity of human reasoning paths.
    
    \subsection{Formalization}
    We formalize \emph{average-to-individual reasoning adaptation} by modeling an individual $i$’s belief state as a distribution over $d$ factors
    \[
    b_i \equiv P_{\phi_i}(\mathbf{s} \mid \mathcal{C}_i), \quad 
    \mathbf{s}\in\mathbb{R}^d,
    \]
    with context $\mathcal{C}_i$ (e.g., demographics, transcripts). Under an intervention $\mathcal{I}_t$, beliefs evolve via
    \[
    b_i^{t+1} = \mathcal{U}(b_i^t,\;\mathcal{I}_t), \quad 
    \Delta b_i^t = \mathbb{E}_{b_i^{t+1}}[\mathbf{s}] - \mathbb{E}_{b_i^t}[\mathbf{s}].
    \]
    
    \textbf{Tasks.}
    (i) \emph{Belief State Inference:} infer stance/factor polarity from $\mathcal{C}_i$.  
    (ii) \emph{Belief Dynamics Update:} predict stance shifts $\widehat{\Delta\mathbf{s}}_i$ given $(\mathcal{C}_i,\mathcal{I})$.

   \subsection{Theoretical Anchors: Probabilistic and Causal Perspectives}

We use normative models as anchors rather than assumptions.
(1) \textbf{Bayesian / PLoT.} Idealized revision follows Bayesian conditioning,
$b_i'(\mathbf{s}) \propto b_i(\mathbf{s})\, p(\mathcal{I}\mid \mathbf{s})$,
treating language as probabilistic evidence over latent stances. 
\textbf{Probabilistic Language of Thought (PLoT)} further motivates viewing such evidence as compositional linguistic structure \citep{goodman2015plot}.
(2) \textbf{Structural Causal Models (SCM).} Interventions act as $do(\mathcal{I})$ on a causal graph of values and reasons, yielding counterfactual shifts
$\mathbb{E}[\mathbf{s}\mid do(\mathcal{I})]$.
A person's value--reason structure can thus be viewed as a signed directed graph $G_i$, whose similarity across domains may bound transfer performance.
Human reasoning deviates from these ideals; the anchors provide principled baselines for analysis.
    
    

    \subsection{Diagnostic Guiding Questions}

Grounding HugAgent in theory leads to four diagnostic questions that serve as lenses for interpreting empirical results, rather than assumptions to be fully verified. 
(1) \textbf{Intra-individual consistency}: with sufficient context (e.g., demographic features or prior transcripts), can LLMs stably capture an individual’s belief state? We test this through per-person prediction from partial reasoning histories. 
(2) \textbf{Cross-domain transfer}: do reasoning patterns transfer across domains for the same person, and is domain-transfer accuracy lower than in-domain performance? 
(3) \textbf{Population prior reliance}: without individual context, do LLMs fall back to aggregate population priors rather than person-specific cues? We compare full-context prediction against a population-prior baseline and identity-shuffle controls. 
(4) \textbf{Context information gain}: does prediction improve monotonically with more context, or does longer context introduce noise and overload? 

Together, these questions move the benchmark beyond performance reporting: they test structural claims about how LLMs approximate, or fail to approximate the individuality of human reasoning.

    \section{HugAgent Benchmark}
    \label{sec:hugagent}
    
    Grounded in the theoretical setup in Section~\ref{sec:problem}, 
    we now introduce \textbf{HugAgent}, 
    which translates these principles into concrete tasks (Sec. \ref{sec:hugagent-task}), 
    a human-grounded data collection pipeline (Sec. \ref{sec:hugagent-pipeline}), 
    and evaluation protocols (Sec. \ref{sec:evaluation}).


    \begin{figure*}[t]
    \centering
    \includegraphics[width=\linewidth]{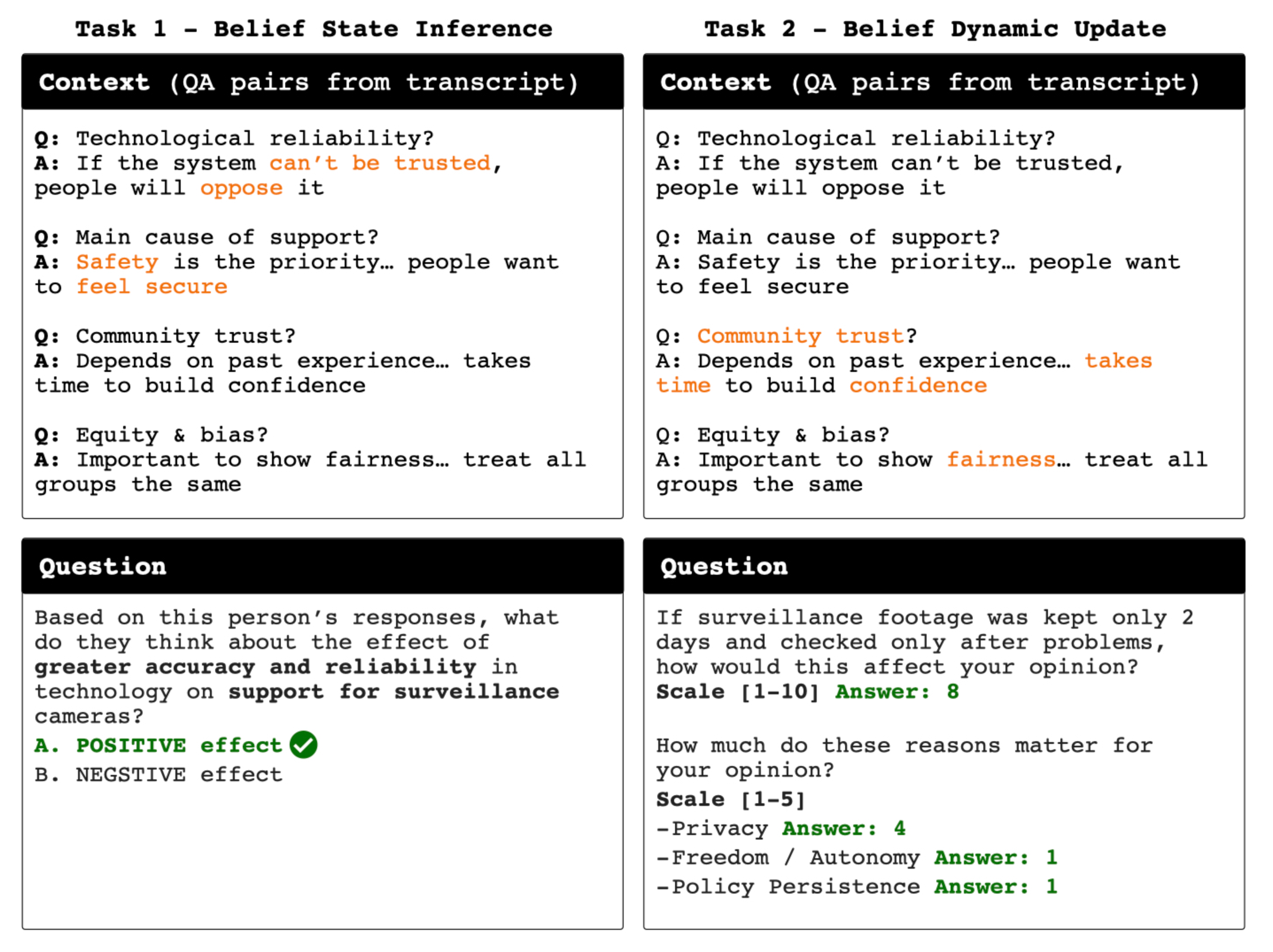}
    \caption{\textbf{Two benchmark tasks.} Task 1 (Belief State Inference) infers stance and reasons from prior context; Task 2 (Belief Dynamics Update) predicts stance shifts and the underlying reasoning under new evidence.}
    \label{fig:tasks}
\end{figure*}
    
    \subsection{Design Principles}
    \label{sec:design_p}
    \noindent (1) \textbf{Open-ended but deeper.} Emphasize depth over breadth: semi-structured dialogue with targeted follow-ups surfaces individuality while avoiding over-scaffolding; ground truth comes from self-reports \citep{kvale1996interviews, srivastava2023imitationgamequantifyingextrapolating, park2023thousand}. (2) \textbf{Two observable proxies.} We evaluate (i) belief state inference and (ii) belief dynamics update---tractable targets that avoid requiring exact trace imitation; cf.\ proxy-label benchmarks \citep{geva2021didaristotleuselaptop, ho2022wikiwhyansweringexplainingcauseandeffect, guerdan2023groundlesstruthcausalframework}. (3) \textbf{Human ceiling.} Test--retest reliability defines the upper bound, aligning with psychology standards \citep{nunnally1994psychometric, cronbach1970essentials} and recent large-scale simulations \citep{toubia2025twin2k500datasetbuildingdigital, park2023thousand}.
    See Appendix~\ref{app:design} for extended discussion.
    
    \subsection{Task Definition}
    \label{sec:hugagent-task}
 We formalize reasoning adaptation as predicting individual belief state changes under new evidence.
    \paragraph{Belief representation.} 
    A belief at time $t$ is $b_t = (s_t, \mathbf{w}_t)$, where $s_t \in \{1,\dots,10\}$ is a stance score and $\mathbf{w}_t$ is a distribution over $K$ reason weights. 
    \paragraph{Belief update.} 
    Given evidence $e$, an update operator $U$ produces $b_{t+1} = U(b_t, e)$. In HugAgent, $b_{t+1}$ is measured through participants' self-reported stance and reason updates.
    
    \paragraph{Tasks.} 
    We instantiate two tasks:  
    (i) \textbf{Belief State Inference}: predict $(s_t,\mathbf{w}_t)$ from prior responses.  
    (ii) \textbf{Belief Dynamics Update}: predict $(s_{t+1},\mathbf{w}_{t+1})$ given $(b_t,e)$.  
    Figure~\ref{fig:tasks} shows concrete examples of both tasks in HugAgent.

    \begin{figure*}[ht]
        \centering
        \includegraphics[width=\linewidth]{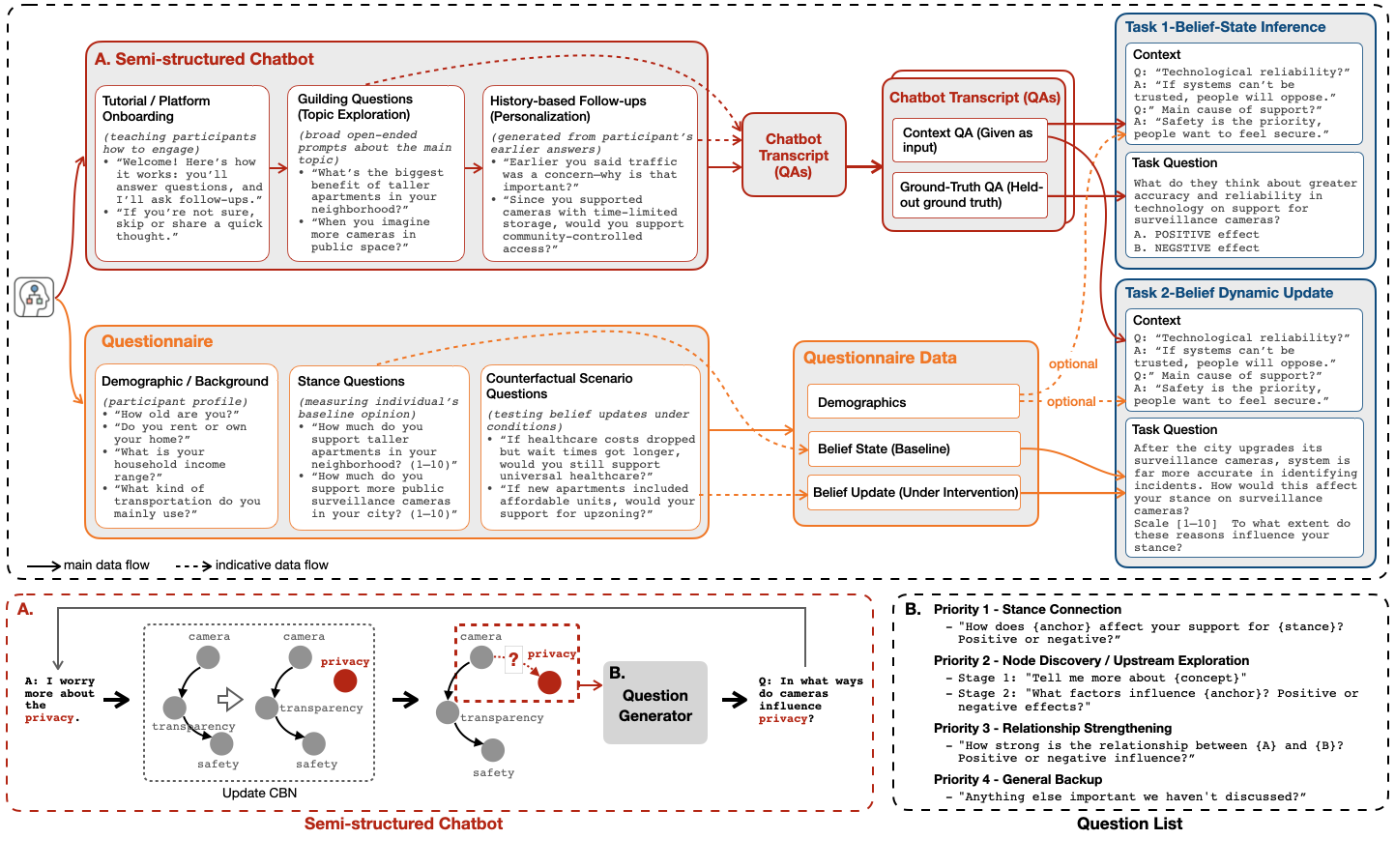}
        
        \caption{
    \textbf{HugAgent} benchmark pipeline. 
    Inputs (demographics, questionnaires, and transcripts) flow through two components: a questionnaire that provides demographic anchors, stance baselines, and counterfactual updates, and a semi-structured chatbot that elicits individualized reasoning. 
    Together these elements define two benchmark tasks: (i) \textit{belief state inference}: recovering stance and factor polarity from context, and (ii) \textit{belief dynamics update}: predicting stance shifts and reweighting under new evidence. 
    (A) The chatbot maintains a causal belief network of factors, used to identify the most critical nodes and edges for follow-up. 
    (B) A question generator derives targeted, context-specific probes from this network to structure the dialogue.}
        \label{fig:pipeline}
    \end{figure*}
    
    \subsection{Building HugAgent: Scalable Elicitation of Individual Reasoning}
    \label{sec:hugagent-pipeline}
    
    A two–stage pipeline builds HugAgent (Figure~\ref{fig:pipeline}).
    
    \textbf{1. The \textit{questionnaire} stage} collects demographics, baseline stances ($s_t$, 1–10), reason weights ($\mathbf{w}_t$, 1–5), and counterfactual interventions. These responses provide gold labels for \textit{Belief Dynamics Update} and anchor free text to factors. 
    
    \textbf{2. The \textit{chatbot} stage} elicits 8–20 question–answer pairs through semi-structured interview, combining open-ended elaborations (\emph{Context QAs}) with concise polarity judgments (\emph{GT QAs}). 
    This setup captures both participants’ belief, reasoning styles and explicit preferences on each decision factor.
    Each transcript thus supports both benchmark tasks: \textit{Belief State Inference} (using Context and GT QAs) and \textit{Belief Dynamics Update} (using Context QAs, questionnaire responses upon interventions). 
    Survey-provided updates are never revealed in dialogue, preventing leakage. 
    
    Finally, we establish a human reliability ceiling via test–retest elicitation, reporting intra-individual consistency with Intraclass Correlation (ICC) and quadratic-weighted kappa (QWK) with 95\% confidence intervals. 
    Further design choices, intervention phrasing, prompt templates, and quality-control rules are in Appendix~\ref{app:design}.

    \subsection{Dataset Statistics}
    Applying this pipeline yields HugAgent, a human-grounded dataset across three socially salient domains: \emph{healthcare}, \emph{surveillance}, and \emph{zoning}, chosen for their ecological validity, viewpoint diversity, and rich trade-offs (e.g., affordability vs.\ neighborhood character, privacy vs.\ safety). 
    From over 120 participants, we retained 54 after predefined quality-control filtering (Appendix~\ref{app:qc-protocol}). 
    The final dataset contains 356 \textit{Belief State Inference} items and 1{,}386 \textit{Belief Dynamics Update} items, summarized in Table~\ref{tab:tasks}.
    
    \begin{table}[t]
    \centering
    \footnotesize
    \renewcommand{\arraystretch}{1.08}
    \setlength{\tabcolsep}{3.5pt}
    \begin{tabular}{lrrrr}
    \toprule
    \textbf{Task} & \textbf{Health} & \textbf{Surv.} & \textbf{Zoning} & \textbf{Total} \\
    \midrule
    BSI   & 108 & 122 & 126 & 356 \\
    BDU   & 472 & 364 & 550 & 1{,}386 \\
    \midrule
    \textbf{Total} & 580 & 486 & 676 & 1{,}742 \\
    \bottomrule
    \end{tabular}
    \caption{Human-grounded dataset statistics by task and domain. BSI denotes Belief State Inference; BDU denotes Belief Dynamics Update.}
    \label{tab:tasks}
    \end{table}

       \subsection{Evaluation Protocols}
    \label{sec:evaluation}

    \paragraph{Evaluation metrics.}

    We evaluate \textit{belief state inference} using \textbf{accuracy} (exact matches). For \textit{belief dynamics update}, we report four metrics: 
(i) \textbf{accuracy}, the proportion of predictions within a tolerance band ($\pm$1 for 5-point, $\pm$2 for 10-point scales); 
(ii) \textbf{mean absolute error (MAE)}, the average deviation magnitude, normalized to a 5-point scale; 
(iii) \textbf{directional accuracy}, measuring whether the predicted update direction (increase, decrease, or no change) matches the ground truth; and 
(iv) \textbf{average to individual (ATI)} score. Following SuperGLUE~\citep{wang2019superglue}, we derive the ATI score via hierarchical aggregation, where task-specific metrics are combined into the update score and averaged with inference, using human and random baselines as normalization bounds. Appendix~\ref{sec:metrics} provides formal definitions.

    \paragraph{Leakage control.}
    We masked attribution targets, drew interventions from external surveys, and presented each item independently with minimal-overlap prompts (see Appendix~\ref{sec:prompt} for templates).
    \paragraph{Human baselines.}
    We re-contacted a subset of participants for a short-interval (14-day) test–retest study. Across all sessions, 54 participants contributed data. Of these, 18 completed the retest, and 13 were retained following a demographic consistency check. \textit{Belief State Inference} yielded an accuracy of \textbf{84.84\%} (SD = 8.90, 95\% CI: [80.00, 89.68]). \textit{Belief Dynamics Update} achieved an accuracy of \textbf{85.66\%} (SD = 7.66, 95\% CI: [80.91, 90.40]) and a mean absolute error of \textbf{0.68} (SD = 0.20, 95\% CI: [0.55, 0.80]). The directional accuracy was \textbf{88.92\%} (SD = 9.89, 95\% CI: [82.09, 96.24]). These scores establish a human consistency ceiling, as outlined in Section~\ref{sec:design_p}, against which model performance can be benchmarked.


    \section{Main Results}
    \label{sec:results}
    \subsection{Baselines} 

    We compare models against three anchors: (i) a \textbf{human upper bound} (test–retest consistency); (ii) a \textbf{random-guess lower bound}; and (iii) pretrained LLMs (\texttt{GPT}, \texttt{Gemini}, \texttt{LLaMA}, \texttt{Qwen}) as non-agentic baselines lacking memory, personalization, or retrieval.
    To evaluate agent-like structures in belief reasoning, we further include two \textbf{agent-style LLM baselines} incorporating memory or retrieval. 
    The first, \textbf{Generative Agents}, is reproduced following \citet{park2023thousandpeople} using \texttt{Qwen2.5-32B-instruct}.
    The second group includes two variants of \textbf{retrieval-augmented generation (RAG)}.  
    The first, \textbf{RAG}, follows the standard setup~\citep{lewis2020retrieval}, replacing the original full QA context with the top-$k$ retrieved QA pairs ($k=5$).  
    The second, \textbf{RAG with Full Context}, appends the retrieved QA pairs to the original input, allowing the model to jointly condition on both.  
    This evaluates whether retrieval serves as an auxiliary signal instead of a substitute for agent context.
    Appendix~\ref{app:hugtom-details} provides detailed settings.
    \subsection{Overall Performance}
    \label{sec:leave-one-out}
    Table~\ref{tab:all_results} summarizes performance. We evaluate using the metrics introduced in Section~\ref{sec:evaluation}.
    For \textbf{belief state inference}, best-performing LLMs approach but do not match human 
    accuracy, trailing by 7--9 points. Open-source \texttt{LLaMA} and \texttt{Qwen} rival \texttt{GPT-4o}, 
    while smaller or less aligned models lag significantly.  
    For \textbf{belief dynamics update}, gaps are larger: models frequently mispredict the direction 
    of stance change or fail to adjust reason weights, yielding higher error than 
    human baseline. 
    We also quantify the uncertainty from the participant pool with a participant-level bootstrap and a leave-one-participant-out check. The human--model gap and the overall ordering hold at this level. The intervals of
    closely ranked systems overlap. See Appendices~\ref{app:bootstrap}
    and~\ref{app:robustness-folds} for details.

\begin{table*}[ht]
\centering
\scriptsize 
\setlength{\tabcolsep}{2.5pt} 
\renewcommand{\arraystretch}{0.95} 

\begin{resizebox}{\textwidth}{!}{%
\begin{tabular}{lccccc}
\toprule
\textbf{Model} & \textbf{Belief State Inference} & \multicolumn{3}{c}{\textbf{Belief Dynamics Update}} & \textbf{ATI} \\
 & Acc. (\% $\uparrow$) & Acc. (\% $\uparrow$) & MAE ($\downarrow$) & Dir. Acc. (\% $\uparrow$) & (\% $\uparrow$) \\
\midrule
\rowcolor{gray!15}
Human 
& \textbf{84.84} 
& \textbf{85.66} 
& \textbf{0.68}
& \textbf{88.92} 
& \textbf{100.00}\\
\midrule
\multicolumn{6}{l}{\textit{\textcolor{gray}{OpenAI Models}}} \\
GPT-4o 
& 74.66$^{\pm0.24}$ 
& 63.11$^{\pm0.14}$ 
& 1.29$^{\pm0.00}$
& 82.27$^{\pm1.02}$ 
& 67.29$^{\pm0.74}$\\
GPT-5-mini 
& 75.30$^{\pm0.79}$ 
& 58.21$^{\pm0.50}$ 
& 1.43$^{\pm0.01}$
& 77.02$^{\pm2.02}$ 
& 61.53$^{\pm2.09}$\\
GPT5.1 Rea. High
& 73.36$^{\pm2.16}$ 
& 67.13$^{\pm0.46}$
& \textbf{1.17$^{\pm0.01}$ }
& 80.94$^{\pm3.28}$ 
& 64.66$^{\pm3.79}$\\
o3-mini 
& 75.12$^{\pm0.78}$ 
& 64.54$^{\pm0.32}$ 
& 1.22$^{\pm0.01}$
& 71.29$^{\pm2.87}$ 
& 60.92$^{\pm1.68}$\\
\midrule
\multicolumn{6}{l}{\textit{\textcolor{gray}{Other Closed-Source Models}}} \\
Claude Sonnet 4.5
& 76.04$^{\pm0.01}$ 
& \textbf{68.61$^{\pm0.09}$ }
& 1.18$^{\pm0.01}$
& 78.73$^{\pm0.10}$ 
& 67.39$^{\pm0.11}$\\
Gemini 2.0 Flash 
& 69.95$^{\pm0.12}$ 
& 60.55$^{\pm0.08}$ 
& 1.35$^{\pm0.00}$
& \textbf{83.31$^{\pm0.19}$ } 
& 59.76$^{\pm0.18}$\\
Gemini 2.5 Pro
& 75.45$^{\pm1.13}$ 
& 64.87$^{\pm0.39}$ 
& 1.27$^{\pm0.01}$
& 78.65$^{\pm1.17}$ 
& 64.29$^{\pm1.58}$\\
DeepSeek-R1 
& 75.43$^{\pm0.34}$ 
& 64.88$^{\pm0.33}$
& 1.29$^{\pm0.01}$
& 79.69$^{\pm0.91}$ 
& 67.20$^{\pm1.03}$\\
Qwen-max 
& 77.40$^{\pm0.27}$ 
& 58.86$^{\pm0.25}$ 
& 1.40$^{\pm0.00}$
& 77.17$^{\pm0.48}$ 
& 65.21$^{\pm0.49}$\\
\midrule
\multicolumn{6}{l}{\textit{\textcolor{gray}{Open-Source Models}}} \\
LLaMA 3.3 70B 
& 76.39$^{\pm0.18}$ 
& 67.57$^{\pm0.20}$ 
& 1.24$^{\pm0.00}$
& 79.56$^{\pm0.36}$ 
& \textbf{69.84$^{\pm0.27}$ }\\
Qwen2.5-32B-instr. 
& 77.17$^{\pm0.32}$ 
& 58.96$^{\pm0.05}$ 
& 1.40$^{\pm0.00}$
& 76.88$^{\pm0.29}$ 
& 64.71$^{\pm0.23}$\\
Qwen2.5-7B-instr. 
& 77.18$^{\pm0.73}$ 
& 58.82$^{\pm0.20}$ 
& 1.40$^{\pm0.00}$
& 77.12$^{\pm0.80}$ 
& 64.83$^{\pm1.21}$\\
\midrule
\multicolumn{6}{l}{\textit{\textcolor{gray}{Memory-Augmented Baselines}}} \\
RAG\footnotemark[1]
& 75.46$^{\pm0.47}$ 
& 51.90$^{\pm0.39}$ 
& 1.57$^{\pm0.05}$
& 72.25$^{\pm0.57}$ 
& 54.96$^{\pm0.87}$\\
RAG-FC\footnotemark[1]
& \textbf{77.56$^{\pm0.38}$} 
& 59.97$^{\pm0.23}$ 
& 1.39$^{\pm0.00}$
& 76.80$^{\pm0.80}$ 
& 65.65$^{\pm0.61}$\\
Generative Agents\footnotemark[1]
& 76.19$^{\pm0.36}$ 
& 58.22$^{\pm0.43}$ 
& 1.40$^{\pm0.01}$
& 76.13$^{\pm2.45}$ 
& 62.43$^{\pm1.74}$\\
\midrule
\multicolumn{6}{l}{\textit{\textcolor{gray}{Non-Learning Baselines}}} \\
Global Majority 
& 65.77$^{\pm0.00}$ 
& 58.18$^{\pm0.00}$ 
& 2.54$^{\pm0.00}$
& 17.93$^{\pm0.00}$ 
& 4.44$^{\pm0.00}$\\
Random Guess 
& 51.89$^{\pm3.96}$ 
& 43.12$^{\pm0.78}$ 
& 1.88$^{\pm0.05}$
& 46.74$^{\pm3.28}$ 
& 0.00$^{\pm5.80}$\\
\bottomrule
\end{tabular}%
}
\end{resizebox}

\caption{Average performance across domains for \textit{Belief State Inference} and \textit{Belief Dynamics Update}. Results are \textit{mean$\pm$std} over 5 runs. Best-performing model (below human upper bound) per column is highlighted in \textbf{bold}.}
\label{tab:all_results}
\end{table*}
\footnotetext[1]{RAG~\citep{lewis2020retrieval}, RAG-FC = RAG with Full Context~\citep{lewis2020retrieval}, Generative Agents~\citep{park2023thousandpeople}.}

    \section{Main Findings}
    \label{sec:main findings}
    
    \subsection*{Preserving Identity Across Domains 
     is Harder Than Expected}
     \label{sec:finding1}

    
    To evaluate whether models generalize personal context across domains, we conduct a cross-domain swap test. Each model receives QA context from one domain and is evaluated on another domain for the same participant. 
    We report \texttt{GPT-4o} (closed-source SOTA) and \texttt{Qwen2.5-32B-instruct} (strong open-source baseline) as representative models.\newline
    Under cross-domain transfer, model performance degrades substantially compared to within-domain evaluation. 
   For instance, \texttt{GPT-4o}’s \textit{belief state inference} score drops from \textbf{74.66}\% to \textbf{58.56}\%, and \textit{belief dynamics update} drops from \textbf{63.11}\% to \textbf{44.64}\%.
    This degradation is most pronounced in \textit{belief dynamics update}, suggesting models depend heavily on domain-specific cues, limiting cross-domain transfer. 
    This underscores \textit{within-person, cross-domain consistency} as a key indicator of robust generalization, suggesting that improving transferability requires focusing on domain-relevant context instead of surface-level correlations.(Appendix Table~\ref{tab:cross-topic})
    
    \subsection*{Belief Updating Depends on High-Signal Context, Not More Context}
    \label{sec:context-ablation}
    

    
    To examine how context length affects the two tasks, we vary the number of Context QAs and evaluate representative closed-source and open-source models. 
    
\begin{figure}[ht]
        \centering
        \includegraphics[width=\linewidth]{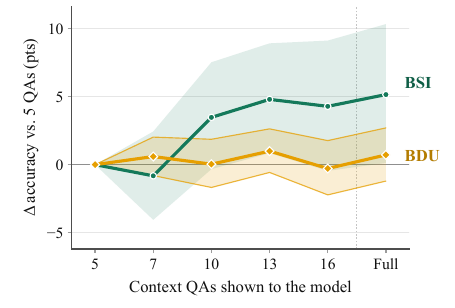}
        
        \caption{Change in accuracy relative to the shortest context, macro-averaged across the three domains. Bands show 95\% confidence intervals from a participant-level cluster bootstrap (over 54 participants), computed separately for each context length without correction for multiple comparisons. Accuracy at the 5-QA baseline is 71.1\% for \textit{Belief State Inference} and 66.2\% for \textit{Belief Dynamics Update}.}
        \label{fig:context not help}
    \end{figure}

    Figure~\ref{fig:context not help} shows a clear asymmetry. \textit{Belief State Inference} generally improves with more context, reaching $+5.2$ points at full context, although the intervals are wide and no individual comparison remains significant after correcting for the five comparisons. \textit{Belief Dynamics Update}, in contrast, barely changes: every estimate stays within 1.0 point of the baseline, with intervals contained within $\pm2.7$ points. The main result is therefore the difference in the overall pattern: more context tends to help belief-state inference, but shows little evidence of helping belief updating.

\begin{figure}[ht]
        \centering
        \includegraphics[width=\linewidth]{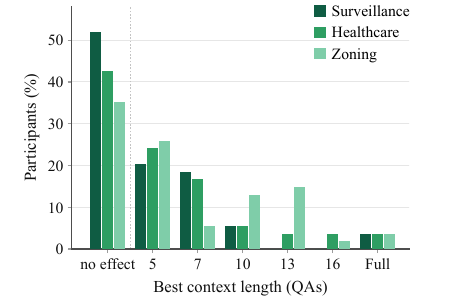}
        
        \caption{Best context length for belief updating, by participant and domain, using \texttt{qwen-plus-2025-09-11} over $\{5,7,10,13,16,\mathrm{full}\}$. Cells with identical accuracy across all context lengths are labeled \emph{no effect} rather than assigned to the shortest-context bucket; these account for 43\% of the 162 participant–domain cells.}
        \label{fig:individual group}
    \end{figure}

    Further analysis suggests that the issue is not length alone, but the location of high-signal evidence. In semi-structured interviews, participants reveal decisive reasons at different points; many provide key information early, while later exchanges add elaboration without improving update prediction. At the individual level, the pattern is even clearer than the average curve suggests (Figure~\ref{fig:individual group}). For 43\% of participant–domain cells, accuracy is identical across all context lengths, meaning that additional context makes no difference at all. Among the cells where context length does matter, shorter context usually performs best, while full context is uniquely best for only 2 of 54 participants in each domain. Masking and reordering experiments show that removing or displacing high-signal QAs hurts more than replacing later low-signal QAs with unrelated text (Appendix~\ref{app:context-ablation}).
    This asymmetry suggests static belief-state inference benefits from broader coverage, while belief updating depends on compact, diagnostic evidence, with additional context diluting earlier signals or inducing a recency bias toward later, less relevant exchanges. Logit-level analyses support this interference: in BDU-Surveillance and BDU-Healthcare, longer context increases the model's confidence margin on wrong updates, suggesting that additional context makes models confidently wrong rather than better informed (Appendix Figure~\ref{fig:logit}).

    \section{Why It Happens: Diagnostic Ablations}
    \label{sec:diagnostic}
   
    Section~\ref{sec:main findings} revealed a \textbf{consistent failure to preserve identity across domains}. Here we ask \textit{\textbf{why}}: is it because models never learned to use personal information, or because they use it in an \textit{associative, non-generalizable} way? 
    To answer this, we conduct two diagnostic ablations.
    
    \subsection{Population Prior vs.\ Individual Context}
    \label{sec:pop-prior}
        
    We first test whether providing individual context leads to better model performance than relying solely on population-level priors. 
    
    The \textit{No-Context} setting uses only demographic background, while \textit{Full-Context} additionally includes transcripts and survey answers. 
    For \texttt{GPT-4o}, full context raises belief-state inference accuracy from \textbf{58.49}\% to \textbf{74.66}\%, and belief-dynamics-update accuracy from \textbf{39.83}\% to \textbf{63.11}\% (Appendix Table~\ref{tab:global-prior}).
    The consistent gains in\textit{ Full-Context} indicate that models leverage individual cues rather than population priors, showing that the benchmark captures identity-sensitive reasoning rather than general demographic trends.

    \subsection{Cross-Person Generalization}
    \label{sec:identity-shuffle}

    We then test whether the observed gains reflect genuine identity modeling or simply the benefit of having richer, more fine-grained context. 
    In the \textit{Cross-Person} setting, QA context from one participant is used to predict another participant’s responses within the same domain. 
Performance drops sharply: for \texttt{GPT-4o}, belief-dynamics-update accuracy falls to \textbf{39.30}\% with an MAE of \textbf{1.93} (Appendix Table~\ref{tab:cross-person}). 
 These results suggest that improvements in the \textit{Full-Context} setting arise from learning identity-specific patterns rather than simply benefiting from additional contextual detail.
    
    \paragraph{Summary.}
    Together with the findings in Section \ref{sec:main findings}, these ablations reveal that current models’ failure to preserve identity across domains is not because they ignore individual context in inference, but rather implies a reliance on associative context matching instead of identity-consistent reasoning.

    \section{Error Analysis: Bias Patterns and Sources of Failure}
    \label{sec:error-analysis}
    \paragraph{Domain dependence and cross-domain generalization.}

    Performance varies systematically across domains. 
    Within-domain, models often perform better on \emph{Surveillance} for \textit{Belief State Inference}, while the strongest \textit{Belief Dynamics Update} performance varies by domain and model. 
    However, cross-domain transfer yields much sharper degradation. 
    For example, when using \emph{Surveillance} context to predict \emph{Zoning}, \texttt{Qwen2.5\mbox{-}32B\mbox{-}instr} drops from \textbf{62.25}\% in-domain BDU accuracy to \textbf{24.84}\%, with MAE increasing from \textbf{1.33} to \textbf{1.94}. 
    This suggests that models rely heavily on domain-specific contextual cues rather than domain-general person-level reasoning patterns. 
    Full domain-level results are provided in Appendix~\ref{app:hugtom-details}.\\
    \textbf{Implication.} LLMs lack shared cross-domain regularities in belief updating; their \textit{belief dynamics} rely on corpus-specific semantic co-occurrences rather than abstract cognitive mechanisms. Therefore, cross-domain degradation is not a simple domain shift, but exposes a lack of domain-general inductive bias for individualized simulation.

    \paragraph{Directional error decomposition.}
    We decompose directional errors into: (1) \textbf{change-detection error} (failure to detect if a belief change occurred), and (2) \textbf{direction-inference error} (failure to predict change direction: \emph{increase}, \emph{decrease}, or \emph{no change}). Appendix~\ref{sec:metrics} provides details.   
    Across systems, \emph{change-detection} errors predominate. In \emph{Healthcare}, \texttt{GPT-4o} attains a change-detection accuracy of \textbf{49.36}\%, compared to \textbf{88.89}\% for direction-inference (Dir. Acc.~77.03\%).  
    Similarly, \texttt{Qwen2.5-32B-Instruct} records \textbf{33.19}\% versus \textbf{84.29}\% (Dir. Acc.~68.96\%). \\
    This cross-domain asymmetry yields high \emph{direction-inference} but lower \emph{change-detection} performance, limiting \emph{overall directional accuracy}.
    These errors suggest models preserve stance change \emph{signs} once detected but fail to capture \emph{when} updates should occur. Instead of inverting polarity, they remain static despite contextual cues, reflecting a change-averse bias toward \textit{no change}.
    This pattern holds for strong models such as \texttt{Qwen-Max} and \texttt{GPT-4o}, which achieve moderate directional accuracy ($\approx$77--82\%), lagging behind human performance (Dir. Acc. 88.92\%) by 7--12 percentage points (Appendix~\ref{app:annotation}).
    Notably, low mean absolute error (e.g., \texttt{o3-mini}, MAE~${1.22}$) does not imply higher directional accuracy, confirming magnitude alignment alone cannot ensure correct update directionality.  \\
    \textbf{Implication.} LLMs tend to be \textit{directionally accurate} yet \textit{change-averse}, preserving prior beliefs instead of updating without strong evidence. This implicit stability bias suggests future models \textbf{require calibrated change detection}, potentially via continuous output representations or confidence-aware designs.

    \section{Mitigation Strategies: Insights and Guiding Principles}
    \label{sec:mitigation}
    
    Our diagnostics reveal three recurring failure modes that suggest directions for future model design. \textbf{(1) Identity-consistent generalization.} Full-context gains disappear under cross-person settings, suggesting associative matching rather than person-specific reasoning; future models may need identity-conditioned representations with cross-domain consistency constraints.
    \textbf{(2) Calibrated belief updating.}
    Directional error decomposition shows that models often preserve prior beliefs even when cues signal change; future systems should better calibrate \textit{when} to update, not only what direction to update toward. \textbf{(3) Context prioritization.}
        Context-length ablations show that more context does not reliably improve belief updating, suggesting the need to prioritize high-signal evidence rather than simply expanding context windows.

        

        

        \paragraph{Synthesis.}
        Together, these patterns position HugAgent as a diagnostic benchmark: it reveals not only accuracy, but also structured biases that standard metrics (accuracy, MAE) often obscure.

    \section{Discussion and Open Challenges}

    Our results highlight three open challenges for individualized reasoning benchmarks. 
    First, \textbf{topic selection is not neutral}: controversial domains surface richer value trade-offs and sharper belief updates, while homogeneous topics may compress individual variation and inflate apparent accuracy. Future evaluations should therefore treat topic choice as an explicit experimental variable rather than a hidden confound. 
    Second, \textbf{elicitation must balance intuition and deliberation}: open-ended interviews can collapse into shallow reactions, but excessive scaffolding may overwrite individual reasoning. 
    Lightweight probes (e.g., counterfactual questions) may help elicit reflective updates without enforcing a single ``correct'' trace. 
    Third, \textbf{cross-domain transfer should remain central}: models that fit one domain may still fail to preserve a person's reasoning pattern elsewhere. 
    Future benchmarks should compare varying controversy levels and report within-person, cross-domain generalization alongside standard accuracy.
    Ultimately, individual reasoning requires not just better models, but better questions: trade-off-eliciting topics, deliberate protocols, and cross-domain fidelity instead of single-domain fit.
    
    \section{Conclusion}
\textbf{HugAgent} introduces a human-grounded benchmark for \emph{average-to-individual reasoning adaptation}, evaluating individualized belief states, belief updates, and open-ended reasoning traces. Across three policy domains, models benefit from a person’s own context, but lose nearly all of that benefit with context from another person or domain. Belief updating remains well below the human test–retest ceiling. The elicitation pipeline, two-task structure, and diagnostics are domain-agnostic, while reported accuracy levels are specific to self-reported belief reasoning in healthcare, surveillance, and zoning.
    
    \section*{Limitations}
    

Our study is constrained by the scale and scope of the human dataset. Collecting high-fidelity individualized reasoning data is substantially more expensive than collecting standard survey responses, since each participant must complete structured questionnaires, semi-structured think-aloud interviews, and counterfactual update tasks. As a result, the current human track contains a modest number of retained participants, and the test--retest subset is smaller still. While this design provides richer individual-level supervision than conventional vignette or outcome-only benchmarks, future work should expand the participant pool and further validate reliability across larger and more diverse populations.

The benchmark currently covers three socially salient policy domains: healthcare, surveillance, and zoning. These domains were chosen because they elicit value trade-offs and belief updates, but they do not exhaust the space of human reasoning. Individualized reasoning may behave differently in lower-stakes topics, interpersonal scenarios, scientific reasoning, consumer choice, or culturally specific contexts. The present results should therefore be interpreted as evidence about individualized policy reasoning rather than a complete account of human reasoning simulation.

HugAgent also relies on self-reported beliefs, reason weights, and belief updates. These reports provide a practical and interpretable proxy for reasoning, but they should not be treated as direct access to underlying cognitive processes. Participants may simplify, rationalize, or revise their explanations during elicitation. In addition, our structured evaluation tasks discretize complex reasoning into stance scores, factor polarities, and update labels, which may underrepresent ambiguity or multiple coexisting motivations.

Finally, the benchmark remains sensitive to elicitation and modeling choices, including chatbot follow-up questions, prompt format, model version, and context length. We release the data collection pipeline, prompts, and evaluation scripts to make these choices transparent and to support future extensions across domains, populations, and model families.

\section*{Ethical Considerations}

This work adheres to the ACL Code of Ethics and was conducted under an approved Institutional Review Board (IRB) protocol.  \\
\textbf{Human participants.} Human data were collected via Prolific with informed consent and fair compensation. All identifying information was removed prior to analysis, and only anonymized transcripts and survey responses are used in the benchmark. Participants were free to withdraw at any time, and only de-identified data will be released. \\ 

\textbf{Risks and mitigations.} Potential risks include the reproduction of demographic or topical biases when models are trained or evaluated on HugAgent. To mitigate these risks, we (i) release data to allow sensitivity analyses, (ii) provide documentation of demographic distributions.  \\
\textbf{Misuse.} HugAgent is a research benchmark for evaluating models.
It must not be used to persuade people, target individuals, or change anyone's beliefs. \\
\textbf{Data security and release.} All human data are securely stored and released only in anonymized form.  Code, evaluation scripts, and data releases will follow open-science best practices with explicit license terms and a retraction mechanism in case of unforeseen issues.  \\
Overall, we commit to transparency, reproducibility, and responsible usage. The benchmark is intended solely for advancing research on individualized reasoning in language models and not for deployment in decision-making that could affect individuals or communities.

\section*{Reproducibility Statement}

We have taken extensive measures to ensure that the HugAgent benchmark and the experiments reported in this paper are fully reproducible.  

\textbf{Human data collection.} The survey and chatbot protocols are provided in full in Appendix~\ref{survey:general},\ref{survey:zoning}, \ref{survey-healthcare}, \ref{survey:surveillance-survey},  including question templates, intervention designs, and filtering criteria for valid participants. Recruitment, compensation, and exclusion rules follow a standardized IRB-approved protocol, ensuring that the human dataset can be replicated with identical procedures.  


\textbf{Evaluation.} All tasks use standardized metrics described in Appendix~\ref{sec:metrics}, including accuracy, F1, mean absolute error (MAE). We release the evaluation scripts and configuration files with fixed random seeds to ensure identical results across runs.  

\textbf{Release and documentation.} We release the full benchmark and pipeline as \textit{open source}, including (i) the \textbf{HugAgent benchmark} (data, evaluation scripts, and leaderboard) and (ii) the \textbf{trace-your-thinking pipeline} (semi-structured chatbot and transcript generation). 
Both repositories are publicly available: \\
\url{https://github.com/jajamoa/HugAgent} \\
\url{https://github.com/jajamoa/trace-your-thinking} \\
The code is released under the MIT license and the de-identified participant data under CC BY-NC 4.0, with intended-use terms restricting the data to research on evaluating individualized reasoning. The release also includes data versioning and a retraction mechanism.

Together, these measures ensure that our benchmark, experiments, and results are reproducible by independent researchers.

\section*{Acknowledgments}

We thank Deb Roy for the conceptual discussions that shaped how this work is
framed, and the anonymous reviewers and the area chair, whose comments prompted
the participant-level bootstrap, leave-one-out, and sensitivity analyses
reported here.

This work received no external funding. Participant compensation, which exceeded
USD 8{,}350, was covered in part by the City Science group at the MIT Media Lab
and in part by the authors.

We used a large language model to polish the language of the manuscript. It was not involved in the ideation, methodology, experiments, or analysis; a full statement is in Appendix~\ref{app:llm-usage}.
\bibliography{latex/custom}

\appendix

\newpage
\section{Related Work}
\label{app:related_work}

\paragraph{Social Simulation, Digital Twins, and Population Panels.}
A growing line of work simulates societies with LLM agents. Early “silicon samples” use LMs to approximate human samples in stylized tasks \citep{argyle2023out}, while \textit{Generative Agents} extend to rich daily-life environments with memory and social coordination~\citep{park2023thousandpeople}. This has scaled to population panels—e.g., simulations of \emph{1,000 people} \citep{park2023thousand} and digital-twin datasets such as \textit{Twin-2K-500} \citep{toubia2025twin2k500datasetbuildingdigital}—as well as community/role-play platforms for real-time social interaction \citep{zhao2023lyfe} and personification benchmarks \citep{xie2024human,xie2024can}. These approaches provide breadth and largely \emph{static} outcome measures; \textbf{HugAgent} adds \emph{depth} via think-aloud transcripts that trace reasoning trajectories, counterfactual interventions to test belief updates, and a human test–retest reliability ceiling to anchor claims \citep{Wurgaft2025ScalingThinkAloud,nunnally1994psychometric,cronbach1970essentials}.

\paragraph{Social Reasoning and Theory of Mind.}
Work on Theory of Mind (ToM) in AI draws from developmental psychology tests such as the false-belief task \citep{wimmer1983beliefaboutbelifs}, Sally-Anne \citep{baroncohen1985tomautism}, and Strange Stories \citep{happe1994advancedtom}, later reformulated as computational tasks \citep{nematzadeh2018evaluating,rabinowitz2018machinetheorymind}. Scaled language models brought ToM into broad benchmarks \citep{le2019revisiting,srivastava2023imitationgamequantifyingextrapolating,chen2024tombenchbenchmarkingtheorymind} and inspired synthetic testbeds such as BigToM \citep{gandhi2023understandingsocialreasoninglanguage}, HI-TOM \citep{he2023hitombenchmarkevaluatinghigherorder}, FANToM \citep{kim2023fantombenchmarkstresstestingmachine}, and MMToM-QA \citep{jin2024mmtom}. More recent directions ground ToM in dialogues and social contexts \citep{chan2024negotiationtombenchmarkstresstestingmachine,sap2019socialiqa,strachan2024testingtom}, or frame it through Bayesian belief attribution \citep{ying2024grounding}. Yet these benchmarks remain synthetic, vignette-based, and decontextualized, missing ecological and demographic variability \citep{wang2025large,stewart2017crowdsourcing}.

Parallel lines in AI reasoning emphasize world and agent models: causal world modeling \citep{wong2022causal,lake2017building,ellis2023dreamcoder} and the LAW framework, which coordinates world, agent, and language models \citep{hu2023languagemodelsagentmodels}. Within this framing, HugAgent extends ToM evaluation by asking whether models can map natural language into personalized belief states and update them consistently under interventions, bridging synthetic ToM tasks and socially grounded reasoning.


\newpage
\section{Auxiliary Dataset}
\label{appendix:aux}
Our core benchmark is designed around interview transcripts, where each data point consists of demographic information, a context in the form of question--answer pairs, and ground-truth first-person self-reports. This setup defines a clean belief inference language task: given demographic cues and conversational context, models must infer individual beliefs and predict reactions.

To complement this benchmark, we also release an auxiliary demographics-only dataset (39 users, each with survey responses and demographic attributes). While these records do not support direct belief inference, they enable principled baselines and transfer settings. Concretely, we implemented a demographic--linear regression model based on survey responses, and further tested Qwen-Plus on the main belief dynamics update task with auxiliary supervision from a subset of 15 users. Results from both the demographic-linear baseline and the augmented Qwen-Plus setting are reported in Table~\ref{tab:auxiliary-results}, illustrating how population-level priors can inform personalized inference.

\begin{table*}[t]
\centering
\caption{Results on the auxiliary demographics-only setting. Comparison of a demographic-only prior baseline and Qwen-Plus with auxiliary supervision across three domains. Higher accuracy and lower MAE indicate better performance.}
\label{tab:auxiliary-results}
\begin{tabular}{lcccc}
\toprule
\multirow{2}{*}{Domain} & \multicolumn{2}{c}{Demographic-only prior} & \multicolumn{2}{c}{Qwen-Plus w/ auxiliary data} \\
\cmidrule(lr){2-3} \cmidrule(lr){4-5}
 & Accuracy & MAE & Accuracy & MAE \\
\midrule
Healthcare   & 0.130 & 2.954 & 0.524 & 1.792 \\
Zoning       & 0.202 & 2.954 & 0.474 & 1.765 \\
Surveillance & 0.081 & 2.797 & 0.589 & 1.819 \\
\bottomrule
\end{tabular}
\end{table*}

\newpage
\section{HugAgent Benchmark Details}
\label{app:hugtom-details}

We introduce the \textbf{HugAgent Benchmark} (\textit{HUman-Grounded Theory of Mind}), a new evaluation suite for reasoning fidelity in generative agents. HugAgent formalizes the task of \textit{causal BN reconstruction from human interviews}, and provides (1) a dataset of annotated causal belief graphs derived from natural language Q\&A, and (2) multi-level evaluation metrics (node, edge, motif) to assess structural alignment between human and agent reasoning.

Unlike prior benchmarks focused on behavioral imitation or chain-of-thought generation, HugAgent directly evaluates whether agents can reconstruct the latent causal structures that underlie human judgments. This benchmark responds to recent concerns that LLM-based agents risk flattening individual identity representations by grounding evaluation in structured, human-annotated causal beliefs.

\subsection*{Data schema and cognitive grounding}
Inspired by cognitive science, we use causal BNs to represent the reasoning structures underlying human decision-making. Rather than modeling surface discourse, our schema captures latent causal dynamics by explicitly linking belief variables, affective states, and behavioral intentions in a directed graph. This design supports psychologically grounded and structurally coherent representations of human reasoning.

\paragraph{Schema Design}
Each participant’s causal BN is represented as a structured JSON object composed of three main components: \texttt{nodes}, \texttt{edges}, and \texttt{qa\_history}. This schema is designed to encode causal beliefs extracted from interviews, with each element indexed by a unique ID to enable motif analysis, simulation, and evidence tracing.

\subparagraph{Nodes}
Each node represents a belief concept and includes a \texttt{label}, a model-generated \texttt{confidence} score (ranging from 0.0 to 1.0), and a list of \texttt{source\_qa} IDs that support the node's existence. Nodes also track their \texttt{incoming\_edges} and \texttt{outgoing\_edges} for efficient graph traversal.

\subparagraph{Edges}
Edges capture directed causal links between nodes. Each edge includes a \texttt{source} node ID, a \texttt{target} node ID, and an \texttt{aggregate\_confidence} score reflecting the model’s overall belief in the causal connection. A \texttt{modifier} (in the range \([-1.0, 1.0]\)) represents the direction and strength of influence: positive values indicate causal support, negative values indicate inhibition. Each edge is backed by a list of individual QA-based evidence entries with associated confidence scores.

\subparagraph{QA History}
The \texttt{qa\_history} component stores raw interview responses, mapping each QA pair to its corresponding extracted causal relations. Each QA entry includes the original question and answer texts, as well as a list of \texttt{extracted\_pairs}, where each pair links a source node to a target node with a confidence score.

This data structure supports fine-grained analysis of belief formation, causal reasoning, and evidence provenance across participants.

\subsection*{Dataset Construction}
We collected over 100 interviews from participants recruited through the Prolific platform. Topics—such as urban upzoning, surveillance cameras, and universal healthcare—were chosen to elicit reflective, ecologically valid reasoning.

\subsection*{Transcript Collection}
To construct structured causal BNs from qualitative interviews, we developed a \textbf{semi-structured, cognitively grounded elicitation framework}. This framework guides LLMs in extracting interpretable causal structures from natural language dialogue and generating follow-up questions that balance open-ended exploration with targeted inquiry~\citep{chickering2002optimal, pohontsch2015cognitive}. 

\subsection*{Human Annotation Protocol}
\label{app:annotation}
We asked annotators to label causal BNs using soft labels, capturing graded beliefs and allowing for variation across annotators. Our human-in-the-loop annotation tool supports annotators in assigning confidence scores to each node and edge\citep{zhang2023human}.

Annotators were also recruited from Prolific\citep{stewart2017crowdsourcing}. During selection, we followed three principles: (1) double-blind annotation, (2) matching annotators with similar backgrounds, and (3) using shared guidelines to maintain consistency. Annotation instructions were carefully designed to ensure reproducibility and interpretability.

Annotators were compensated at \$15 per hour, which exceeds the prevailing minimum wage in their countries of residence. The instruction page explicitly stated that the annotations would be used for academic research and released as part of a public benchmark, and annotators provided informed consent before starting. The data collection protocol was reviewed and approved by the MIT Committee on the Use of Humans as Experimental Subjects (COUHES).

\subsection*{Evaluation Settings}
All models are evaluated under a consistent inference setup. 
We fix the random seed to \textbf{42} and set the temperature to \textbf{0.1} for all experiments. 
Models from the \texttt{GPT} and \texttt{Gemini} series are executed in \textit{batch inference} mode, 
while all other models use \textit{real-time completion} inference. 
All outputs are constrained using \textbf{function calling} to ensure structured and valid responses. 
Prompts adopt a pure \textit{in-context learning} format without any examples or reasoning demonstrations. 

For the \textbf{Generative Agents} setting, following \citep{park2023thousandpeople}, we employ \texttt{Qwen2.5-32B-instruct} to analyze the dialogue transcript and generate three high-level expert reflections that serve as auxiliary reasoning cues during inference. 
For the two retrieval-augmented variants of RAG\citep{lewis2020retrieval}, we adopt a TF-IDF retriever to identify the top five most relevant QA pairs. 

\subsection*{Human Baselines}
We re-contacted a subset of participants for a short-interval (14-day) test–retest study. Across all sessions, 54 participants contributed data. Of these, 18 completed the retest, and 13 were retained following a demographic consistency check.

\textit{Belief State Inference} yielded an accuracy of \textbf{84.84\%} (SD = 8.90, 95\% CI: [80.00, 89.68]). 
By topic, accuracies were \textbf{87.50\%} for \textit{Surveillance} (SD = 8.29, 95\% CI: [82.81, 92.19]), \textbf{81.54\%} for \textit{Zoning} (SD = 15.11, 95\% CI: [73.32, 89.75]), and \textbf{84.53\%} for \textit{Healthcare} (SD = 13.90, 95\% CI: [76.97, 92.09]). 

\textit{Belief Dynamics Update} achieved an accuracy of \textbf{85.66\%} (SD = 7.66, 95\% CI: [80.91, 90.40]) and a mean absolute error of \textbf{0.68} (SD = 0.20, 95\% CI: [0.55, 0.80]). 
Across topics, accuracy was \textbf{85.75\%} for \textit{Healthcare} (SD = 10.68), \textbf{85.33\%} for \textit{Surveillance} (SD = 12.98), and \textbf{85.86\%} for \textit{Zoning} (SD = 10.08). 
Corresponding mean absolute errors were \textbf{0.66} (SD = 0.28, 95\% CI: [0.49, 0.83]) for Healthcare, \textbf{0.72} (SD = 0.35, 95\% CI: [0.50, 0.93]) for Surveillance, and \textbf{0.67} (SD = 0.28, 95\% CI: [0.49, 0.84]) for Zoning.

In the \textit{Belief Dynamics Update} tasks,
we decompose directional accuracy into two components: 
(1) \textbf{change detection}—the model’s ability to detect whether a belief change has occurred, and 
(2) \textbf{direction inference}—the model’s ability to predict the direction of that change (increase, decrease, or no change)
(detailed definitions are provided in  Appendix~\ref{sec:metrics}).

Overall, the directional accuracy was \textbf{88.92\%} (SD = 9.89, 95\% CI: [82.09, 96.24]).
The results across topics are summarized as follows:
\begin{itemize}
   \item \textbf{Healthcare:} change detection = 80.00\% (SD = 25.82, 95\% CI: [61.53, 98.47]); direction inference = 96.67\% (SD = 10.54, 95\% CI: [89.13, 100.00]); directional accuracy = 91.67\% (SD = 11.06, 95\% CI: [83.76, 99.58]).
    \item \textbf{Surveillance:} change detection = 90.00\% (SD = 16.10, 95\% CI: [78.48, 100.00]); direction inference = 88.33\% (SD = 19.33, 95\% CI: [74.51, 100.00]); directional accuracy = 88.83\% (SD = 15.15, 95\% CI: [77.99, 99.67]).
    \item \textbf{Zoning:} change detection = 80.00\% (SD = 23.31, 95\% CI: [63.33, 96.67]); direction inference = 90.00\% (SD = 21.08, 95\% CI: [74.92, 100.00]); directional accuracy = 87.00\% (SD = 18.14, 95\% CI: [74.03, 99.97]).

\end{itemize}

\subsection*{Human Demographic Breakdown}
To provide a clearer view of the diversity represented in our human participant cohort, we report detailed demographic distributions across age, income, education level, and occupation. As shown below, the 54 participants span a broad range of backgrounds, supporting the use of this sample for individualized reasoning analysis.

\subsection*{High-signal evidence matters more than context volume.}
\label{app:context-ablation}
\begin{figure*}[h]
    \centering
    \includegraphics[width=0.75\linewidth]{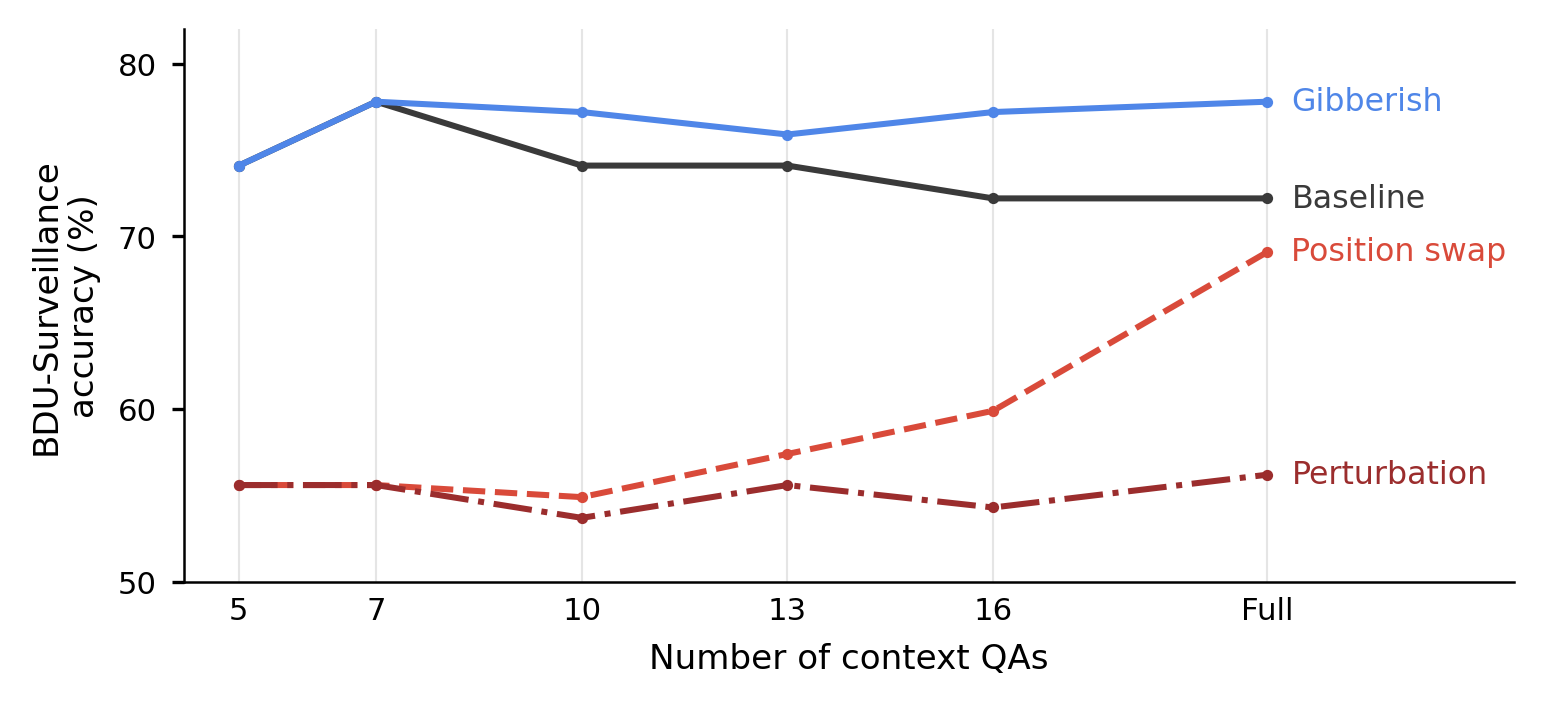}
    \caption{\textbf{Ablating high-signal context in BDU-Surveillance.} Using \texttt{qwen-plus-2025-09-11} on a 23-user BDU-Surveillance subset, we compare the original chronological context with three ablations: moving early QAs to the end, replacing later QAs with unrelated text, and removing early QAs. Preserving early high-signal QAs maintains performance, while removing or displacing them substantially reduces accuracy, showing that belief updating depends on locating diagnostic evidence rather than simply adding more context.}
    \label{fig:context-ablation}
\end{figure*}

To test whether longer context helps because it provides more evidence, or whether performance depends on the placement of a few diagnostic QAs, we conduct an ablation study on BDU-Surveillance using \texttt{qwen-plus-2025-09-11}. We use the 23-user subset from the full-context sweep and compare four context variants: the original chronological context, a \textit{position-swap} variant that moves the first seven QAs to the end, a \textit{gibberish} variant that keeps the first seven QAs but replaces later QAs with unrelated weather QA pairs, and a \textit{perturbation} variant that removes the first seven QAs and retains only later context.
Figure~\ref{fig:context-ablation} shows that the early QAs contain most of the useful evidence for belief updating. Keeping these early QAs while replacing later context with unrelated text preserves, and sometimes slightly improves, performance compared with the original context. In contrast, removing the early QAs causes a large drop in accuracy, and moving them to the end substantially hurts performance under shorter context windows. These results indicate that BDU performance is not driven by context length alone. Models benefit when high-signal evidence is accessible in the effective context window; adding or preserving low-signal later context is much less important. This supports our interpretation that belief updating requires identifying diagnostic personal evidence rather than simply conditioning on more text.

\subsection*{Full Results}

Full results are shown in Table \ref{tab:sota-model}.


\begin{table*}[t]
    \centering
    \begin{adjustbox}{max totalsize={\textwidth}{\textheight},keepaspectratio}
    \begin{tabular}{p{3.8cm}ccccc}
    \toprule
    \multirow{2}{*}{\textbf{Model}} 
    & \textbf{Belief State Inference} 
    & \multicolumn{3}{c}{\textbf{Belief Dynamics Update}} 
    & \multirow{2}{*}{\textbf{ATI (\% $\uparrow$)}}\\
    \cmidrule(lr){2-2}\cmidrule(lr){3-5}
    & Acc. (\% $\uparrow$) 
    & Acc. (\% $\uparrow$) 
    & MAE ($\downarrow$)
    & Dir. Acc. (\% $\uparrow$) 
    & \\
    \midrule
    \rowcolor{gray!15}
    Human\rule{0pt}{2.8ex}\rule[-1.2ex]{0pt}{0pt}
    & \textbf{84.84} 
    & \textbf{85.66} 
    & \textbf{0.68}
    & \textbf{88.92} 
    & \textbf{100.00}\\
    \midrule
    \multicolumn{6}{l}{\footnotesize\textit{\textcolor{gray}{OpenAI Models}}} \\
    GPT-4o 
    & 74.66$^{\pm0.24}$ 
    & 63.11$^{\pm0.14}$ 
    & 1.29$^{\pm0.00}$
    & 82.27$^{\pm1.02}$ 
    & 67.29$^{\pm0.74}$\\
    GPT-5-mini 
    & 75.30$^{\pm0.79}$ 
    & 58.21$^{\pm0.50}$ 
    & 1.43$^{\pm0.01}$
    & 77.02$^{\pm2.02}$ 
    & 61.53$^{\pm2.09}$\\
    GPT5.1 Rea. High
    & 73.36$^{\pm2.16}$ 
    & 67.13$^{\pm0.46}$
    & \textbf{1.17$^{\pm0.01}$}
    & 80.94$^{\pm3.28}$ 
    & 64.66$^{\pm3.79}$\\
    o3-mini 
    & 75.12$^{\pm0.78}$ 
    & 64.54$^{\pm0.32}$ 
    & 1.22$^{\pm0.01}$
    & 71.29$^{\pm2.87}$ 
    & 60.92$^{\pm1.68}$\\
    \midrule
    \multicolumn{6}{l}{\footnotesize\textit{\textcolor{gray}{Other Closed-Source Models}}} \\
    Claude Sonnet 4.5
    & 76.04$^{\pm0.01}$ 
    & \textbf{68.61$^{\pm0.09}$ }
    & 1.18$^{\pm0.01}$
    & 78.73$^{\pm0.10}$ 
    & 67.39$^{\pm0.11}$\\
    Gemini 2.0 Flash 
    & 69.95$^{\pm0.12}$ 
    & 60.55$^{\pm0.08}$ 
    & 1.35$^{\pm0.00}$
    & \textbf{83.31$^{\pm0.19}$} 
    & 59.76$^{\pm0.18}$\\
    Gemini 2.5 Pro
    & 75.45$^{\pm1.13}$ 
    & 64.87$^{\pm0.39}$ 
    & 1.27$^{\pm0.01}$
    & 78.65$^{\pm1.17}$ 
    & 64.29$^{\pm1.58}$\\
    DeepSeek-R1 
    & 75.43$^{\pm0.34}$ 
    & 64.88$^{\pm0.33}$
    & 1.29$^{\pm0.01}$
    & 79.69$^{\pm0.91}$ 
    & 67.20$^{\pm1.03}$\\
    Qwen-plus 
    & \textbf{77.57$^{\pm0.44}$} 
    & 58.93$^{\pm0.22}$ 
    & 1.40$^{\pm0.00}$
    & 77.17$^{\pm0.61}$ 
    & 65.48$^{\pm0.87}$\\
    Qwen-max 
    & 77.40$^{\pm0.27}$ 
    & 58.86$^{\pm0.25}$ 
    & 1.40$^{\pm0.00}$
    & 77.17$^{\pm0.48}$ 
    & 65.21$^{\pm0.49}$\\
    \midrule
    \multicolumn{6}{l}{\footnotesize\textit{\textcolor{gray}{Open-Source Models}}} \\
    LLaMA 3.3 70B 
    & 76.39$^{\pm0.18}$ 
    & 67.57$^{\pm0.20}$ 
    & 1.24$^{\pm0.00}$
    & 79.56$^{\pm0.36}$ 
    & \textbf{69.84$^{\pm0.27}$}\\
    Qwen2.5-32B-instr. 
    & 77.17$^{\pm0.32}$ 
    & 58.96$^{\pm0.05}$ 
    & 1.40$^{\pm0.00}$
    & 76.88$^{\pm0.29}$ 
    & 64.71$^{\pm0.23}$\\
    Qwen2.5-7B-instr. 
    & 77.18$^{\pm0.73}$ 
    & 58.82$^{\pm0.20}$ 
    & 1.40$^{\pm0.00}$
    & 77.12$^{\pm0.80}$ 
    & 64.83$^{\pm1.21}$\\
    \midrule
    \multicolumn{6}{l}{\footnotesize\textit{\textcolor{gray}{Memory-Augmented Baselines}}} \\
    RAG\footnotemark[1]
    & 75.46$^{\pm0.47}$ 
    & 51.90$^{\pm0.39}$ 
    & 1.57$^{\pm0.05}$
    & 72.25$^{\pm0.57}$ 
    & 54.96$^{\pm0.87}$\\
    RAG-FC\footnotemark[1]
    & 77.56$^{\pm0.38}$ 
    & 59.97$^{\pm0.23}$ 
    & 1.39$^{\pm0.00}$
    & 76.80$^{\pm0.80}$ 
    & 65.65$^{\pm0.61}$\\
    Generative Agents\footnotemark[1]
    & 76.19$^{\pm0.36}$ 
    & 58.22$^{\pm0.43}$ 
    & 1.40$^{\pm0.01}$
    & 76.13$^{\pm2.45}$ 
    & 62.43$^{\pm1.74}$\\
    \midrule
    \multicolumn{6}{l}{\footnotesize\textit{\textcolor{gray}{Non-Learning Baselines}}} \\
    Global Majority 
    & 65.77$^{\pm0.00}$ 
    & 58.18$^{\pm0.00}$ 
    & 2.54$^{\pm0.00}$
    & 17.93$^{\pm0.00}$ 
    & 4.44$^{\pm0.00}$\\
    Random Guess    
    & 51.89$^{\pm3.96}$ 
    & 43.12$^{\pm0.78}$ 
    & 1.88$^{\pm0.05}$
    & 46.74$^{\pm3.28}$ 
    & 0.00$^{\pm5.80}$\\
    \bottomrule
    \end{tabular}
    \end{adjustbox}
    \caption{Average performance across all domains for \textit{Belief State Inference} and \textit{Belief Dynamics Update}.
    Results are reported as \textit{mean$\pm$std} over 5 runs. 
    Best-performing model (below human upper bound) per column is highlighted in \textbf{bold}. Real data only. }
    \label{tab:sota-model}
\end{table*}
\footnotetext[1]{RAG~\citep{lewis2020retrieval}, RAG-FC = RAG with Full Context~\citep{lewis2020retrieval}, Generative Agents~\citep{park2023thousandpeople}.}

\begin{table*}[ht]
\centering
\small
\setlength{\tabcolsep}{2.2pt}
\resizebox{\textwidth}{!}{%
\begin{tabular}{lcccccccccc}
\toprule
\multirow{3}{*}{\textbf{Model}} & \multicolumn{9}{c}{\textbf{Human Dataset (N=54)}} \\
\cmidrule(lr){2-10}
 & \multicolumn{3}{c}{\textbf{Belief State Inference}} & \multicolumn{6}{c}{\textbf{Belief Dynamics Update}} \\
 & \multicolumn{3}{c}{Acc. (\% $\uparrow$)} & \multicolumn{3}{c}{Acc. (\% $\uparrow$)} & \multicolumn{3}{c}{MAE ($\downarrow$)} \\
\cmidrule(lr){2-4}\cmidrule(lr){5-7}\cmidrule(lr){8-10}
 & Health & Surveil. & Zoning & Health & Surveil. & Zoning & Health & Surveil. & Zoning \\
\midrule
\rowcolor{gray!15}
\textbf{Human}\rule{0pt}{2.8ex}\rule[-1.2ex]{0pt}{0pt}
& \textbf{84.53} & \textbf{87.50} & \textbf{81.54} 
& \textbf{85.75} & \textbf{85.33} & \textbf{85.86} 
& \textbf{0.66} & \textbf{0.72} & \textbf{0.67} \\
\midrule
\multicolumn{10}{l}{\footnotesize\textit{\textcolor{gray}{OpenAI Models}}} \\
GPT-4o & 71.11$^{\pm0.77}$ & 75.25$^{\pm0.90}$ & 77.62$^{\pm0.35}$ & 61.57$^{\pm0.19}$ & 63.74$^{\pm0.58}$ & 64.04$^{\pm0.20}$ & 1.27$^{\pm0.01}$ & 1.35$^{\pm0.01}$ & 1.26$^{\pm0.01}$ \\
GPT5.1 Rea. High & 72.22$^{\pm1.01}$ & 77.54$^{\pm1.91}$ & 70.32$^{\pm0.46}$ & 63.43$^{\pm0.39}$ & 68.19$^{\pm1.13}$ & 69.78$^{\pm0.65}$ & 1.22$^{\pm0.02}$ & 1.16$^{\pm0.02}$ & 1.12$^{\pm0.01}$ \\
GPT-5-mini & 73.15$^{\pm2.36}$ & 76.89$^{\pm0.90}$ & 75.87$^{\pm0.43}$ & 59.92$^{\pm1.05}$ & 55.16$^{\pm1.09}$ & 59.56$^{\pm0.73}$ & 1.37$^{\pm0.02}$ & 1.54$^{\pm0.02}$ & 1.40$^{\pm0.01}$ \\
o3-mini & 69.26$^{\pm1.21}$ & 77.38$^{\pm1.70}$ & \textbf{78.73$^{\pm0.87}$} & 60.08$^{\pm0.28}$ & 62.53$^{\pm1.16}$ & 71.02$^{\pm0.79}$ & 1.31$^{\pm0.02}$ & 1.29$^{\pm0.03}$ & \textbf{1.07$^{\pm0.01}$} \\
\midrule
\multicolumn{10}{l}{\footnotesize\textit{\textcolor{gray}{Other Closed-Source Models}}} \\
Claude Sonnet 4.5 & 73.15$^{\pm0.01}$ & 81.15$^{\pm0.01}$ & 73.81$^{\pm0.01}$ & \textbf{65.21$^{\pm0.08}$} & \textbf{72.69$^{\pm0.22}$} & 67.93$^{\pm0.09}$ & \textbf{1.17$^{\pm0.01}$} & \textbf{1.13$^{\pm0.01}$} & 1.24$^{\pm0.01}$ \\
Gemini 2.0 Flash & 63.89$^{\pm0.00}$ & 68.20$^{\pm0.37}$ & 77.78$^{\pm0.01}$ & 56.23$^{\pm0.12}$ & 62.25$^{\pm0.37}$ & 63.16$^{\pm0.10}$ & 1.39$^{\pm0.01}$ & 1.33$^{\pm0.01}$ & 1.33$^{\pm0.01}$ \\
Gemini 2.5 Pro & 73.70$^{\pm1.99}$ & 78.03$^{\pm1.31}$ & 74.60$^{\pm1.00}$ & 62.12$^{\pm0.36}$ & 62.53$^{\pm0.57}$ & 69.96$^{\pm0.53}$ & 1.29$^{\pm0.01}$ & 1.31$^{\pm0.01}$ & 1.21$^{\pm0.01}$ \\
DeepSeek-R1-0528 & 76.11$^{\pm0.41}$ & 75.41$^{\pm1.16}$ & 74.76$^{\pm0.66}$ & 61.44$^{\pm0.58}$ & 68.30$^{\pm0.63}$ & 64.91$^{\pm0.41}$ & 1.33$^{\pm0.00}$ & 1.27$^{\pm0.02}$ & 1.26$^{\pm0.01}$ \\
Qwen-plus\_2025-07-28 & \textbf{78.70$^{\pm0.65}$} & 76.39$^{\pm1.22}$ & 77.62$^{\pm0.66}$ & 56.74$^{\pm0.59}$ & 57.86$^{\pm0.15}$ & 62.18$^{\pm0.18}$ & 1.44$^{\pm0.01}$ & 1.43$^{\pm0.01}$ & 1.34$^{\pm0.00}$ \\
Qwen-max\_2024-10-15 & 78.52$^{\pm0.77}$ & 76.39$^{\pm0.37}$ & 77.30$^{\pm0.43}$ & 56.78$^{\pm0.64}$ & 57.69$^{\pm0.27}$ & 62.11$^{\pm0.35}$ & 1.43$^{\pm0.01}$ & 1.44$^{\pm0.01}$ & 1.34$^{\pm0.00}$ \\
\midrule
\multicolumn{10}{l}{\footnotesize\textit{\textcolor{gray}{Open-Source Models}}} \\
LLaMA\_3.3\_70B & 70.74$^{\pm0.51}$ & \textbf{80.66$^{\pm0.45}$} & 77.78$^{\pm0.00}$ & 64.28$^{\pm0.38}$ & 65.33$^{\pm0.23}$ & \textbf{73.09$^{\pm0.31}$} & 1.31$^{\pm0.01}$ & 1.29$^{\pm0.01}$ & 1.14$^{\pm0.01}$ \\
Qwen2.5-32B-instr. & 77.96$^{\pm0.41}$ & 76.56$^{\pm0.45}$ & 76.98$^{\pm0.79}$ & 56.82$^{\pm0.41}$ & 57.80$^{\pm0.25}$ & 62.25$^{\pm0.16}$ & 1.43$^{\pm0.00}$ & 1.43$^{\pm0.01}$ & 1.33$^{\pm0.01}$ \\
Qwen2.5-7B-instr & 78.33$^{\pm0.51}$ & 75.74$^{\pm2.06}$ & 77.46$^{\pm0.43}$ & 56.82$^{\pm0.41}$ & 57.64$^{\pm0.45}$ & 62.00$^{\pm0.34}$ & 1.43$^{\pm0.01}$ & 1.43$^{\pm0.01}$ & 1.34$^{\pm0.01}$ \\
\midrule
\multicolumn{10}{l}{\footnotesize\textit{\textcolor{gray}{Memory-Augmented Baselines}}} \\
RAG\footnotemark[1] & 71.67$^{\pm0.51}$ & 78.36$^{\pm1.37}$ & 76.35$^{\pm0.35}$ & 50.64$^{\pm0.58}$ & 53.68$^{\pm0.60}$ & 51.38$^{\pm0.24}$ & 1.60$^{\pm0.01}$ & 1.50$^{\pm0.00}$ & 1.60$^{\pm0.00}$ \\
RAG-FC\footnotemark[1] & 77.22$^{\pm0.83}$ & 76.72$^{\pm0.93}$ & \textbf{78.73$^{\pm0.87}$} & 58.73$^{\pm0.55}$ & 59.34$^{\pm0.34}$ & 61.85$^{\pm0.15}$ & 1.43$^{\pm0.01}$ & 1.41$^{\pm0.01}$ & 1.33$^{\pm0.01}$ \\
Generative Agents\footnotemark[2] & 76.11$^{\pm0.41}$ & 77.21$^{\pm0.37}$ & 75.24$^{\pm0.66}$ & 57.67$^{\pm0.76}$ & 55.66$^{\pm0.69}$ & 61.35$^{\pm0.38}$ & 1.43$^{\pm0.01}$ & 1.44$^{\pm0.01}$ & 1.34$^{\pm0.00}$ \\
\bottomrule
\end{tabular}%
}
\caption{Results on \textbf{Human} dataset for \textit{belief state inference} and \textit{belief dynamics update} tasks across three policy topics (Health, Surveillance, Zoning). Values are mean$^{\pm\text{std}}$ over 5 runs. For \textit{belief dynamics update}, both Accuracy and MAE are reported separately. Best non-human results per column are in \textbf{bold}.}
\label{tab:human_results}
\end{table*}


\begin{table*}[ht]
\centering
\small
\setlength{\tabcolsep}{2.2pt}
\resizebox{\textwidth}{!}{%
\begin{tabular}{lcccccccccc}
\toprule
\multirow{2}{*}{\textbf{Model}} & \multicolumn{9}{c}{\textbf{Belief Dynamics Update}} \\
\cmidrule(lr){2-10}
 & \multicolumn{3}{c}{Change Detection Acc. (\%$\uparrow$)} & \multicolumn{3}{c}{Direction Inference Acc. (\%$\uparrow$)} & \multicolumn{3}{c}{Dir. Acc. (\%$\uparrow$)} \\
\cmidrule(lr){2-4}\cmidrule(lr){5-7}\cmidrule(lr){8-10}
 & Health & Surveil. & Zoning & Health & Surveil. & Zoning & Health & Surveil. & Zoning \\
\midrule
\rowcolor{gray!15}
\textbf{Human}\rule{0pt}{2.8ex}\rule[-1.2ex]{0pt}{0pt}
& \textbf{80.00} & \textbf{90.00} & \textbf{80.00} 
& \textbf{96.67} & \textbf{88.33} & \textbf{90.00} 
& \textbf{91.67} & \textbf{88.83} & \textbf{87.00} \\
\midrule
\multicolumn{10}{l}{\footnotesize\textit{\textcolor{gray}{OpenAI Models}}} \\
GPT-4o & 49.36$^{\pm1.59}$ & 65.83$^{\pm3.12}$ & 57.14$^{\pm3.01}$ & 88.89$^{\pm0.00}$ & \textbf{91.49$^{\pm0.53}$} & 98.33$^{\pm3.33}$ & 77.03$^{\pm0.48}$ & \textbf{83.79$^{\pm1.31}$} & 85.98$^{\pm1.71}$ \\
GPT5.1 Rea. High & 54.47$^{\pm6.67}$ & 67.50$^{\pm5.53}$ & 59.05$^{\pm9.80}$ & 83.06$^{\pm6.60}$ & 88.26$^{\pm3.57}$ & 98.00$^{\pm4.00}$ & 74.48$^{\pm6.22}$ & 82.03$^{\pm3.37}$ & 86.31$^{\pm5.11}$ \\
GPT-5-mini & 46.81$^{\pm4.66}$ & 65.83$^{\pm6.12}$ & 60.95$^{\pm3.56}$ & 82.95$^{\pm2.56}$ & 90.67$^{\pm5.33}$ & 82.05$^{\pm9.48}$ & 72.11$^{\pm2.84}$ & 83.22$^{\pm3.06}$ & 75.72$^{\pm7.17}$ \\
o3-mini & 50.64$^{\pm4.54}$ & 55.83$^{\pm4.25}$ & 59.05$^{\pm5.71}$ & 76.43$^{\pm4.84}$ & 79.58$^{\pm3.64}$ & 78.59$^{\pm12.06}$ & 68.69$^{\pm3.33}$ & 72.45$^{\pm3.10}$ & 72.72$^{\pm8.67}$ \\
\midrule
\multicolumn{10}{l}{\footnotesize\textit{\textcolor{gray}{Other Closed-Source Models}}} \\
Claude Sonnet 4.5 & 45.96$^{\pm1.04}$ & 62.50$^{\pm0.01}$ & 61.90$^{\pm0.01}$ & 87.50$^{\pm0.01}$ & 84.62$^{\pm0.01}$ & 92.31$^{\pm0.01}$ & 75.04$^{\pm0.31}$ & 
77.98$^{\pm0.01}$ & 83.19$^{\pm0.01}$ \\
Gemini 2.0 Flash & 38.30$^{\pm0.00}$ & \textbf{70.83$^{\pm0.00}$} & 62.86$^{\pm1.90}$ & \textbf{100.00$^{\pm0.00}$} & 83.33$^{\pm0.00}$ & \textbf{100.00$^{\pm0.00}$} & \textbf{81.49$^{\pm0.00}$} & 79.58$^{\pm0.00}$ & 88.86$^{\pm0.57}$ \\
Gemini 2.5 Pro & 
49.36$^{\pm1.59}$ & 54.17$^{\pm5.27}$ & \textbf{70.48$^{\pm3.56}$} & 76.19$^{\pm5.83}$ & 86.32$^{\pm3.85}$ & \textbf{100.00$^{\pm0.01}$} & 68.14$^{\pm3.73}$ & 76.67$^{\pm2.43}$ & 
91.14$^{\pm1.07}$ \\
DeepSeek-R1-0528 & 54.47$^{\pm2.17}$ & 64.67$^{\pm5.20}$ & 52.38$^{\pm3.01}$ & 83.62$^{\pm2.10}$ & 84.42$^{\pm3.75}$ & \textbf{100.00$^{\pm0.00}$} & 74.87$^{\pm1.86}$ & 78.50$^{\pm1.50}$ & 85.71$^{\pm0.90}$ \\
Qwen-plus\_2025-07-28 & 34.04$^{\pm2.33}$ & 63.01$^{\pm1.94}$ & 66.67$^{\pm0.00}$ & 85.24$^{\pm0.95}$ & 83.03$^{\pm0.61}$ & 92.31$^{\pm0.00}$ & 69.88$^{\pm0.97}$ & 77.02$^{\pm0.89}$ & 84.62$^{\pm0.00}$ \\
Qwen-max\_2024-10-15 & 34.89$^{\pm1.04}$ & 63.33$^{\pm1.67}$ & 66.67$^{\pm0.00}$ & 85.71$^{\pm0.00}$ & 82.05$^{\pm2.56}$ & 92.31$^{\pm0.00}$ & 70.47$^{\pm0.31}$ & 76.44$^{\pm1.29}$ & 84.62$^{\pm0.00}$ \\
\midrule
\multicolumn{10}{l}{\footnotesize\textit{\textcolor{gray}{Open-Source Models}}} \\
LLaMA\_3.3\_70B & \textbf{68.09$^{\pm3.01}$} & 54.17$^{\pm0.00}$ & 53.33$^{\pm1.90}$ & 80.00$^{\pm0.00}$ & 85.71$^{\pm0.00}$ & \textbf{100.00$^{\pm0.00}$} & 76.43$^{\pm0.90}$ & 76.25$^{\pm0.00}$ & 86.00$^{\pm0.57}$ \\
Qwen2.5-32B-instr. & 33.19$^{\pm1.04}$ & 62.50$^{\pm0.00}$ & \textbf{66.67$^{\pm3.01}$} & 84.29$^{\pm1.17}$ & 83.33$^{\pm0.00}$ & 92.29$^{\pm0.38}$ & 68.96$^{\pm1.05}$ & 77.08$^{\pm0.00}$ & 84.60$^{\pm1.17}$ \\
Qwen2.5-7B-instr & 33.62$^{\pm1.59}$ & 62.50$^{\pm0.00}$ & 65.63$^{\pm1.40}$ & 84.29$^{\pm1.17}$ & 83.33$^{\pm0.00}$ & 93.57$^{\pm3.26}$ & 69.09$^{\pm1.03}$ & 77.08$^{\pm0.00}$ & 85.19$^{\pm2.30}$ \\
\midrule
\multicolumn{10}{l}{\footnotesize\textit{\textcolor{gray}{Memory-Augmented Baselines}}} \\
RAG\footnotemark[1] & 42.55$^{\pm1.90}$ & 55.83$^{\pm5.00}$ & 66.67$^{\pm0.00}$ & 66.67$^{\pm0.00}$ & 86.51$^{\pm3.85}$ & 85.71$^{\pm0.00}$ & 59.43$^{\pm0.57}$ & 77.30$^{\pm1.94}$ & 80.00$^{\pm0.00}$ \\
RAG-FC\footnotemark[1] & 45.11$^{\pm2.08}$ & 62.43$^{\pm6.00}$ & 60.95$^{\pm1.90}$ & 80.00$^{\pm0.00}$ & 90.91$^{\pm0.00}$ & 86.03$^{\pm2.82}$ & 69.53$^{\pm0.63}$ & 82.36$^{\pm1.80}$ & 78.50$^{\pm1.40}$ \\
Generative Agents\footnotemark[2] & 36.17$^{\pm1.90}$ & 61.06$^{\pm2.27}$ & 60.95$^{\pm1.90}$ & 81.00$^{\pm9.70}$ & 83.64$^{\pm3.64}$ & 93.85$^{\pm3.08}$ & 67.55$^{\pm6.41}$ & 76.86$^{\pm2.20}$ & 83.98$^{\pm1.58}$ \\
\bottomrule
\end{tabular}%
}
\caption{Directional accuracy results on \textbf{Human} dataset for \textit{belief dynamics update} task across three policy topics (Health, Surveillance, Zoning). Values are mean$^{\pm\text{std}}$ over 5 runs. For each topic, we report Change Detection Accuracy, Direction Inference Accuracy, and Directional Accuracy (Dir. Acc). Best results per column are in \textbf{bold}.}
\label{tab:full_human_results}
\end{table*}


\begin{table*}[h!]
\centering
\begin{adjustbox}{max width=\textwidth}
{\scriptsize
\setlength{\tabcolsep}{5pt}
\renewcommand{\arraystretch}{1.25}
\begin{tabular}{llcccc}
\toprule
\textbf{Metric} & \textbf{Domain Transfer} 
& \multicolumn{2}{c}{GPT-4o} 
& \multicolumn{2}{c}{Qwen2.5-32B-instr.} \\
\cmidrule(lr){3-4}\cmidrule(lr){5-6}
& & Cross-Domain & In-Domain & Cross-Domain & In-Domain \\
\midrule
\multicolumn{6}{l}{\textbf{Belief State Inference (Accuracy \% $\uparrow$)}} \\
& Health $\rightarrow$ Surv & 58.52$\pm$0.77 & 71.11$\pm$0.77 & 53.52$\pm$1.52 & 77.96$\pm$0.41 \\
& Surv $\rightarrow$ Zone & 65.41$\pm$1.58 & 75.25$\pm$0.90 & 58.69$\pm$1.10 & 76.56$\pm$0.45 \\
& Zone $\rightarrow$ Health & 51.75$\pm$1.03 & 77.62$\pm$0.35 & 55.08$\pm$1.20 & 76.98$\pm$0.79 \\
& \textit{Average} & \textit{58.56$\pm$0.82} & \textit{74.66$\pm$0.24} & \textit{55.76$\pm$1.09} & \textit{77.17$\pm$0.32} \\
\midrule
\multicolumn{6}{l}{\textbf{Belief Dynamics Update (Accuracy \% $\uparrow$)}} \\
& Health $\rightarrow$ Surv & 49.49$\pm$0.87 & 61.57$\pm$0.19 & 44.41$\pm$0.98 & 56.82$\pm$0.41 \\
& Surv $\rightarrow$ Zone & 42.47$\pm$0.50 & 63.74$\pm$0.58 & 24.84$\pm$0.60 & 57.80$\pm$0.25 \\
& Zone $\rightarrow$ Health & 41.96$\pm$0.30 & 64.04$\pm$0.20 & 30.15$\pm$0.35 & 62.25$\pm$0.16 \\
& \textit{Average} & \textit{44.64$\pm$0.36} & \textit{63.11$\pm$0.14} & \textit{33.13$\pm$0.48} & \textit{58.96$\pm$0.05} \\
\midrule
\multicolumn{6}{l}{\textbf{Belief Dynamics Update (MAE $\downarrow$)}} \\
& Health $\rightarrow$ Surv & 1.55$\pm$0.01 & 1.27$\pm$0.01 & 1.69$\pm$0.01 & 1.43$\pm$0.00 \\
& Surv $\rightarrow$ Zone & 1.60$\pm$0.00 & 1.35$\pm$0.01 & 1.94$\pm$0.01 & 1.43$\pm$0.01 \\
& Zone $\rightarrow$ Health & 1.71$\pm$0.00 & 1.26$\pm$0.01 & 2.05$\pm$0.01 & 1.33$\pm$0.01 \\
& \textit{Average} & \textit{1.62$\pm$0.00} & \textit{1.29$\pm$0.00} & \textit{1.89$\pm$0.01} & \textit{1.40$\pm$0.00} \\
\midrule
\multicolumn{6}{l}{\textbf{ATI (\% $\uparrow$)}} \\
& Health $\rightarrow$ Surv & 23.96$\pm$3.83 & 59.43$\pm$1.36 & 10.04$\pm$2.71 & 60.56$\pm$1.19 \\
& Surv $\rightarrow$ Zone & -19.05$\pm$2.32 & 66.01$\pm$1.49 & 12.54$\pm$1.80 & 60.25$\pm$0.65 \\
& Zone $\rightarrow$ Health & 24.72$\pm$1.67 & 78.92$\pm$1.48 & 21.78$\pm$2.00 & 75.41$\pm$0.95 \\
& \textit{Average} & \textit{9.64$\pm$1.66} & \textit{67.29$\pm$0.74} & \textit{14.36$\pm$1.49} & \textit{64.71$\pm$0.23} \\
\bottomrule
\end{tabular}
}
\end{adjustbox}
\caption{Cross-domain swap test: models are trained with QA context from one domain and evaluated on another domain for the \emph{same participant}. 
We report performance for \textit{Healthcare $\rightarrow$ Surveillance}, \textit{Surveillance $\rightarrow$ Zoning}, \textit{Zoning $\rightarrow$ Healthcare}, and their average. Reported as \textit{mean $\pm$ std} over 5 runs.}

\end{table*}

    \begin{table*}[ht]
    \centering
    \begin{adjustbox}{max totalsize={\textwidth}{\textheight},keepaspectratio}
    {\scriptsize
    \begin{tabular}{lccccc}
    \toprule
    \multirow{2}{*}{\textbf{Method}} 
    & \textbf{Belief State Inference} 
    & \multicolumn{3}{c}{\textbf{Belief Dynamics Update}} 
    & \multirow{2}{*}{\textbf{ATI (\% $\uparrow$)}} \\
    \cmidrule(lr){2-2}\cmidrule(lr){3-5}
    & Accuracy (\% $\uparrow$) & Accuracy (\% $\uparrow$) & MAE ($\downarrow$) & Dir. Acc. (\% $\uparrow$) & \\
    \midrule
    GPT-4o (No-Context) & 58.49$^{\pm0.43}$ & 39.83$^{\pm0.45}$ & 1.70$^{\pm0.01}$ & 75.59$^{\pm0.81}$ & 26.98$^{\pm1.01}$ \\
    GPT-4o (Full-Context) & 74.66$^{\pm0.24}$ & 63.11$^{\pm0.14}$ & 1.29$^{\pm0.00}$ & 82.27$^{\pm1.02}$ & 67.29$^{\pm0.74}$ \\
    \midrule
    Qwen2.5-32B-instr. (No-Context) & 50.92$^{\pm0.55}$ & 32.12$^{\pm0.24}$ & 1.88$^{\pm0.01}$ & 69.00$^{\pm1.07}$ & 6.88$^{\pm0.81}$ \\
    Qwen2.5-32B-instr. (Full-Context) & 77.17$^{\pm0.32}$ & 58.96$^{\pm0.05}$ & 1.40$^{\pm0.00}$ & 76.88$^{\pm0.29}$ & 64.71$^{\pm0.23}$ \\
    \bottomrule
    \end{tabular}
    }
    \end{adjustbox}
    \caption{Comparison of \emph{No-Context} (population prior) and \emph{Full-Context} (with individual transcripts) settings. Reported as \textit{mean $\pm$ std} over 5 runs.}
    \label{tab:global-prior}
    \end{table*}
\begin{table*}[h!]
\centering
\begin{adjustbox}{max width=\textwidth}
{\scriptsize
\setlength{\tabcolsep}{5pt}
\renewcommand{\arraystretch}{1.25}
\begin{tabular}{lcccc}
\toprule
\textbf{Metric} 
& \multicolumn{2}{c}{GPT-4o} 
& \multicolumn{2}{c}{Qwen2.5-32B-instr.} \\
\cmidrule(lr){2-3}\cmidrule(lr){4-5}
& Cross-Domain & In-Domain & Cross-Domain & In-Domain \\
\midrule
Belief State Inference (Accuracy \% $\uparrow$) & 58.56$\pm$0.82 & 74.66$\pm$0.24 & 55.76$\pm$1.09 & 77.17$\pm$0.32 \\
Belief Dynamics Update (Accuracy \% $\uparrow$) & 44.64$\pm$0.36 & 63.11$\pm$0.14 & 33.13$\pm$0.48 & 58.96$\pm$0.05 \\
Belief Dynamics Update (MAE $\downarrow$) & 1.62$\pm$0.00 & 1.29$\pm$0.00 & 1.89$\pm$0.01 & 1.40$\pm$0.00 \\
ATI (\% $\uparrow$) & 9.64$\pm$1.66 & 67.29$\pm$0.74 & 14.36$\pm$1.49 & 64.71$\pm$0.23 \\
\bottomrule
\end{tabular}
}
\end{adjustbox}
\caption{Cross-domain swap test: models are trained with QA context from one domain and evaluated on another domain for the \emph{same participant}. 
Reported as \textit{mean $\pm$ std} over 5 runs averaged across all domain transfer pairs.}
\label{tab:appendix-cross-topic}
\end{table*}

\begin{table*}[h!]
\centering
\begin{adjustbox}{max width=\textwidth}
{\scriptsize
\setlength{\tabcolsep}{4pt}
\renewcommand{\arraystretch}{1.25}
\begin{tabular}{llcccccccccc}
\toprule
\textbf{Metric} & \textbf{Domain} 
& \multicolumn{3}{c}{GPT-4o} 
& \multicolumn{3}{c}{Gemini 2.0 Flash}
& \multicolumn{3}{c}{Qwen2.5-32B-instr.} \\
\cmidrule(lr){3-5}\cmidrule(lr){6-8}\cmidrule(lr){9-11}
& & 5 QAs & 10 QAs & 20+ QAs & 5 QAs & 10 QAs & 20+ QAs & 5 QAs & 10 QAs & 20+ QAs \\
\midrule
\multicolumn{11}{l}{\textbf{Belief State Inference (Accuracy \% $\uparrow$)}} \\
\midrule
& Health & 68.89$\pm$0.51 & 67.59$\pm$1.73 & 71.11$\pm$0.77 & 61.30$\pm$0.41 & 60.19$\pm$0.65 & 63.89$\pm$0.00 & 68.33$\pm$1.21 & 75.37$\pm$4.47 & \textbf{77.96$\pm$0.41} \\
& Surv. & 73.93$\pm$1.07 & 72.30$\pm$0.37 & 75.25$\pm$0.90 & 65.25$\pm$0.45 & 64.75$\pm$0.00 & 68.20$\pm$0.37 & 73.11$\pm$2.68 & 75.90$\pm$1.80 & \textbf{76.56$\pm$0.45} \\
& Zone & 72.22$\pm$0.00 & 72.06$\pm$0.66 & 77.62$\pm$0.35 & 65.08$\pm$0.00 & 71.27$\pm$0.35 & \textbf{77.78$\pm$0.00} & 65.08$\pm$0.56 & 67.78$\pm$1.06 & 76.98$\pm$0.79 \\
& Avg & 71.68$\pm$0.47 & 70.65$\pm$0.49 & 74.66$\pm$0.24 & 63.87$\pm$0.13 & 65.40$\pm$0.25 & 69.95$\pm$0.12 & 68.84$\pm$0.73 & 73.02$\pm$1.66 & \textbf{77.17$\pm$0.32} \\
\midrule
\multicolumn{11}{l}{\textbf{Belief Dynamics Update (Accuracy \% $\uparrow$)}} \\
\midrule
& Health & 61.48$\pm$0.35 & \textbf{62.29$\pm$0.33} & 61.57$\pm$0.19 & 59.70$\pm$0.09 & 58.22$\pm$0.28 & 56.23$\pm$0.12 & 59.45$\pm$0.55 & 58.52$\pm$0.74 & 56.82$\pm$0.41 \\
& Surv. & \textbf{67.80$\pm$0.45} & 64.56$\pm$0.39 & 63.74$\pm$0.58 & 65.71$\pm$0.30 & 62.58$\pm$0.23 & 62.25$\pm$0.37 & 62.53$\pm$0.57 & 60.66$\pm$0.60 & 57.80$\pm$0.25 \\
& Zone & 63.27$\pm$0.36 & \textbf{65.82$\pm$0.39} & 64.04$\pm$0.20 & 62.25$\pm$0.21 & 65.20$\pm$0.16 & 63.16$\pm$0.10 & 63.38$\pm$0.69 & 63.02$\pm$0.47 & 62.25$\pm$0.16 \\
& Avg & 64.19$\pm$0.22 & \textbf{64.22$\pm$0.20} & 63.11$\pm$0.14 & 62.56$\pm$0.18 & 62.00$\pm$0.13 & 60.55$\pm$0.08 & 61.79$\pm$0.46 & 60.73$\pm$0.46 & 58.96$\pm$0.05 \\
\midrule
\multicolumn{11}{l}{\textbf{Belief Dynamics Update (MAE $\downarrow$)}} \\
\midrule
& Health & \textbf{1.26$\pm$0.01} & 1.28$\pm$0.00 & 1.27$\pm$0.01 & 1.36$\pm$0.00 & 1.37$\pm$0.00 & 1.39$\pm$0.00 & 1.36$\pm$0.00 & 1.41$\pm$0.01 & 1.43$\pm$0.00 \\
& Surv. & \textbf{1.19$\pm$0.01} & 1.26$\pm$0.01 & 1.35$\pm$0.01 & 1.23$\pm$0.00 & 1.30$\pm$0.01 & 1.33$\pm$0.00 & 1.31$\pm$0.01 & 1.36$\pm$0.01 & 1.43$\pm$0.01 \\
& Zone & 1.24$\pm$0.00 & \textbf{1.21$\pm$0.01} & 1.26$\pm$0.01 & 1.34$\pm$0.00 & 1.26$\pm$0.00 & 1.33$\pm$0.00 & 1.27$\pm$0.01 & 1.27$\pm$0.01 & 1.33$\pm$0.01 \\
& Avg & \textbf{1.23$\pm$0.00} & 1.25$\pm$0.00 & 1.29$\pm$0.00 & 1.31$\pm$0.00 & 1.31$\pm$0.00 & 1.35$\pm$0.00 & 1.31$\pm$0.00 & 1.35$\pm$0.00 & 1.40$\pm$0.00 \\
\midrule
\multicolumn{11}{l}{\textbf{ATI (\% $\uparrow$)}} \\
\midrule
& Health & 55.33$\pm$1.05 & 55.54$\pm$2.52 & 59.43$\pm$1.36 & 31.13$\pm$0.84 & 34.19$\pm$0.78 & 50.03$\pm$0.06 & 47.57$\pm$1.16 & 58.18$\pm$5.94 & \textbf{60.56$\pm$1.19} \\
& Surv. & \textbf{69.38$\pm$1.54} & 60.07$\pm$0.27 & 66.01$\pm$1.49 & 50.19$\pm$0.21 & 46.56$\pm$1.71 & 52.20$\pm$0.51 & 57.95$\pm$5.12 & 59.36$\pm$2.68 & 60.25$\pm$0.65 \\
& Zone & 70.60$\pm$0.49 & 67.45$\pm$1.02 & 78.92$\pm$1.48 & 62.63$\pm$0.11 & 67.57$\pm$0.57 & \textbf{80.62$\pm$0.57} & 62.15$\pm$0.94 & 65.47$\pm$2.27 & 75.41$\pm$0.95 \\
& Avg & 64.46$\pm$0.81 & 60.46$\pm$0.98 & \textbf{67.29$\pm$0.74} & 46.87$\pm$0.35 & 48.24$\pm$0.76 & 59.76$\pm$0.18 & 55.28$\pm$1.46 & 60.58$\pm$2.58 & 64.71$\pm$0.23 \\
\bottomrule
\end{tabular}
}
\end{adjustbox}
\caption{Question masking / length scaling for both Belief State Inference and Belief Dynamics Update. Reported as \textit{mean $\pm$ std} over 5 runs. Best results per row are highlighted in \textbf{bold}.}
\label{tab:appendix-masking-combined}
\end{table*}

  \begin{table*}[ht]
    \centering
    \setlength{\tabcolsep}{3pt}       
    \renewcommand{\arraystretch}{0.92} 
   
    \begin{adjustbox}{max width=\textwidth}
    {\scriptsize
    \begin{tabular}{lccccc}
    \toprule
    \multirow{2}{*}{\textbf{Method}} 
    & \multicolumn{1}{c}{\textbf{Belief State Inference}} 
    & \multicolumn{3}{c}{\textbf{Belief Dynamics Update}} 
    & \multirow{2}{*}{\textbf{ATI (\% $\uparrow$)}} \\
    \cmidrule(lr){2-2}\cmidrule(lr){3-5}
    & Accuracy (\% $\uparrow$) 
    & Accuracy (\% $\uparrow$) & MAE ($\downarrow$) & Dir. Acc. (\% $\uparrow$) & \\
    \midrule
    GPT\mbox{-}4o (Cross-Person)       & 54.45$^{\pm2.30}$ & 39.30$^{\pm0.22}$ & 1.93$^{\pm0.01}$ & 63.56$^{\pm0.76}$ & 10.29$^{\pm3.88}$ \\
    GPT\mbox{-}4o (Same-Person)       & 74.66$^{\pm0.24}$ & 63.11$^{\pm0.14}$ & 1.29$^{\pm0.00}$ & 82.27$^{\pm1.02}$ & 67.29$^{\pm0.74}$ \\
    \midrule
    Qwen2.5\mbox{-}32B-instr. (Cross-Person)  & 60.44$^{\pm0.47}$ & 38.16$^{\pm0.44}$ & 2.02$^{\pm0.01}$ & 69.17$^{\pm1.06}$ & 22.15$^{\pm1.24}$ \\
    Qwen2.5\mbox{-}32B-instr. (Same-Person)   & 77.17$^{\pm0.32}$ & 58.96$^{\pm0.05}$ & 1.40$^{\pm0.00}$ & 76.88$^{\pm0.29}$ & 64.71$^{\pm0.23}$\\
    \bottomrule
    \end{tabular}
    }
    \end{adjustbox}
    \caption{Cross-person swap test: QA context from one participant is used to predict another participant's responses (Cross-Person) versus the same participant (Same-Person). Results are reported as \textit{mean $\pm$ std} over 5 runs, averaged across domains.}
    \label{tab:cross-person}
    \end{table*}
\begin{table*}[ht]
\centering
\begin{adjustbox}{max width=\textwidth}
{\scriptsize
\setlength{\tabcolsep}{4pt}
\renewcommand{\arraystretch}{1.25}
\begin{tabular}{llcccc}
\toprule
\textbf{Metric} & \textbf{Domain} 
& \multicolumn{2}{c}{GPT-4o} 
& \multicolumn{2}{c}{Qwen2.5-32B-instr.} \\
\cmidrule(lr){3-4}\cmidrule(lr){5-6}
& & No-Context & Full-Context & No-Context & Full-Context \\
\midrule
\multicolumn{6}{l}{\textbf{Belief State Inference (Accuracy \% $\uparrow$)}} \\
\midrule
& Health & 55.93$\pm$1.40 & 71.11$\pm$0.77 & 45.19$\pm$1.66 & 77.96$\pm$0.41 \\
& Surv. & 66.07$\pm$1.37 & 75.25$\pm$0.90 & 59.02$\pm$1.16 & 76.56$\pm$0.45 \\
& Zone & 53.49$\pm$1.99 & 77.62$\pm$0.35 & 48.57$\pm$1.30 & 76.98$\pm$0.79 \\
& Avg & 58.49$\pm$0.43 & 74.66$\pm$0.24 & 50.92$\pm$0.55 & 77.17$\pm$0.32 \\
\midrule
\multicolumn{6}{l}{\textbf{Belief Dynamics Update (Accuracy \% $\uparrow$)}} \\
\midrule
& Health & 34.62$\pm$0.68 & 61.57$\pm$0.19 & 34.58$\pm$0.55 & 56.82$\pm$0.41 \\
& Surv. & 48.24$\pm$0.31 & 63.74$\pm$0.58 & 33.90$\pm$0.60 & 57.80$\pm$0.25 \\
& Zone & 36.62$\pm$0.52 & 64.04$\pm$0.20 & 27.89$\pm$0.38 & 62.25$\pm$0.16 \\
& Avg & 39.83$\pm$0.45 & 63.11$\pm$0.14 & 32.12$\pm$0.24 & 58.96$\pm$0.05 \\
\midrule
\multicolumn{6}{l}{\textbf{Belief Dynamics Update (MAE $\downarrow$)}} \\
\midrule
& Health & 1.77$\pm$0.01 & 1.27$\pm$0.01 & 1.89$\pm$0.01 & 1.43$\pm$0.00 \\
& Surv. & 1.51$\pm$0.01 & 1.35$\pm$0.01 & 1.75$\pm$0.01 & 1.43$\pm$0.01 \\
& Zone & 1.84$\pm$0.01 & 1.26$\pm$0.01 & 1.99$\pm$0.01 & 1.33$\pm$0.01 \\
& Avg & 1.70$\pm$0.01 & 1.29$\pm$0.00 & 1.88$\pm$0.01 & 1.40$\pm$0.00 \\
\midrule
\multicolumn{6}{l}{\textbf{ATI (\% $\uparrow$)}} \\
\midrule
& Health & 19.76$\pm$2.31 & 59.43$\pm$1.36 & -7.58$\pm$2.70 & 60.56$\pm$1.19 \\
& Surv. & 35.66$\pm$2.50 & 66.01$\pm$1.49 & 15.31$\pm$1.64 & 60.25$\pm$0.65 \\
& Zone & 26.20$\pm$3.73 & 78.92$\pm$1.48 & 14.92$\pm$2.15 & 75.41$\pm$0.95 \\
& Avg & 26.98$\pm$1.01 & 67.29$\pm$0.74 & 6.88$\pm$0.81 & 64.71$\pm$0.23 \\
\bottomrule
\end{tabular}
}
\end{adjustbox}
\caption{Comparison of \emph{No-Context} (population prior) and \emph{Full-Context} (with individual transcripts) settings. Reported as \textit{mean $\pm$ std} over 5 runs.}
\label{tab:appendix-global-prior}
\end{table*}

 \begin{table*}[h!]
    \centering
    \begin{adjustbox}{max width=\textwidth}
    {\footnotesize
    \setlength{\tabcolsep}{8pt}
    \renewcommand{\arraystretch}{1.15}
    \begin{tabular}{llcccc}
    \toprule
    \textbf{Task} & \textbf{Model} 
    & Health $\rightarrow$ Surv & Surv $\rightarrow$ Zone & Zone $\rightarrow$ Health & Avg \\
    \midrule
    \multirow{2}{*}{Belief State Inference (Accuracy \% $\uparrow$)} 
    & Qwen2.5-32B-instr. & 
    53.52$^{\pm1.52}$ & 
    58.69$^{\pm1.10}$ & 
    55.08$^{\pm1.20}$ & 
    55.76$^{\pm1.09}$ \\
    & GPT-4o & 58.52$^{\pm0.77}$ & 65.41$^{\pm1.58}$ & 51.75$^{\pm1.03}$ & 58.56$^{\pm0.82}$ \\
    \midrule
    \multirow{2}{*}{Belief Dynamics Update (Accuracy \% $\uparrow$)} 
    & Qwen2.5-32B-instr. & 44.41$^{\pm0.98}$ & 24.84$^{\pm0.60}$ & 30.15$^{\pm0.35}$ & 33.13$^{\pm0.48}$ \\
    & GPT-4o & 49.49$^{\pm0.87}$ & 42.47$^{\pm0.50}$ & 41.96$^{\pm0.30}$ & 44.64$^{\pm0.36}$ \\
    \midrule
    \multirow{2}{*}{Belief Dynamics Update (MAE $\downarrow$)} 
    & Qwen2.5-32B-instr. & 1.69$^{\pm0.01}$ & 1.94$^{\pm0.01}$ & 2.05$^{\pm0.01}$ & 1.89$^{\pm0.01}$ \\
    & GPT-4o & 1.55$^{\pm0.01}$ & 1.60$^{\pm0.00}$ & 1.71$^{\pm0.00}$ & 1.62$^{\pm0.00}$ \\
    \bottomrule
    \end{tabular}
    }
    \end{adjustbox}
    \caption{Cross-domain swap test: models are trained with QA context from one domain and evaluated on another domain for the \emph{same participant}. 
    We report performance for \textit{Healthcare $\rightarrow$ Surveillance}, \textit{Surveillance $\rightarrow$ Zoning}, \textit{Zoning $\rightarrow$ Healthcare}, and their average. Reported as \textit{mean $\pm$ std} over 5 runs.}
    \label{tab:cross-topic}
    \end{table*}

\begin{table*}[h!]
\centering
\begin{adjustbox}{max width=\textwidth}
{\scriptsize
\setlength{\tabcolsep}{4pt}
\renewcommand{\arraystretch}{1.25}
\begin{tabular}{llccc}
\toprule
\textbf{Metric} & \textbf{Domain} 
& Gemini 2.0 Flash 
& Qwen2.5-32B-instr. 
& GPT-4o \\
\midrule
\multicolumn{5}{l}{\textbf{Belief State Inference (Accuracy \% $\uparrow$)}} \\
\midrule
& Health & 52.78$\pm$0.65 & 58.15$\pm$0.77 & 50.93$\pm$7.29 \\
& Surv. & 52.79$\pm$1.49 & 64.75$\pm$0.00 & 61.31$\pm$0.37 \\
& Zone & 47.46$\pm$0.66 & 58.41$\pm$0.90 & 51.11$\pm$0.71 \\
& Avg & 51.01$\pm$0.52 & 60.44$\pm$0.47 & 54.45$\pm$2.30 \\
\midrule
\multicolumn{5}{l}{\textbf{Belief Dynamics Update (Accuracy \% $\uparrow$)}} \\
\midrule
& Health & 34.19$\pm$0.12 & 33.77$\pm$0.73 & 37.37$\pm$0.19 \\
& Surv. & 40.71$\pm$0.36 & 40.77$\pm$0.63 & 43.13$\pm$0.75 \\
& Zone & 39.35$\pm$0.16 & 39.93$\pm$0.46 & 37.38$\pm$0.28 \\
& Avg & 38.08$\pm$0.13 & 38.16$\pm$0.44 & 39.30$\pm$0.22 \\
\midrule
\multicolumn{5}{l}{\textbf{Belief Dynamics Update (MAE $\downarrow$)}} \\
\midrule
& Health & 2.13$\pm$0.01 & 2.19$\pm$0.01 & 2.02$\pm$0.01 \\
& Surv. & 1.85$\pm$0.01 & 1.87$\pm$0.01 & 1.80$\pm$0.01 \\
& Zone & 2.05$\pm$0.00 & 2.01$\pm$0.00 & 1.98$\pm$0.01 \\
& Avg & 2.01$\pm$0.00 & 2.02$\pm$0.01 & 1.93$\pm$0.01 \\
\bottomrule
\end{tabular}
}
\end{adjustbox}
\caption{Human \textit{Cross\_Person} test: models trained on one participant and evaluated on another. 
Reported as mean $\pm$ std over 5 runs.}
\label{tab:human-swap-person}
\end{table*}

\begin{figure*}[t]
    \centering
    \includegraphics[width=0.75\linewidth]{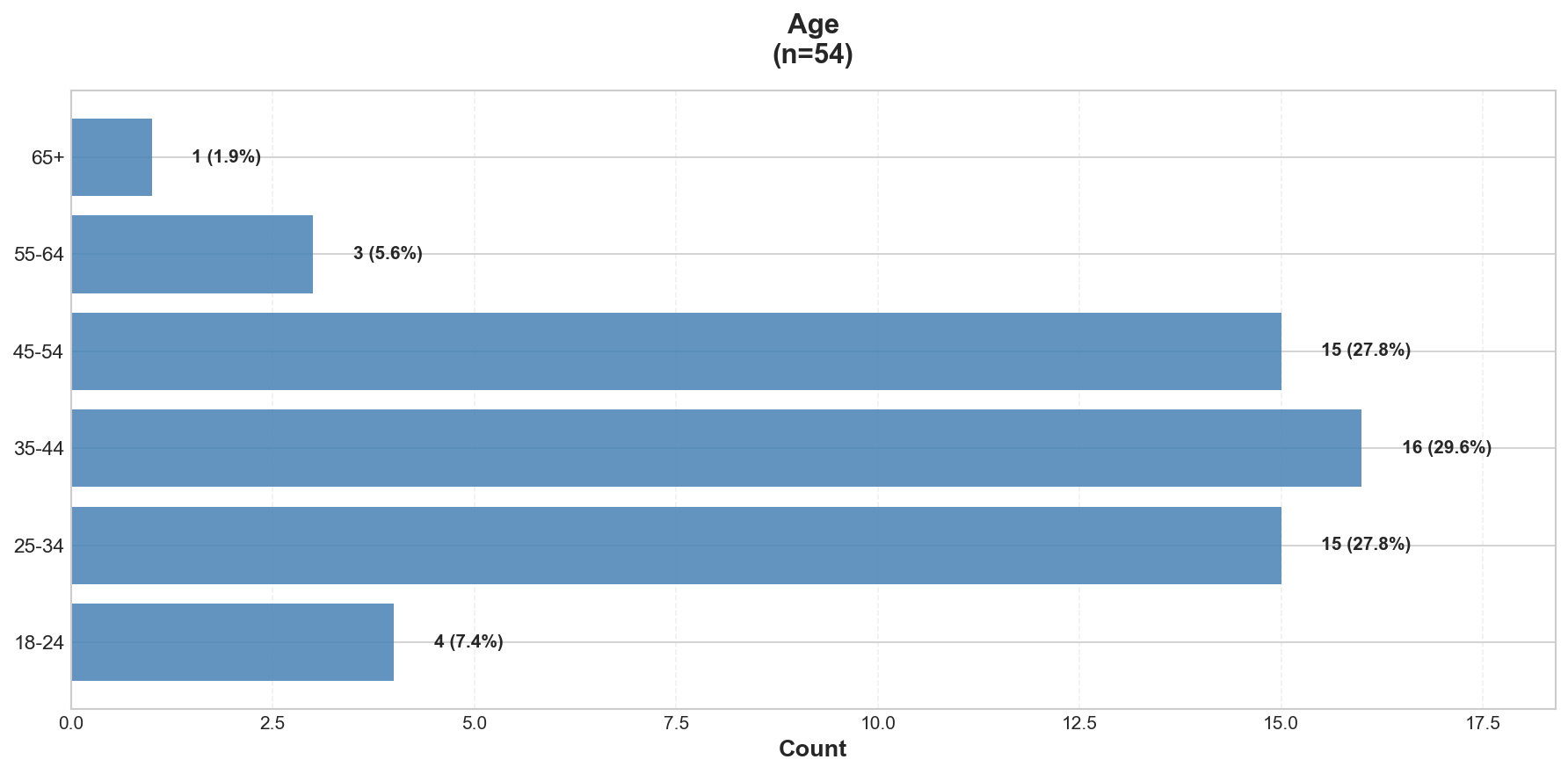}
    \caption{Age distribution of the 54 participants.}
    \label{fig:age}
\end{figure*}

\begin{figure*}[t]
    \centering
    \includegraphics[width=0.75\linewidth]{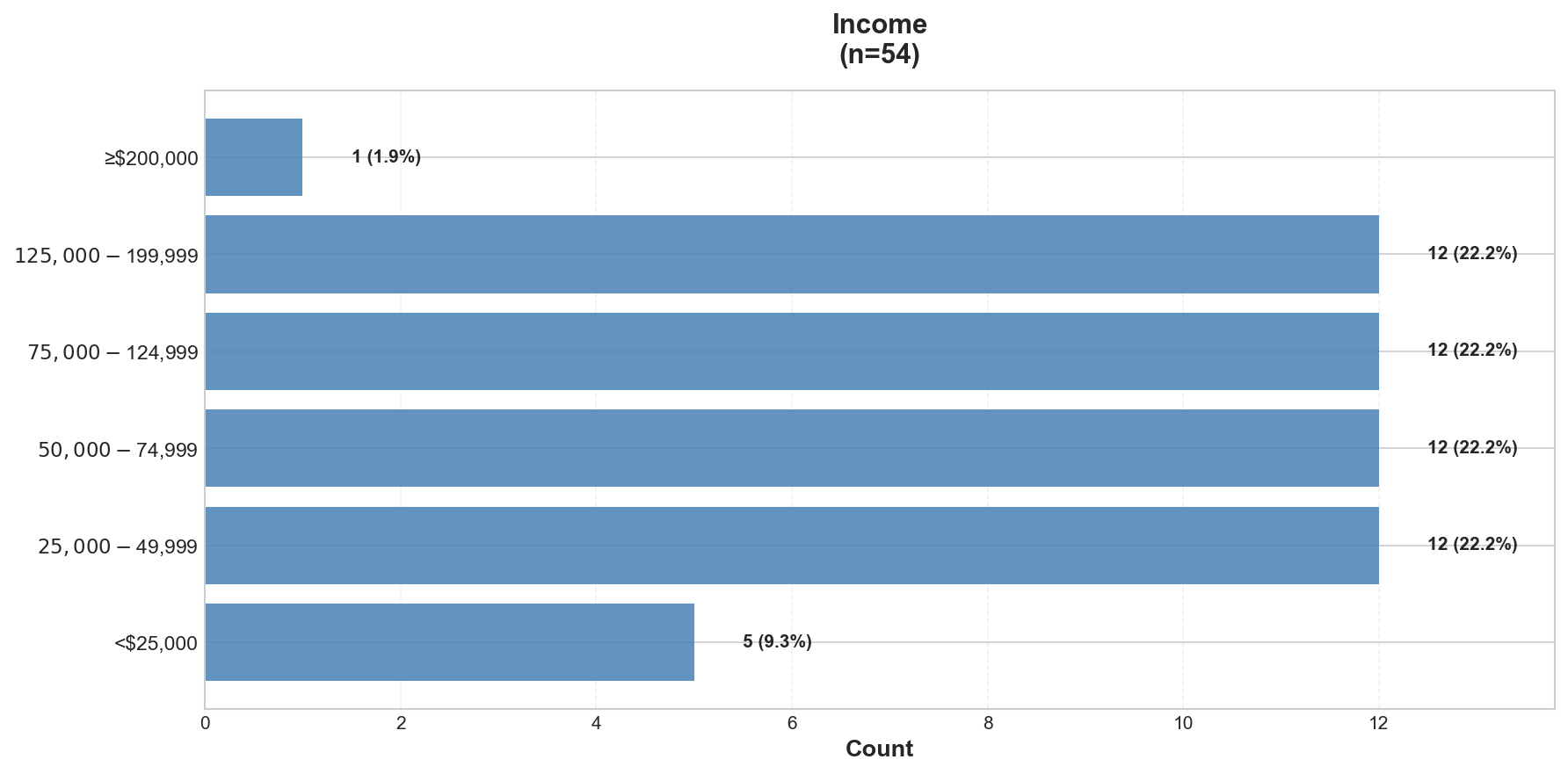}
    \caption{Income distribution of the 54 participants.}
    \label{fig:income}
\end{figure*}

\begin{figure*}[t]
    \centering
    \includegraphics[width=0.75\linewidth]{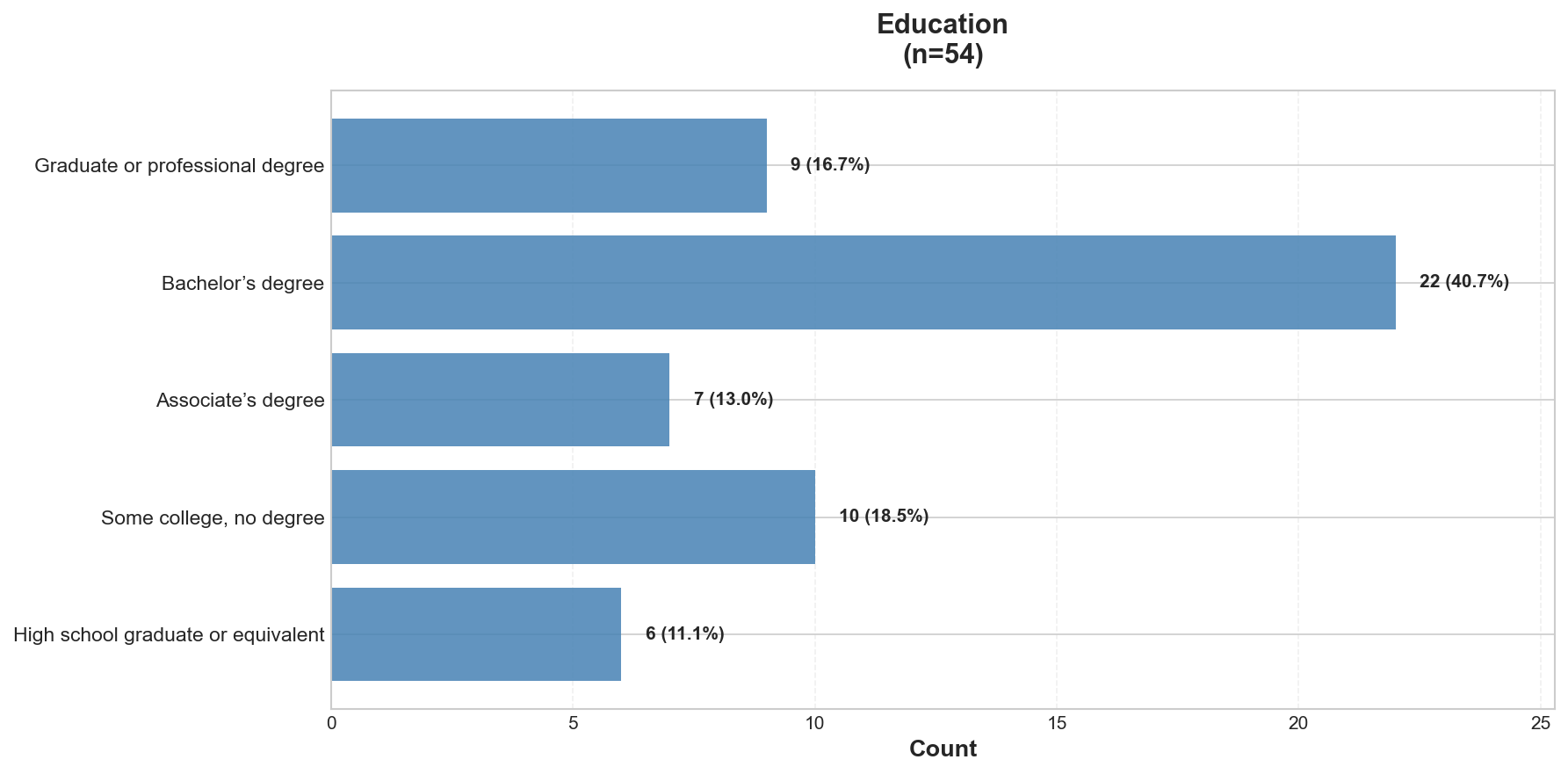}
    \caption{Education level distribution of the 54 participants.}
    \label{fig:education}
\end{figure*}

\begin{figure*}[t]
    \centering
    \includegraphics[width=0.75\linewidth]{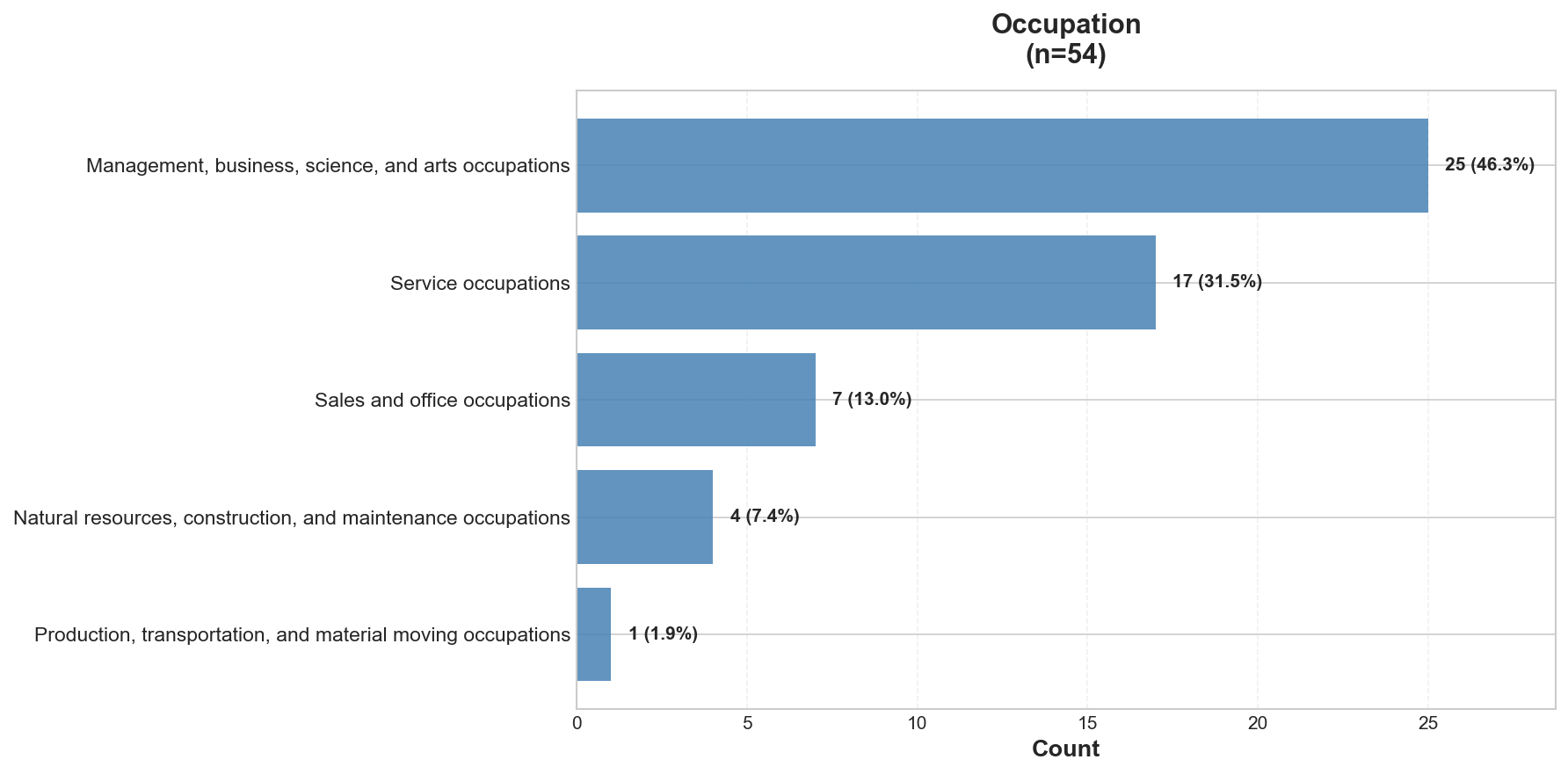}
    \caption{Occupation distribution of the 54 participants.}
    \label{fig:occupation}
\end{figure*}

\begin{figure*}[t]
    \centering
    \includegraphics[width=0.75\linewidth]{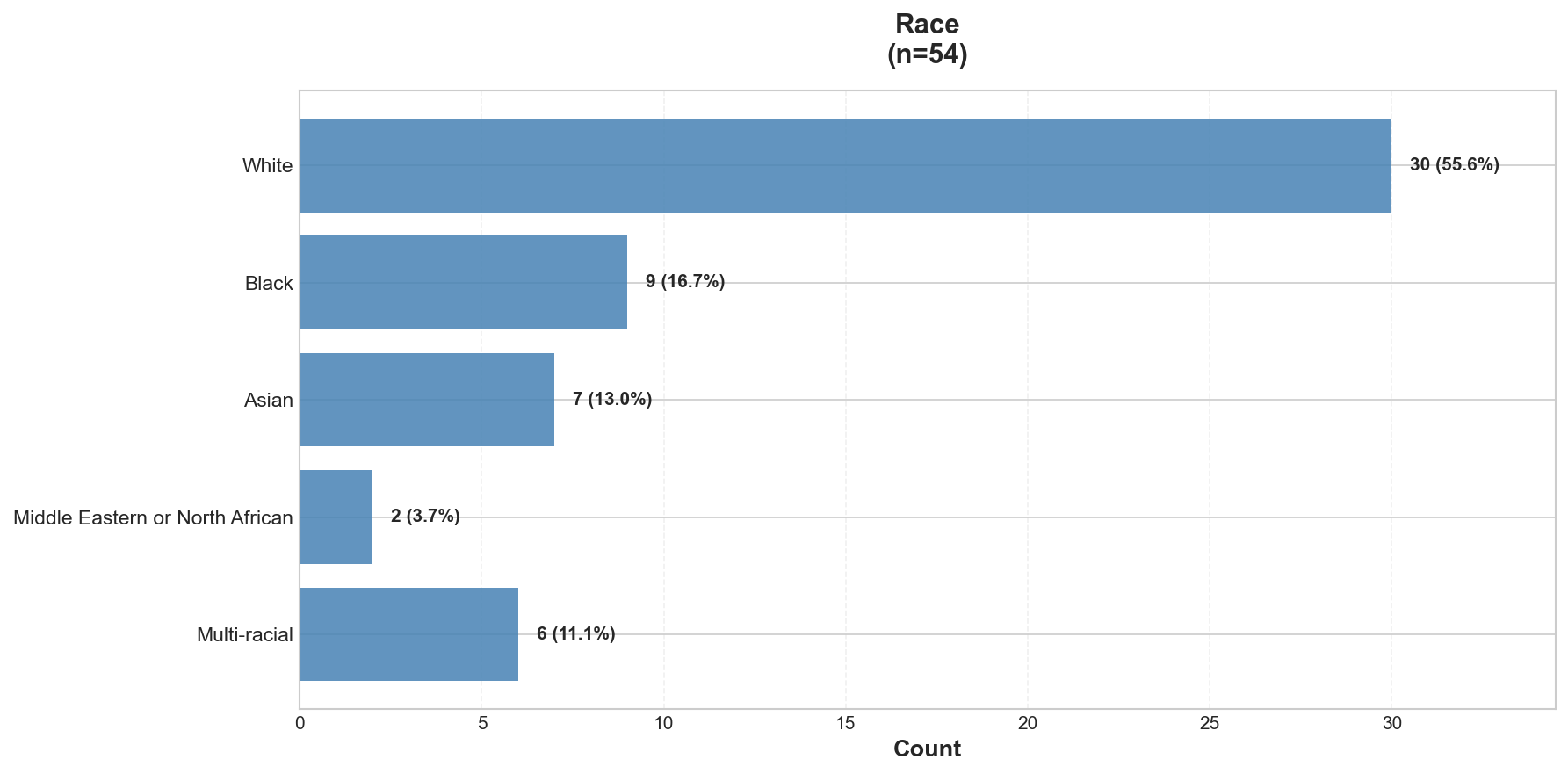}
    \caption{Race distribution of the 54 participants.}
    \label{fig:ethnicity}
\end{figure*}

   \begin{figure*}
        \centering

        \includegraphics[width=\linewidth]{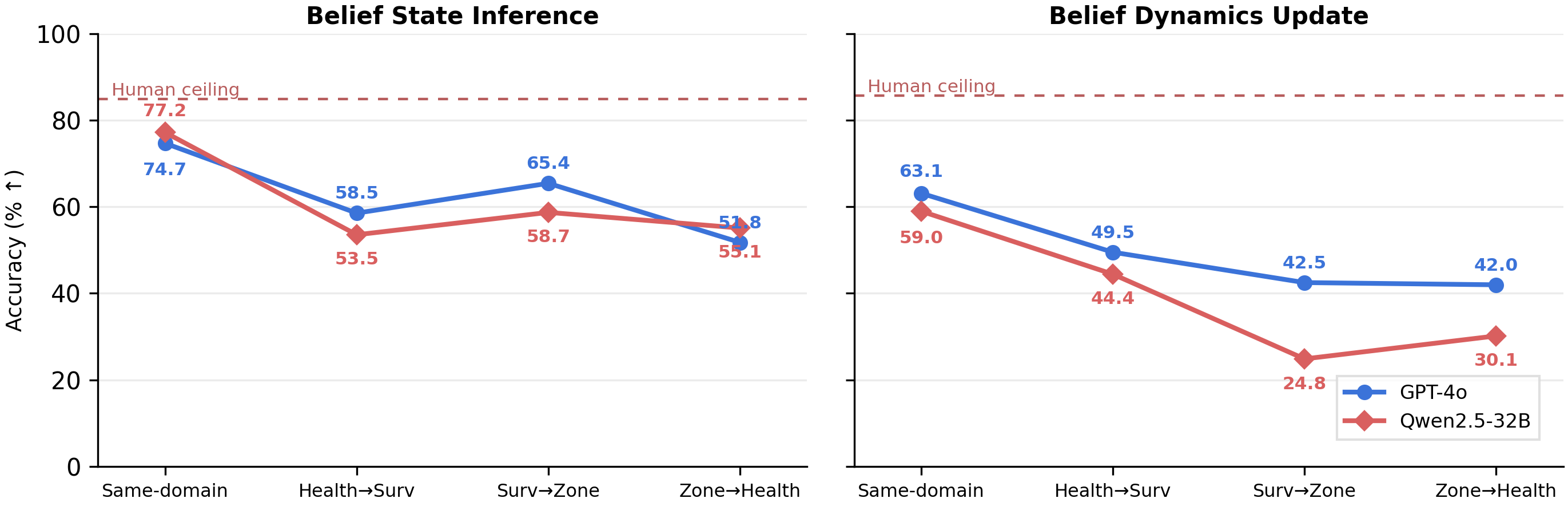}
        
    \caption{The results for cross-domain.}
        \label{fig:cross-domain-fig}

    \end{figure*}

 \begin{figure}[ht]
    \centering
    \includegraphics[width=\linewidth]{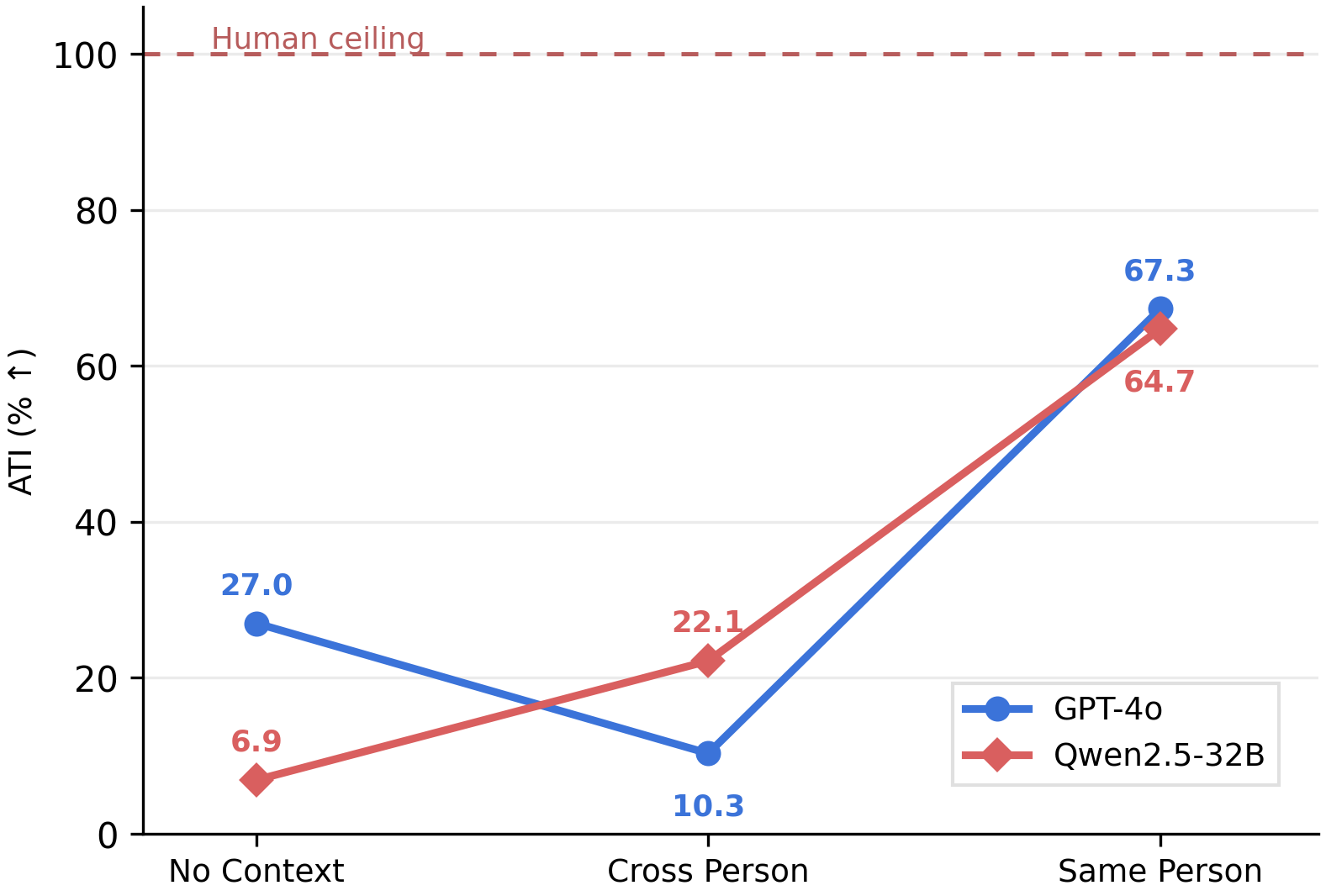}

    \caption{\textbf{Same-person context drives belief updating performance.} Models achieve substantially higher ATI with same-person context than with no context or cross-person context, indicating that useful evidence is person-specific rather than merely contextual.}

    \label{ati-three-condition}
    \end{figure}

\begin{figure}[ht]
    \centering
    \includegraphics[width=\linewidth]{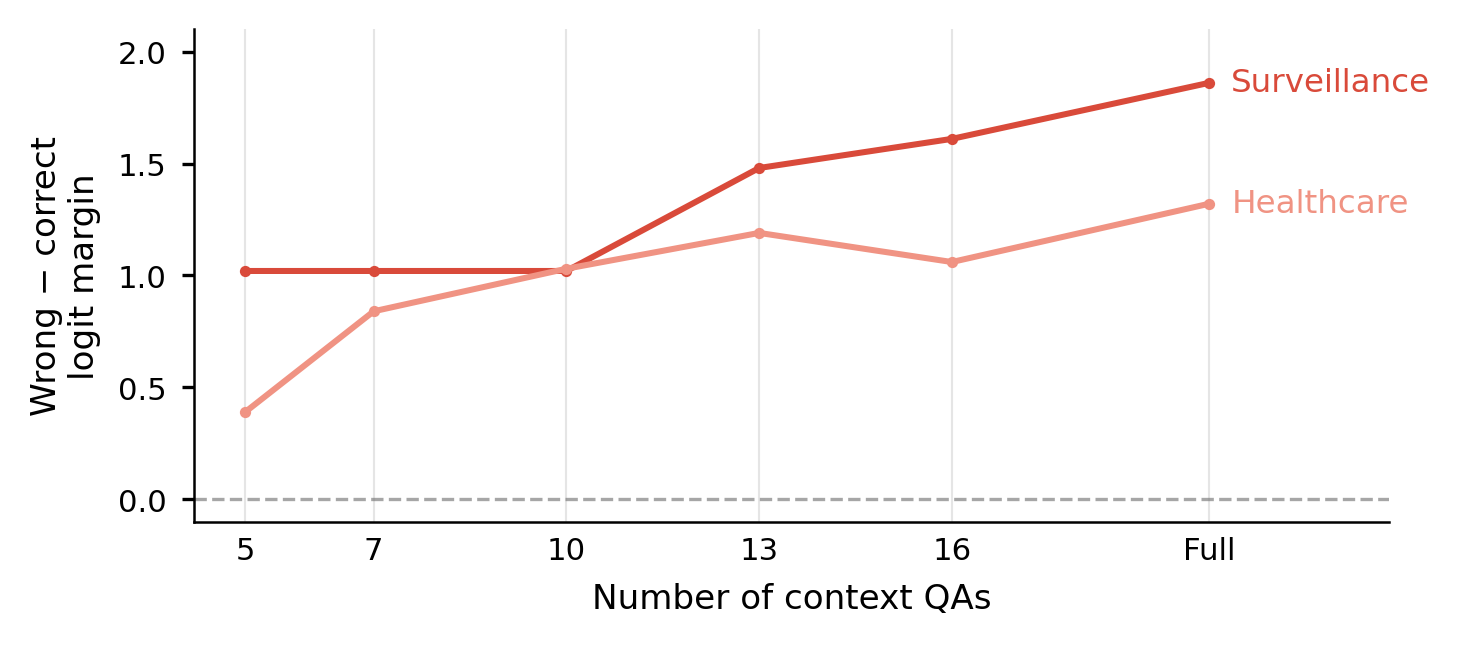}
    \caption{ \textbf{Longer context can increase confidence in wrong updates.} For BDU-Surveillance and BDU-Healthcare, the wrong-minus-correct logit margin increases as more context QAs are added, indicating that additional context can make the model more confidently wrong rather than better informed.}
    \label{fig:logit}
    \end{figure}

\clearpage

\newpage

\begin{figure*}[t]
    \centering
    \includegraphics[width=\linewidth]{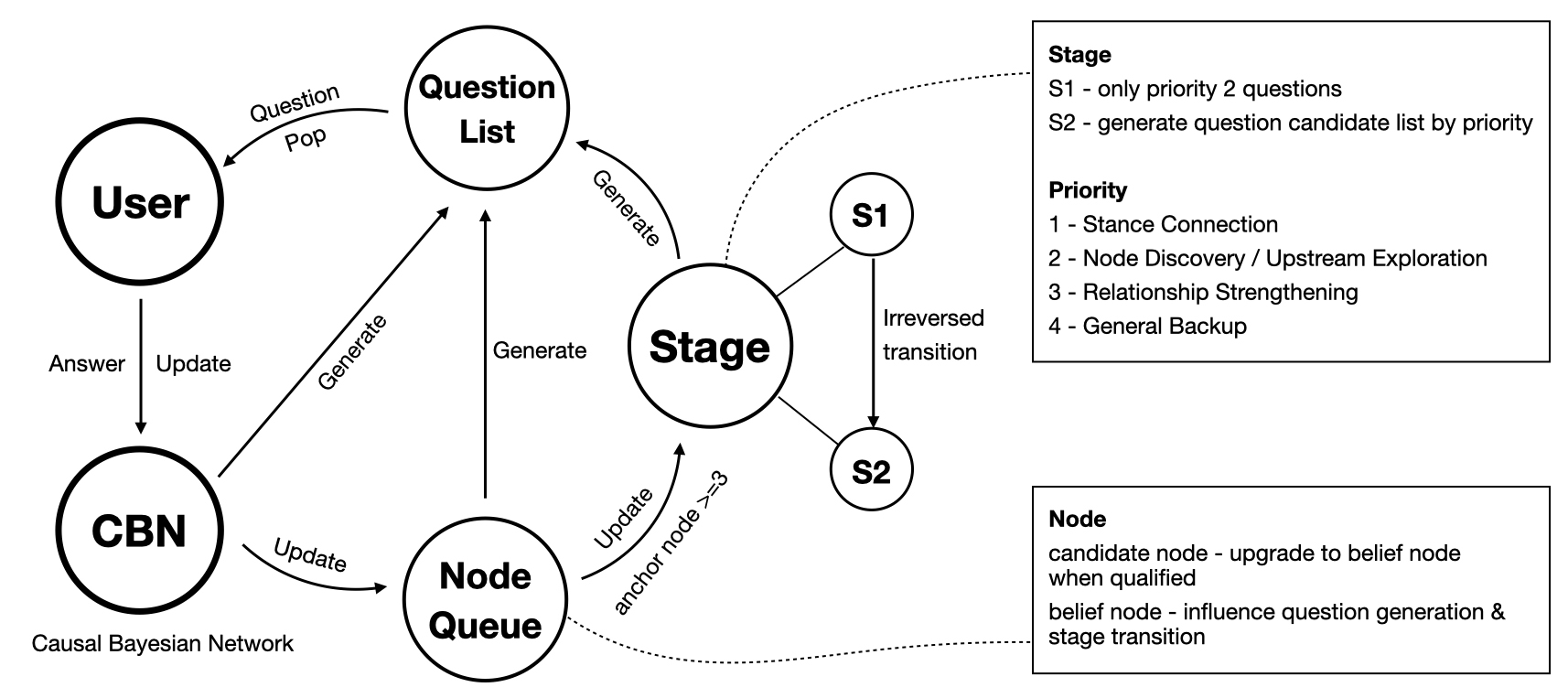}
    \caption{Overview of the QA loop and data structures. The system integrates the Causal Belief Network (CBN), Node Queue, and Question List to guide interaction. Stages regulate question priorities, with an irreversible transition from Stage~1 to Stage~2 once anchor nodes $\geq 3$.}
    \label{fig:qa-loop}
\end{figure*}

\section{Chatbot Design}
\label{app:chatbot}

\subsection*{Core Data Structures}

The system is built on three key data structures:  
(i) the \textbf{Causal Belief Network (CBN)},  
(ii) the \textbf{Node Queue}, and  
(iii) the \textbf{Question List}.  
Together with a staged QA loop, these structures support the dynamic modeling of user beliefs and the generation of targeted questions.  

\subsubsection*{Causal Belief Network (CBN)}
The CBN is the central representation of the user’s belief system. It organizes concepts as nodes and captures their relations as edges.  

\begin{itemize}
    \item \textbf{Nodes.} Nodes represent concepts in the belief system.  
    \begin{itemize}
        \item \emph{Candidate Nodes}: new concepts detected in user answers, under evaluation.  
        \item \emph{Belief Nodes}: stable concepts that have been upgraded from candidates (e.g., due to repeated mentions or high user confidence).  
        \item \emph{Anchor Nodes}: a subset of belief nodes that play a special role in question generation and stage transition.  
    \end{itemize}

    \item \textbf{Edges.} Edges encode causal or influence relations between nodes. They specify direction (source → target), polarity (positive/negative), and optionally strength.  
\end{itemize}

\subsubsection*{Node Queue}
The Node Queue maintains candidate nodes that may become belief nodes.  
\begin{itemize}
    \item \textbf{Entry Condition:} a new concept first appears in user responses.  
    \item \textbf{Upgrade Condition:} node is promoted to belief node when thresholds are met (e.g., frequency of mention, confidence expressed by the user).  
\end{itemize}

\subsubsection*{Question List}
The Question List stores both guiding questions and follow-up questions. It is dynamically updated based on the current CBN and node queue, and it serves as the buffer for delivering the next question to the user.

\subsection*{QA Loop Overview}
The overall interaction loop proceeds as follows:
\begin{enumerate}
    \item \textbf{Initialization.} The CBN contains only a stance node.  
    \item \textbf{User Input.} User provides a new answer.  
    \item \textbf{Update.} Update the CBN and node queue based on the answer.  
    \item \textbf{Question Generation.} Generate new questions using the CBN and queues; append them to the question list.  
    \item \textbf{Next Question.} Select and ask the next question from the list.  
\end{enumerate}

\subsubsection*{Two-Stage Design}
The QA loop has two stages. The transition occurs when the number of anchor nodes $\geq 3$. This transition is one-way; Stage~2 never returns to Stage~1.
\begin{itemize}
    \item \textbf{Stage 1.} Only Priority~2 questions are allowed. Rationale: with too few anchors, meaningful relationship questions are not possible. The system must first accumulate important concepts to avoid premature exploration. 
    \item \textbf{Stage 2.} Questions are selected from a candidate list according to priority. Higher-priority items are chosen first.
\end{itemize}

\subsubsection*{Question Priorities}
\begin{itemize}
    \item \textbf{Priority 1 -- Stance Connection.}  \\
    Purpose: connect essential concepts to the user’s stance.  \\
    Condition: isolated anchor (out-degree = 0, not connected to stance). \\
    \emph{Format:} \textit{``How does \{anchor\} affect your support for \{stance\}? Positive or negative? How strong?''} \\  
    \emph{Example:} \small ``How does privacy protection affect your support for surveillance?'' 
    
    \item \textbf{Priority 2 -- Node Discovery / Upstream Exploration.}  \\
    Purpose: discover new concepts or explore influencing factors of anchors.  \\
    Condition: in Stage~1 or anchor has fewest in-degrees.  \\
    \emph{Format (Stage~1):} \textit{``Tell me more about \{concept\}.''}  \\
    \emph{Format (Stage~2):} \textit{``What factors influence \{anchor\}? Positive or negative?''}  \\
    \emph{Examples:} \small ``Tell me more about public safety.''; ``What factors influence government oversight?'' 
    
    \item \textbf{Priority 3 -- Relationship Strengthening.}  \\
    Purpose: quantify the strength and direction of existing relationships.  \\
    Condition: edge requires parameters or graph pattern needs completion.  \\
    \emph{Format:} \textit{``How strong is the relationship between \{A\} and \{B\}? Positive or negative?''}  \\
    \emph{Example:} \small ``How does technological advancement affect privacy protection? Strong or weak?'' 
    
    \item \textbf{Priority 4 -- General Backup.}  \\
    Purpose: fill in missing information at the end of the interview.  \\
    Condition: remaining questions $\leq 3$ and candidate pool insufficient.  \\
    \emph{Format:} \textit{``Anything else important we have not discussed?''}  \\
    \emph{Example:} \small ``Any clarifications on your previous answers?'' 
\end{itemize}

\newpage
\newpage
\section{Semi-Structured Interview and Causal Belief Network Formalization}
\label{chatbot:CBN}
\begin{figure*}[ht]
    \centering
    \includegraphics[width=\linewidth]{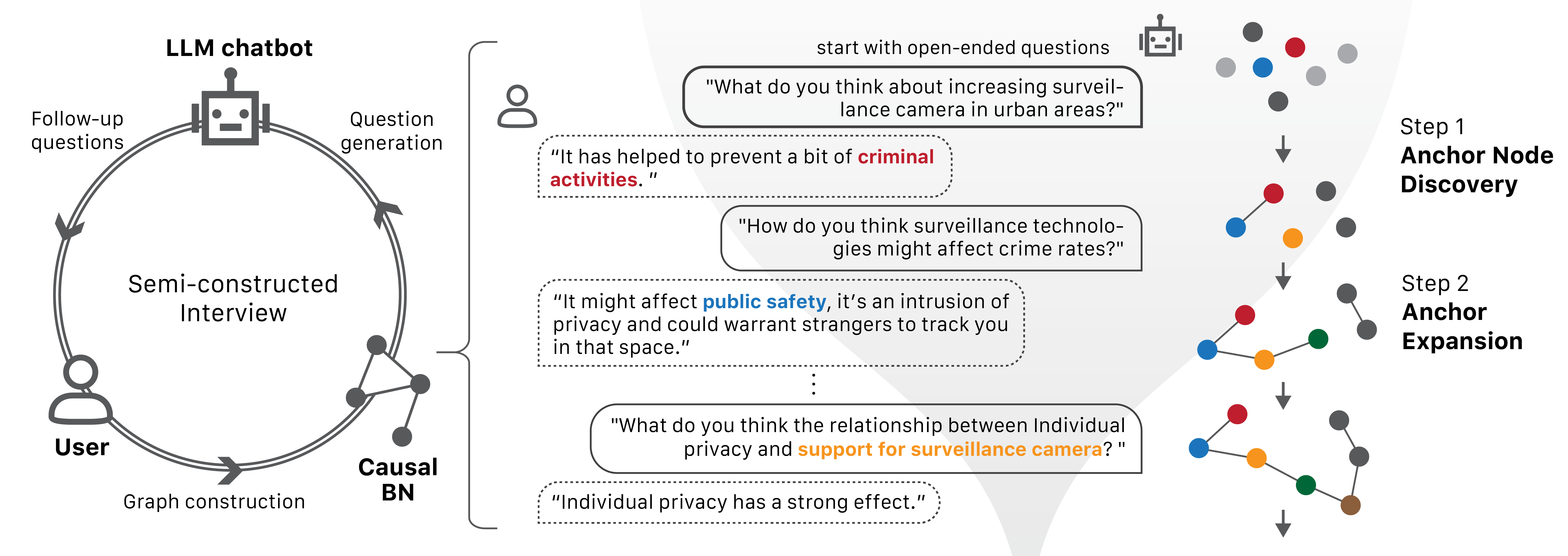}
    \caption{Illustration of the semi-structured interview process and causal belief network construction. 
    The chatbot begins with open-ended questions and extracts candidate concepts from user responses (Anchor Node Discovery). 
    Once three or more anchors are identified, it transitions to targeted follow-ups to expand causal relations (Anchor Expansion). 
    Edges represent directional influences with polarity, forming the evolving CBN.}
    \label{fig:chatbot-formalization}
\end{figure*}
\paragraph{Semi-Constructed Interview Design.}
We use GPT-4 (or Qwen for open-source deployments) as the backbone of a semi-structured interviewer. The model follows a two-phase logic:

\begin{enumerate}
    \item \textbf{Anchor Node Discovery:} From initial open-ended responses, the system uses noun-phrase mining and causal phrase detection to extract candidate belief variables. Candidates that appear in multiple QA pairs or show causal centrality are promoted to \textit{anchor nodes}, representing key ideas around which reasoning is structured.
    \item \textbf{Anchor Expansion:} For each anchor node, the system asks targeted follow-ups (e.g., “What causes this?” or “What does this influence?”). These responses are parsed into edges, which represent directional causal relations with confidence scores and modifiers (positive or negative influence).
\end{enumerate}

\paragraph{Causal BN Formalization.}
Each participant’s graph is a Directed Acyclic Graph (DAG), with nodes $v_i$ labeled by semantically grounded belief variables, and edges $e_{ij}$ denoting belief in the causal influence from $v_i \rightarrow v_j$. We capture the following metadata for each element:
\begin{itemize}
    \item \textbf{Node-level:} Label, frequency across QAs, semantic role (\texttt{external\_state}, \texttt{internal\_affect}, \texttt{behavioral\_intention}), layer depth (e.g., experience $\rightarrow$ value $\rightarrow$ stance).
    \item \textbf{Edge-level:} Confidence (based on question phrasing), polarity (positive or negative), and QA provenance.
\end{itemize}

\paragraph{Edge Probability Estimation.}
Each edge is assigned a probability $P(v_j | v_i)$ based on linguistic indicators in the answer and motif alignment scores:
\begin{equation}
P(v_j | v_i) = \sigma(w_1 \cdot s_{\text{causal}} + w_2 \cdot s_{\text{linguistic}} + w_3 \cdot s_{\text{motif}})
\end{equation}
where $s_{\text{causal}}$ captures explicit causal phrasing, $s_{\text{linguistic}}$ measures structural confidence from the model, and $s_{\text{motif}}$ reflects alignment to previously seen cognitive motif patterns. $\sigma$ is the logistic function.

\paragraph{Demographic Consideration.}
To support downstream generalization and population modeling (Phase III), each interview is paired with structured demographic data (age, housing status, transportation mode, etc.). These attributes allow later stages to interpolate motif distributions and simulate representative reasoning across diverse population groups.

\paragraph{Stopping Criteria.}
The system continues alternating between node discovery and causal expansion until one or more termination conditions are met: (1) no new anchor nodes emerge, (2) motif-based reasoning paths reach convergence, or (3) information gain across simulated stances falls below a threshold.

\paragraph{Forward Simulation and Inference.}
Once an intervention is identified, the causal BN is used to simulate the effects of this intervention. The intervention is applied to the graph as a DO-operation which cuts all incoming edges to the intervened node and updates its distribution. This is followed by a forward simulation to propagate the effects through the network.

Post-processing includes analyzing changes in node probabilities and identifying significant shifts, particularly those related to policy objectives. These results help explain the agent’s behavior and evaluate proposed interventions. This structured method empowers stakeholders to make data-driven decisions based on causal dynamics.

\newpage
\section{Questionnaire General Design}
\label{survey:general}
The questionnaire serves as the foundational layer of HugAgent, designed to capture both baseline beliefs and structured reasoning factors before participants engage in interactive chatbot interviews. The survey was administered through the Prolific platform, ensuring a diverse and demographically balanced pool of respondents. Importantly, not all participants were asked to complete the chatbot phase; instead, all participants began with the questionnaire, and only a subset was later recruited for semi-structured chatbot interviews. This two-stage design allows us to ground conversational transcripts in an already standardized and validated set of structured responses.

The questionnaire is structured into three complementary components.  
First, participants provide \textbf{demographic information}, including age, gender, education, income, housing status, neighborhood context, and transportation habits. These variables are aligned with U.S. Census and urban planning survey standards, enabling stratified analyses of systematic variability in beliefs across groups (e.g., renters versus homeowners, high-income versus low-income).  
Second, participants answer \textbf{stance and intervention items}, rating their support on a 1--10 scale and updating their stance under hypothetical scenarios (e.g., reduced rent under upzoning, reduced household costs under universal healthcare, reduced crime under surveillance).  
Third, each topic includes a standardized \textbf{reason pool}, a set of common factors such as affordability, fairness, privacy, safety, and neighborhood character. After reporting their stance, participants rate on a 1--5 scale how strongly each reason influences their opinion. This structure provides interpretable ground-truth (GT) data for reasoning dimensions and enables cross-participant comparability, since all individuals evaluate the same set of reasons. By aggregating these structured ratings, we can test whether models not only predict overall support levels but also recover the latent weighting of reasons that drive human decision-making. These structured ratings also serve as a reference for aligning open-ended chatbot responses with quantitative belief factors, creating a consistent bridge between free-text explanations and structured data.

\subsection*{Question Types}
We define distinct question types to systematically probe both interpretive reasoning (inferring hidden beliefs) and predictive reasoning (anticipating belief change).  

\paragraph{Type 1.1: Stance elicitation (baseline beliefs).}  
Participants report initial support levels on a 1--10 scale (e.g., ``How much do you support allowing taller apartment buildings in your neighborhood?''). 
This provides the starting point for belief state modeling.

\paragraph{Type 1.2: Reason evaluation.}  
Participants rate how strongly predefined reasons (e.g., economic benefits, fairness, neighborhood character, privacy, efficiency) influence their stance on a 1--5 scale. 
The reason pools are shared across all respondents within a topic, allowing structured comparison across individuals and providing ground-truth data on how value dimensions shape beliefs.

\paragraph{Type 1.3: Contextualized interview beliefs.}  
Through chatbot dialogue, participants explain or justify their stance in natural language. 
These free-form responses provide latent belief evidence, which models must interpret to infer hidden attitudes. The transcripts can be cross-validated against the structured reason evaluations for consistency.

\paragraph{Type 2.1: Scenario-based interventions.}  
Participants evaluate counterfactual scenarios (e.g., ``If rent prices fall by 15\% after upzoning, how would your stance change?''). 
This probes dynamic updating of beliefs in response to outcomes.

\paragraph{Type 2.2: Normative fairness interventions.}  
Scenarios manipulate fairness dimensions (e.g., ``If upzoning applied equally to wealthy neighborhoods'' or ``If cameras were controlled by local boards''). 
These tasks test whether models capture fairness-based belief shifts.

\paragraph{Type 2.3: Conditional trade-offs.}  
Participants consider hybrid conditions (e.g., ``Universal healthcare exists alongside private insurance'' or ``Surveillance footage stored for 48 hours only''). 
These tasks require reasoning under institutional or design constraints.

\begin{table*}[h!]
\centering
\renewcommand{\arraystretch}{1.3}
\setlength{\tabcolsep}{3pt}
\begin{tabular}{p{0.15\linewidth} p{0.25\linewidth} p{0.25\linewidth} p{0.25\linewidth}}
\toprule
\textbf{Task Type} & \textbf{Upzoning} & \textbf{Surveillance Cameras} & \textbf{Universal Healthcare} \\
\midrule
\textbf{Belief Inference} & 
Q: ``On a scale from 1 to 10, how much do you support allowing taller apartment buildings in your neighborhood?''  
A: ``Probably around 3. I worry it changes the character of the area.''  
\newline \textbf{Target:} Low support (3/10); belief: upzoning harms neighborhood character. & 
Q: ``How comfortable do you feel being monitored by public cameras?''  
A: ``Honestly, it makes me uneasy. I don’t trust how the footage is used.''  
\newline \textbf{Target:} Low comfort; belief: privacy concerns about surveillance. & 
Q: ``Do you feel your current health insurance provides adequate coverage?''  
A: ``Not really, I often avoid going to specialists due to cost.''  
\newline \textbf{Target:} Insurance inadequate; belief: high costs limit access. \\
\midrule
\textbf{Reaction Prediction} & 
Scenario: ``After the city allows more apartments, rent prices drop 15\%. Your monthly rent is noticeably lower.''  
\newline \textbf{Target:} Support increases (e.g., +2 on 1--10 scale). & 
Scenario: ``After installing cameras, neighborhood break-ins fall and robberies drop by 20\%.''  
\newline \textbf{Target:} Support increases (stronger acceptance). & 
Scenario: ``After switching to universal healthcare, household out-of-pocket costs fall by \$3{,}000 annually.''  
\newline \textbf{Target:} Support increases (e.g., from 6/10 to 9/10). \\
\bottomrule
\end{tabular}
\caption{Illustrative examples of HugAgent questionnaire and interview tasks. Each domain includes both belief inference and reaction prediction items, enabling evaluation of models on stance attribution and dynamic belief updating.}
\end{table*}


\clearpage
\raggedbottom
\section{Scalar Sensitivity Analysis}
Scalar responses are sometimes viewed as potentially sensitive to sampling noise in LLMs.
To evaluate the stability of LLM-generated scalar outputs, we analyze the variance of model responses on both 1–5 and 1–10 scales across three domains (healthcare, surveillance, and zoning).
For clarity, we visualize one representative model from each category (OpenAI, other closed-source, and open-source).
All remaining models demonstrate similar stability patterns; full results are available upon request.

\subsection{Consistency of Model Predictions}
Across all scalar questions, model predictions remain highly consistent across five independent runs.
With temperature fixed at 0.1 and identical prompting conditions, the variance across runs is minimal, indicating that LLMs produce stable scalar outputs.
Figures below report the standard deviation across runs for each model.

\subsection{Alignment with Human Response Distributions}
Beyond run-to-run stability, we also evaluate whether model-generated scalar scores align with empirical human response distributions.
Across all three domains and both scalar ranges, the models achieve Jensen--Shannon Divergence (JSD) values below $0.10$ and Pearson correlation coefficients of $r \ge 0.2$. 
JSD quantifies the similarity between two probability distributions, where values below $0.10$ are commonly interpreted as indicating close alignment in distributional shape. 
Pearson's $r$ measures the correlation between the model and human mean scores, capturing alignment in central tendencies. 
Together, these complementary metrics provide evidence that models capture both the overall distributional patterns and the relative ordering of human scalar judgments. 
Across most settings, we observe low JSD ($0.02$--$0.08$) and moderate positive correlations ($r \approx 0.2$--$0.4$), consistent with prior findings on LLM stability. 
Although certain tasks (e.g., zoning on the 1--10 scale) exhibit slightly higher variability, the broader distributional trends remain similar to human responses. 
Taken together, these results demonstrate that model-generated scalar outputs are stable, well-aligned with human response patterns, and that any numerical noise at the item level does not affect the main conclusions of the study.

\begin{figure*}[h]
    \centering
    \includegraphics[width=0.75\linewidth]{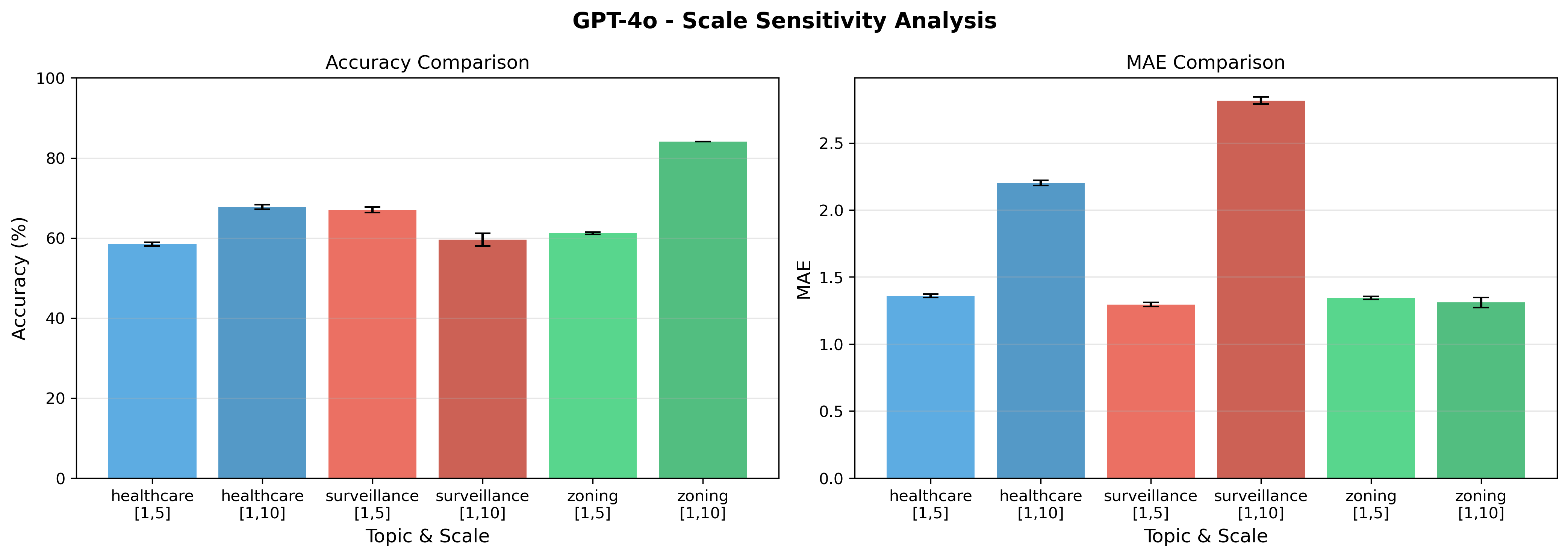}
    \caption{Scalar sensitivity analysis for GPT-4o.}
    \label{fig:4o-error-bar}
\end{figure*}

\begin{figure*}[h]
    \centering
    \includegraphics[width=0.75\linewidth]{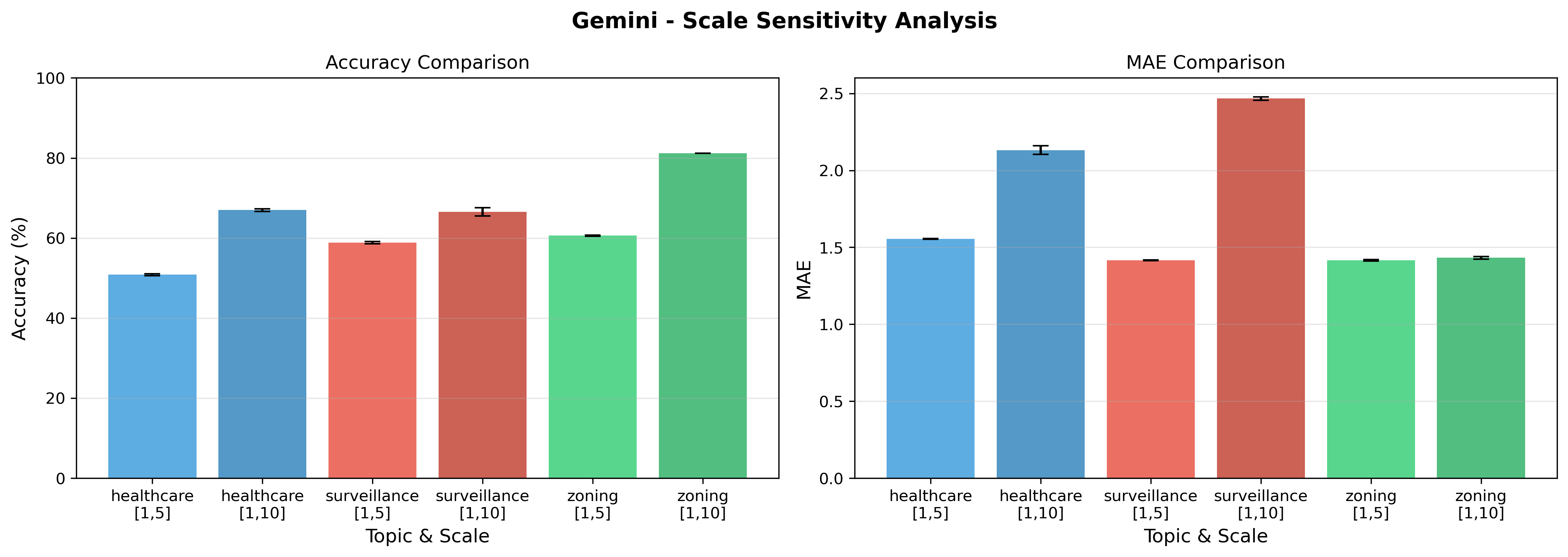}
    \caption{Scalar sensitivity analysis for Gemini 2.0 Flash.}
    \label{fig:gemini-error-bar}
\end{figure*}

\begin{figure*}[h]
    \centering
    \includegraphics[width=0.75\linewidth]{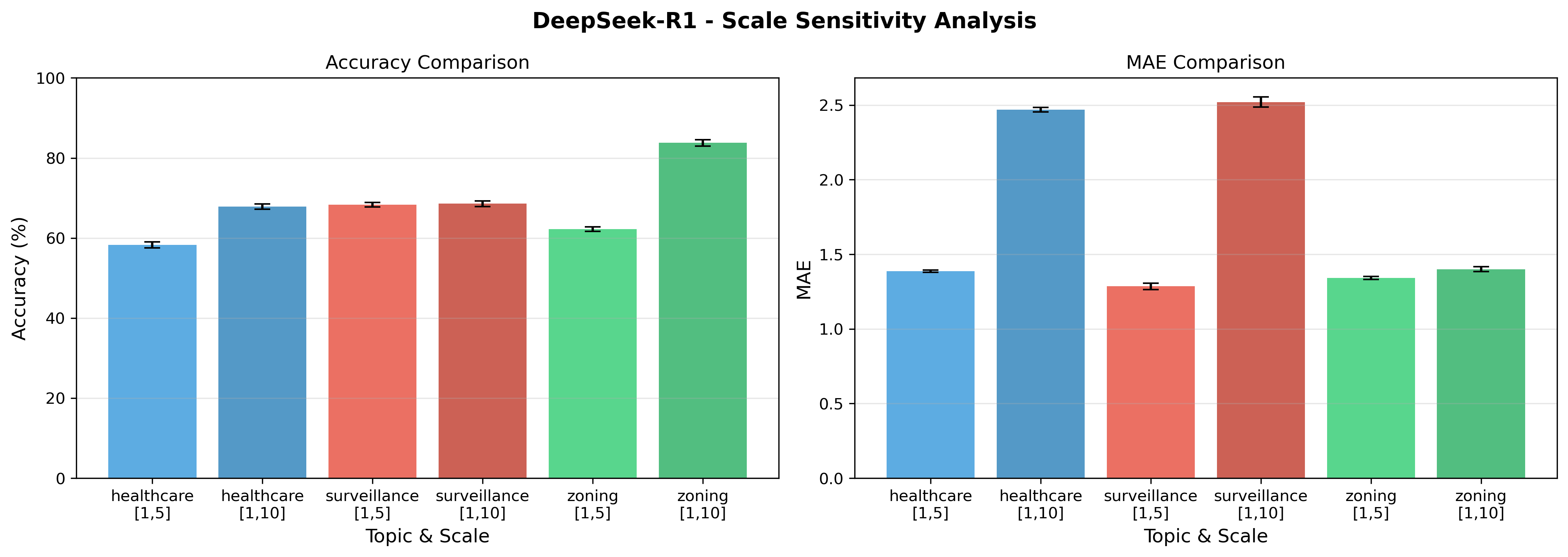}
    \caption{Scalar sensitivity analysis for DeepSeek-R1. }
    \label{fig:deepseek-error-bar}
\end{figure*}

\begin{figure*}[h]
    \centering
    \includegraphics[width=0.75\linewidth]{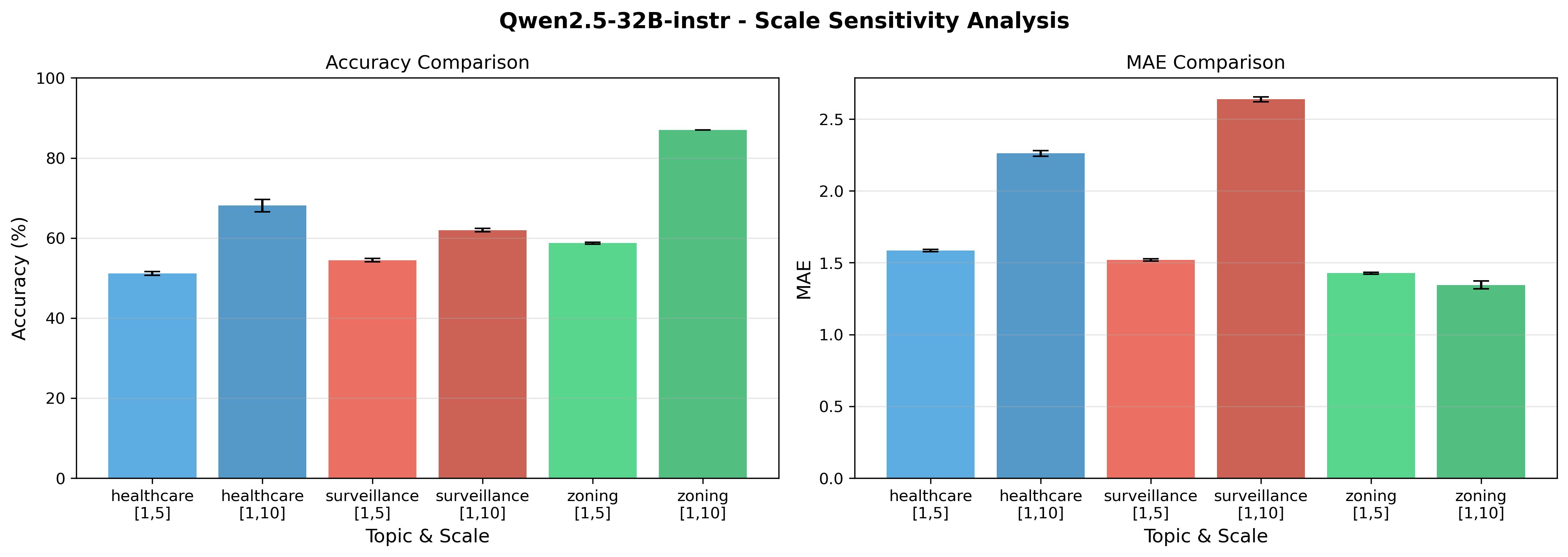}
    \caption{Scalar sensitivity analysis for Qwen2.5-32B-Instr.}
    \label{fig:qwen-error-bar}
\end{figure*}

\begin{figure*}[ht]
    \centering
    \includegraphics[width=0.75\linewidth]{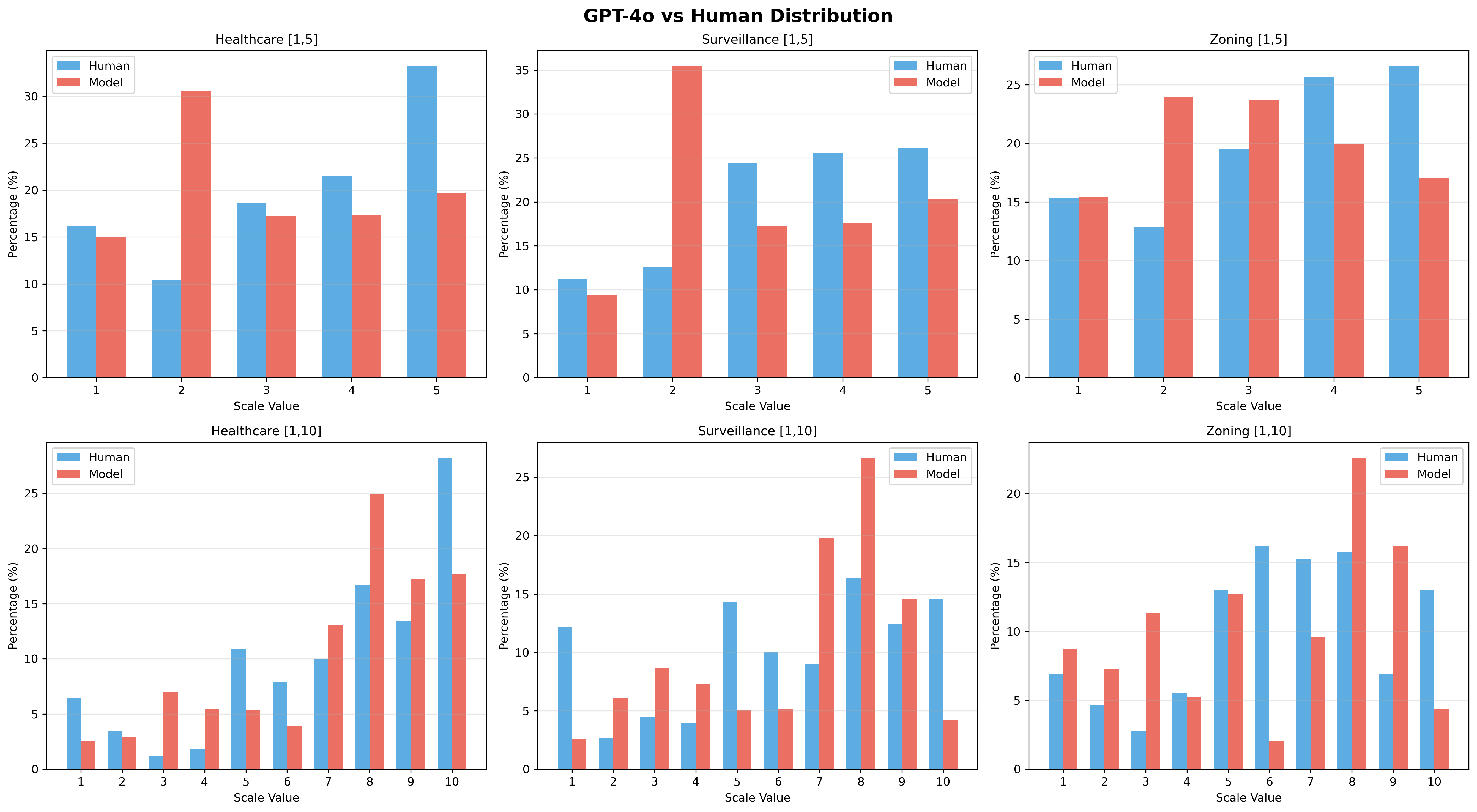}
    \caption{Scalar response distributions for GPT-4o across three topics and two scales.}
    \label{fig:4o-distribution}
\end{figure*}

\begin{figure*}[h]
    \centering
    \includegraphics[width=0.75\linewidth]{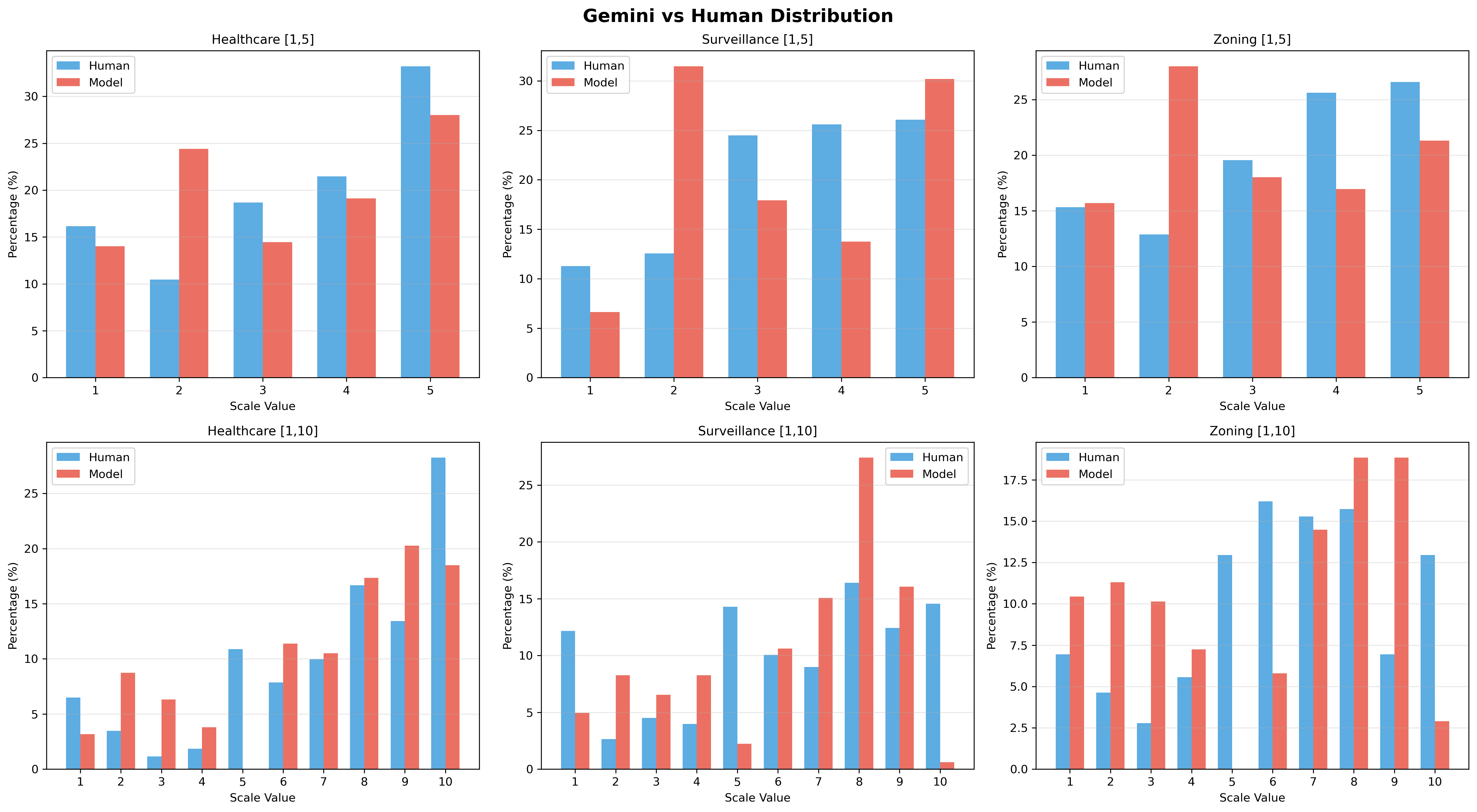}
    \caption{Scalar response distributions for Gemini-2.0-Flash across three topics and two scales.}
    \label{fig:gemini-distribution}
\end{figure*}

\begin{figure*}[h]
    \centering
    \includegraphics[width=0.75\linewidth]{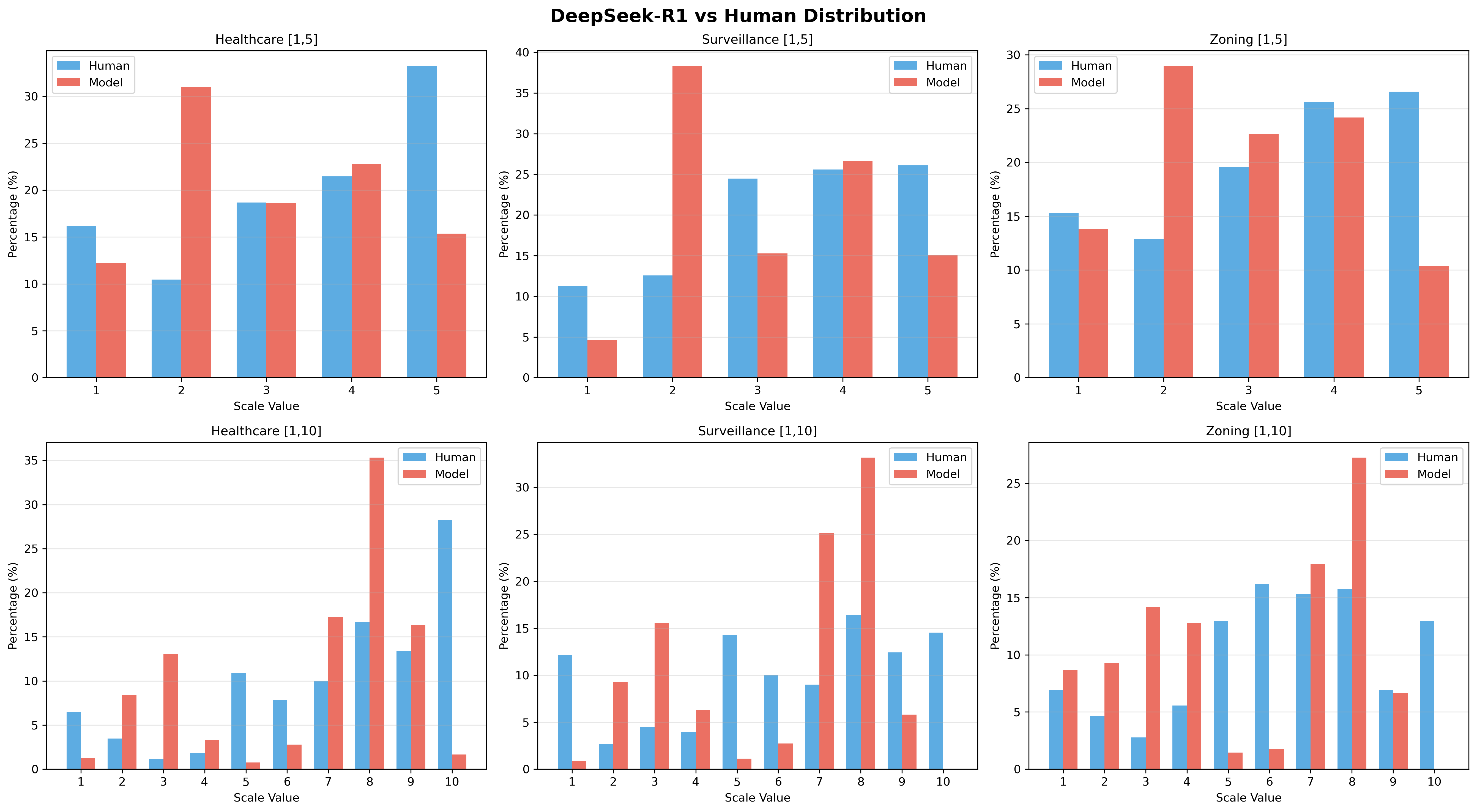}
    \caption{Scalar response distributions for DeepSeek-R1 across three topics and two scales.}
    \label{fig:deepseek-distribution}
\end{figure*}

\begin{figure*}[h]
    \centering
    \includegraphics[width=0.75\linewidth]{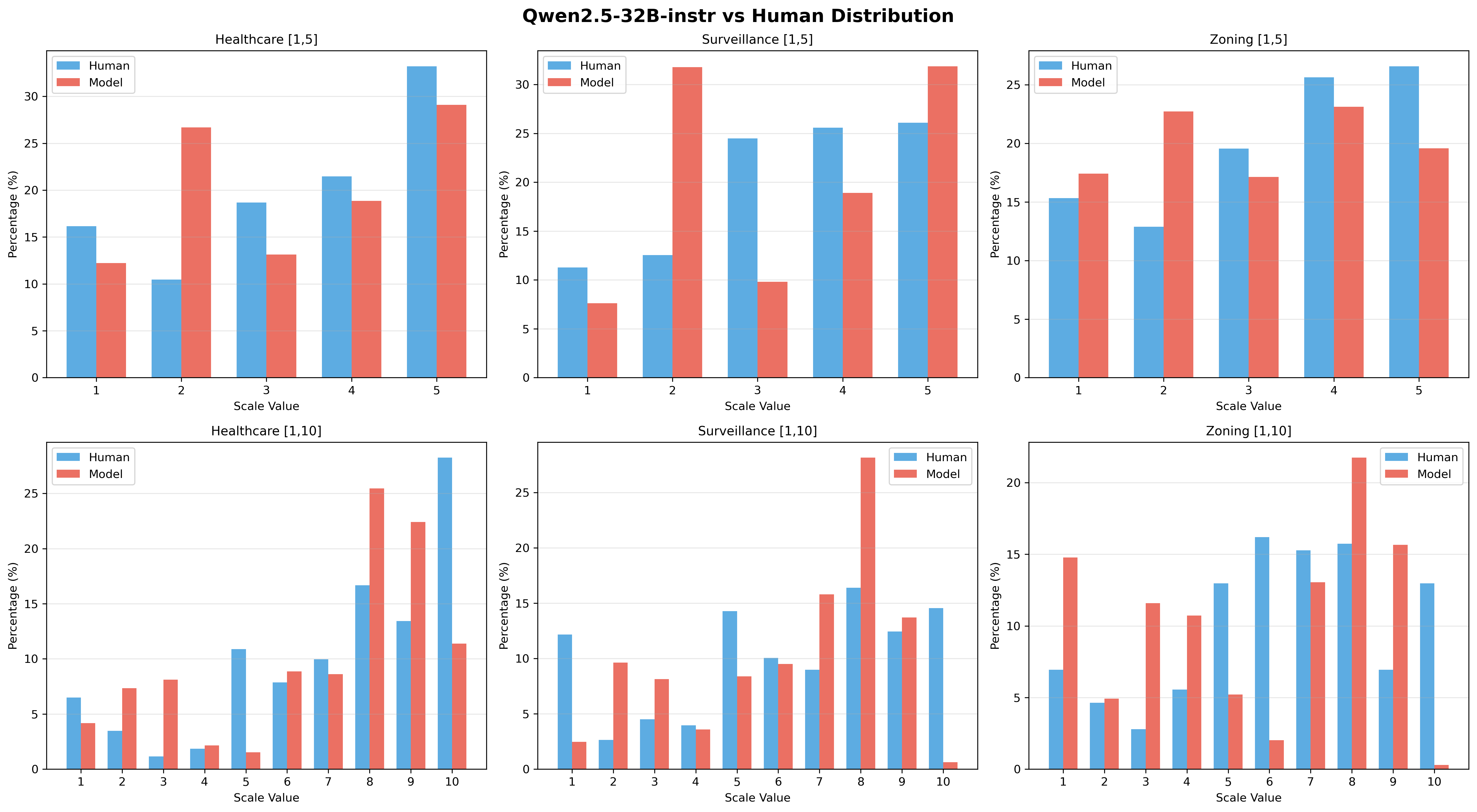}
    \caption{Scalar response distributions for Qwen2.5-32B-Instr. across three topics and two scales.}
    \label{fig:qwen-distribution}
\end{figure*}

\clearpage
\section{Zoning Opinion Questionnaire (Human Evaluation)}
\label{survey:zoning}

To rigorously evaluate the fidelity of our generative agents' responses against real human participants, we conducted a structured public opinion survey titled \textit{General Housing \& Upzoning Public Opinion Survey}. The survey was carefully designed to facilitate comparison between human-generated responses and those from LLM-based agents, specifically targeting residents of United states.

\subsection*{Motivation and Objectives}
This survey aimed to assess public opinion on urban upzoning scenarios, capturing nuanced attitudes toward housing policies and their underlying reasoning. Our goal was to determine whether generative agents could reliably replicate human response patterns, especially regarding sensitive issues such as neighborhood change, density increases, and emotional responses like YIMBY (Yes In My Backyard) and NIMBY (Not In My Backyard).

\subsection*{Survey Structure and Methodology}
The survey comprised two primary sections:

\textbf{Section 1: Demographic and Background Information}

Participants provided detailed demographic data aligned with U.S. Census Bureau categories: Age, Housing status (owner or renter), Income levels, Occupation, Marital status, Presence of children, Transportation mode, Monthly rent as a percentage of income, Residential mobility, ZIP code or proximity-based location verification.

To ensure data quality, participants were required to explicitly answer an attention check question.

\textbf{Section 2: Scenario-Based Opinion Measurement}

Participants were first asked general zoning questions and rated their support for allowing larger, taller apartment buildings in their neighborhood on a 1–10 Likert scale (1 = strongly oppose, 10 = strongly support). Each scenario was accompanied by a set of related factors, which participants evaluated on a 1–5 scale (1 = no impact, 5 = very large impact), regardless of whether the impact was positive or negative.
The factors included: Housing supply and availability
, Affordability for low- and middle-income residents
, Neighborhood character and visual compatibility
, Traffic and parking availability
, Walkability and access to amenities
, Noise, congestion, or infrastructure strain
, Fairness and distribution of development
, Economic vitality for local businesses
, Building height/scale relative to surroundings
, Property values or homeownership concerns.

Clarifying examples were provided to ensure consistent interpretation of impact ratings.

\subsection*{Data Collection and Implementation}

The survey was implemented using Google Forms and distributed via the Prolific platform, with compensation set at \$12/hour. Participants were guided through the survey flow with embedded instructions and examples to ensure comprehension and engagement.

\subsection*{Transparency}

All survey items, design rationales, and filtering criteria are publicly documented to support reproducibility and public trust. This enables rigorous evaluation of generative agents' ability to simulate human attitudes under complex, emotionally and politically sensitive policy conditions.

\clearpage

\section{Universal Healthcare Questionnaire}
\label{survey-healthcare}

\subsection*{Motivation and Objective}
This survey was designed to evaluate whether a structured reasoning system—based on Bayesian networks extracted from interviews and conditioned large language models (LLMs)—can simulate or recover human judgments on complex policy issues. In this case, we focus on universal healthcare, a topic involving tradeoffs across fairness, cost, autonomy, and trust.

Rather than simply measuring stance, the survey was constructed to expose the participant's reasoning pathway, enabling fidelity evaluation at both outcome and process levels.

\subsection*{Survey Structure and Methodology}
The survey design draws on the four-stage cognitive model of survey response \citep{tourangeau2000psychology}:
\begin{itemize}
    \item \textbf{Comprehension}: Questions were phrased clearly and definitions were provided (e.g., what universal healthcare entails).
    \item \textbf{Retrieval}: Participants were asked to recall relevant experiences (e.g., delays in care, interactions with public systems).
    \item \textbf{Judgment}: Participants evaluated tradeoffs and reflected on personal values.
    \item \textbf{Response}: Structured Likert scales captured quantified opinions.
\end{itemize}

\subsection*{Survey Components}
The survey includes:
\begin{itemize}
    \item \textbf{Stance Rating}: Support for universal healthcare on a 1–10 scale.
    \item \textbf{Personal Experience}: Items capturing healthcare access and insurance adequacy.
    \item \textbf{Baseline Reason Evaluation}: Participants rated 13 carefully constructed reasons (e.g., fairness, efficiency, innovation) for their general influence on stance.
\end{itemize}

\subsection*{Counterfactual Scenarios}
To probe reasoning dynamics and test the model's sensitivity to causal perturbations, four counterfactual scenarios were introduced, each followed by a stance re-rating and a focused subset of reasons. Scenarios included:
\begin{enumerate}
    \item National cost reduction with increased wait times.
    \item Household savings of \$3,000 annually.
    \item Retention of private insurance alongside a public system.
    \item Coverage limited to essential services.
\end{enumerate}

Participants re-evaluated selected reasons in the context of each scenario (e.g., “I worry about tax increases” or “Universal healthcare might reduce personal choice in care”) on a 1–5 scale, allowing analysis of belief shifts.

\subsection*{Reason Design}
Reasons were drawn from qualitative policy discourse and refined to:
\begin{itemize}
    \item Reflect distinct value dimensions (e.g., equality, responsibility, institutional trust).
    \item Avoid biasing language (neutral framing, no moral triggers).
    \item Enable both positive and negative stance justifications across political orientations.
\end{itemize}

Each reason was independently interpretable and mapped to latent causal factors in the underlying Bayesian model. Subsets of reasons were assigned to each counterfactual scenario to ensure relevance while reducing redundancy.

\newpage
\section{Surveillance Camera Questionnaire}
\label{survey:surveillance-survey}

\subsection*{Motivation and Objective}

This survey is designed to evaluate the reasoning fidelity of structured models such as Bayesian Networks (BNs) when paired with large language models (LLMs). Specifically, it tests whether a BN+LLM system can simulate human responses to policy questions about public surveillance more faithfully than a baseline persona-based LLM. To do this, we use controlled question design inspired by cognitive science and causal reasoning frameworks.

\subsection*{Survey Structure and Methodology}

The survey design draws on the four-stage cognitive model of survey response \citep{tourangeau2000psychology}:
\begin{enumerate}
    \item \textbf{Comprehension}: Understand the question and context.
    \item \textbf{Retrieval}: Recall relevant experiences and beliefs.
    \item \textbf{Judgment}: Synthesize and evaluate relevant considerations.
    \item \textbf{Response}: Map judgment to a scale-based response.
\end{enumerate}

This model guides both our baseline attitude elicitation and our counterfactual design. The survey consists of:

\paragraph{Section 1: Baseline Stance and Experience}
Participants rate their general support for public surveillance (1--10), followed by personal experiences such as feelings of safety, comfort, and negative interactions with surveillance technology.

\paragraph{Section 2: General Reason Evaluation}
Participants evaluate the importance of twelve potential reasons (1--5 Likert scale) influencing their baseline stance, including factors like privacy, crime prevention, power misuse, and behavioral impacts.

\paragraph{Section 3: Counterfactual Scenarios and Dynamic Reasoning}
Participants are then presented with three hypothetical surveillance policy changes:
\begin{itemize}
    \item Crime Reduction vs. False Arrest Tradeoff
    \item Limited Data Retention (48h)
    \item Community-Controlled Surveillance
\end{itemize}

For each scenario:
\begin{itemize}
    \item Participants rate how the new information affects their stance (1--10 scale).
    \item Then, they re-evaluate a scenario-specific subset of 3--5 reasons (1--5 scale) that are most relevant under the new condition.
\end{itemize}

This design allows us to evaluate whether the model (and human) responses adjust not only the final stance, but also the internal reasoning paths---a critical distinction for validating structural cognitive models.

\subsection*{Design Highlights}

\begin{itemize}
    \item \textbf{Cognitive fidelity}: Question wording avoids surface cues and forces reasoning across multiple values (e.g., privacy vs. safety, trust vs. control).
    \item \textbf{Counterfactual sensitivity}: Each scenario targets a specific edge in the causal BN, enabling us to observe how reason weights shift under perturbation.
    \item \textbf{Explanation delta}: By comparing reason weights before and after each scenario, we quantify whether the model exhibits structural adaptation or static stance mimicry.
\end{itemize}

\subsection*{Data Collection and Implementation}

The survey was implemented using Google Forms and distributed via the Prolific platform, with compensation set at \$12/hour. Participants were guided through the survey flow with embedded instructions and examples to ensure comprehension and engagement.

\subsection*{Transparency}

All survey items, design rationales, and filtering criteria are publicly documented to support reproducibility and public trust. This enables rigorous evaluation of generative agents' ability to simulate human attitudes under complex, emotionally and politically sensitive policy conditions.

\newpage
\section{Prompt}
\subsection{Task formatting prompt}
\label{sec:prompt}
\begin{tcolorbox}[
    title= System Prompt,
    colback=gray!5,
    colframe=black!80,
    coltitle=white,
    fonttitle=\bfseries,
    breakable
]

    zoning: "You are an expert at analyzing conversations about urban policy to extract causal beliefs.\\
    surveillance": "You are an expert at analyzing conversations about surveillance and public safety to extract causal beliefs.\\
    healthcare": "You are an expert at analyzing conversations about healthcare policy to extract causal beliefs.

\end{tcolorbox}

\begin{tcolorbox}[
    title=User Prompt,
    colback=gray!5,
    colframe=black!80,
    coltitle=white,
    fonttitle=\bfseries,
    breakable
]
Based on the following conversation about \{conversation\_topic\}, identify ALL question-answer pairs that reveal the person's beliefs about causal relationships between different factors.

Conversation:
\{context\_text\}
Your task:
1. Find ALL Q\&A pairs that show how the person believes one factor affects another (up to 10 pairs)
2. For each pair, create a direct question asking about the influence level using everyday language
3. Based on the person's answer, determine their belief about the effect.\\
Selection rule:
- PRIORITIZE items with dependency\_level >= 1 (needs-context). If fewer than 10 such items exist, then fill the remainder with the best dependency\_level = 0 items.
- Prefer diverse factor pairs; avoid near-duplicates.\\

Return JSON format as an array.

\{answer\_options\_text\}

Use simple, everyday language for the factors. Examples by topic:  \\
Zoning: "building more housing" instead of "upzoning policies", "traffic congestion", "neighborhood character"  .\\
Surveillance: "installing cameras" instead of "surveillance systems", "crime rates", "privacy concerns"  .\\
Healthcare: "universal coverage" instead of "healthcare policy", "wait times", "healthcare costs".

Return up to 10 belief inference questions maximum.
\end{tcolorbox}

\subsection{Evaluation prompt}
\begin{tcolorbox}[
    title=System Prompt,
    colback=gray!5,
    colframe=black!80,
    coltitle=white,
    fonttitle=\bfseries,
    breakable
]
    You are an expert psychologist specializing in Theory of Mind and belief inference.\\
    Your task: analyze conversation transcripts to infer what the participant believes about causal relationships.
    Focus on understanding their mental model - what they think causes what, not what is objectively true.\\
    Consider their background, conversation patterns, and implicit beliefs expressed through their responses.
    Base your inference strictly on evidence from their statements, not general assumptions.
\end{tcolorbox}

\begin{tcolorbox}[
    title=User Prompt,
    colback=gray!5,
    colframe=black!80,
    coltitle=white,
    fonttitle=\bfseries,
    breakable
]

\{Context QA + Demographic information\}\\
Based on the evidence above (including Conversation History and Person's Background), respond with ONLY the single letter ({options\_str}) that best represents this person's belief.)  

\end{tcolorbox}
\newpage
\section{Extended Design Principles}
\label{app:design}

\paragraph{Open-ended reasoning as principle}
Our benchmark targets \textit{reasoning as a dynamic and individualized process}, rather than static prediction. 
We therefore adopt an \textbf{open-ended elicitation principle}: instead of pre-defining fixed question banks, HugAgent uses a single \textit{guiding question} to initiate a semi-structured conversation. 
All follow-up questions are generated adaptively within the same dialogue, grounded in the participant’s own responses. 
This design enables \textit{deep, conversational reasoning} to unfold while minimizing artificial scaffolding from the chatbot itself. 
We do not claim fully open-world coverage; rather, we emphasize \textit{open-domain extensibility}: by simply swapping the guiding question, the benchmark can be ported to new domains while maintaining consistency in evaluation. 
Such minimal-interaction protocols align with prior work showing that lightweight conversational scaffolds preserve ecological validity in human reasoning studies \citep{van1994think,clark1996using,sap2019socialiqa,driess2023palm}. 
\textbf{This principle directly motivates the guiding-question chatbot protocol we describe in Appendix \ref{app:chatbot}.}

\paragraph{Proxy tasks of reasoning}
To evaluate whether models capture not only what individuals believe but also how their beliefs evolve, we operationalize reasoning through two proxy tasks: 
\textit{belief state inference} (recovering stance and factor polarity from context) and \textit{belief dynamics update} (predicting stance shifts and reweighting under new evidence). 
These tasks follow the tradition of modeling belief revision as a tractable proxy for underlying cognitive processes \citep{gopnik2004theory,sloman2009causal}. 
While other proxies could be envisioned, these two are the most direct operationalizations of \textit{individual reasoning trajectories}, balancing interpretability and task difficulty. 
\textbf{This motivates our benchmark’s two-task structure, detailed in Appendix \ref{app:chatbot}.}

\paragraph{Upper bound via test--retest reliability}
A natural question is whether human annotators could serve as the benchmark baseline. 
While this is common in many benchmarks, HugAgent tasks present unique challenges: they involve \textit{long, naturalistic transcripts} and fine-grained belief trajectories. 
In principle, annotators could be asked to re-read transcripts and label stance updates, but such procedures are slow, error-prone, and risk conflating annotators’ own heuristics with the original participant’s reasoning. 
This creates a fidelity–feasibility tradeoff: while feasible, the outcome would be a proxy of \textit{third-party interpretation}, rather than a faithful measure of the individual’s reasoning process.

Instead, we adopt \textit{test--retest reliability} as the human ceiling. 
Here, the same participant is re-sampled or re-interviewed, and the consistency of their own responses provides a direct measure of reliability. 
This practice is well established in psychology and survey research, and has been adopted in recent large-scale reasoning datasets facing similar challenges~\citep{park2023thousandpeople,toubia2025twin2k500datasetbuildingdigital}. 
Compared to annotator baselines, test--retest reliability offers a more precise and ecologically valid upper bound for model performance, aligned with the benchmark’s goal of capturing intra-individual reasoning fidelity. 
\textbf{This principle defines how we report the human ceiling in HugAgent.}

\newpage

\section{Data Full Example}
User's
Demographic:
\begin{table*}[ht]
\centering
\begin{adjustbox}{max width=\textwidth}
\small
\begin{tabular}{ll}
\toprule
\textbf{Attribute} & \textbf{Example (Anonymized Participant)} \\
\midrule
Housing Experience & Has lived in the same residence for several years \\
Age & 30 \\
Moved Last Year & Same house 1 year ago \\
Housing Status & Owner-occupied \\
Transportation & Car / Truck / Van \\
Household Income & \$75{,}000--\$99{,}999 \\
Occupation & Sales and office occupations \\
Marital Status & Not married \\
Children & Has children \\
Neighborhood Safety & Very safe. I rarely worry about crime \\
Health Insurance & Private insurance, no disability \\
Education & High school graduate or equivalent \\
Citizenship & Native-born U.S. citizen \\
Financial Situation & Gets by, but money is tight \\
\bottomrule
\end{tabular}
\end{adjustbox}
\caption{Example anonymized participant profile used in analysis (for illustration only). Personally identifiable details have been generalized or omitted.}
\label{tab:appendix-example-profile}
\end{table*}

\subsection*{Interview QA}
\addcontentsline{toc}{section}{Interview QA}

\textit{Note: The following excerpt reflects a simulated or anonymized participant’s responses. 
It may contain biased or stereotypical opinions that do not represent the authors’ or dataset creators’ views. 
It is included purely for analysis of belief attribution and reasoning behavior.}

\begin{enumerate}

\item \textbf{Q:} To what extent do you support or oppose upzoning policies that allow for higher density housing in traditionally single-family neighborhoods? Please explain your reasoning.\\
\textbf{A:} I don't support it at all. I'm worried that it'll cause overcrowding if cheaper apartments or housing were made. Aside from that, we know that statistically, lower income people tend to have more of the criminal population in them, isn't that right? So this might cause the crime rates to go up!

\item \textbf{Q:} What do you think are the most significant impacts, positive or negative, of increasing housing density in residential neighborhoods?\\
\textbf{A:} I've mentioned the potential for crime rates to go up, that's the real worry here. Lots of new lower income people, lots of potential criminals.

\item \textbf{Q:} How do you think upzoning policies might affect housing affordability in urban areas?\\
\textbf{A:} They'd most likely lower the price of rent because of ``competition''. But at what cost? The safety of the people!

\item \textbf{Q:} What impact do you believe increased housing density might have on neighborhood character and quality of life?\\
\textbf{A:} Safety for sure. Low income places simply have more potential for crimes due to people being tempted to commit criminal acts for survival.

\item \textbf{Q:} How do you think upzoning might affect transportation systems and traffic congestion in cities?\\
\textbf{A:} It's going to worsen! Look, there was a time when I used to take the bus to get to work every day when I still didn't have a car. I live in a big city and sometimes, the bus couldn't take all of us! That caused me to get late a couple of times since there wasn't even any standing room. So imagine, a rush of new low income people to this area, probably they don't have cars so they'll rely on buses, it'll just be extra strain on the buses and not everyone would be able to get on the bus at all.

\item \textbf{Q:} What role do you believe local government should play in regulating housing development and density?\\
\textbf{A:} The government really shouldn't be too involved with many things. Just minimally involved. Less government involvement, the better.

\item \textbf{Q:} How might environmental concerns factor into decisions about urban density and zoning?\\
\textbf{A:} I don't personally care about these so-called ``environmental concerns''. I'm not some kind of environmental activist or terrified climate change believer. As long as something doesn't dump toxic waste or all sorts of hazardous material in my area, then it's good.

\item \textbf{Q:} What economic effects, both positive and negative, might result from changing zoning laws to allow more multi-family housing?\\
\textbf{A:} More new people, more potential customers for businesses in the area obviously. BUT we also have to think that these are low income people if we're talking about low income housing. So businesses targeting low income people would most likely benefit, but the more upscale ones wouldn't.

\item \textbf{Q:} How do you think the interests of current residents versus future residents should be balanced when making zoning decisions?\\
\textbf{A:} The current residents should ALWAYS be prioritized, they were there first. New people should always be considerate of the people living wherever they're planning to move to. It's just basic human decency.

\item \textbf{Q:} What role do you think social equity and access to opportunity play in discussions about zoning and housing policy?\\
\textbf{A:} I am totally against EQUITY. Equity means taking opportunities away from someone in order to give it to somebody else who probably didn't earn it. I don't like the idea of redistributing what a successful person has.

\item \textbf{Q:} How confident are you that changes in Higher density housing lead to changes in Support for Upzoning? Does it have a positive effect (increasing it) or a negative effect (decreasing it)? How strong is this effect?\\
\textbf{A:} It's going to be NEGATIVE. If we're talking about people, it's not just quantity that we're supposed to worry about, but also the quality. So we can say ``Don't judge a book by their cover'', but we also must think that people are in the situation they are for a reason. So if we're going to get flooded by low income people, we have to ask, ``Why are they low income?'' Of course not all low income people are bad, but majority of criminals are low income people.

\item \textbf{Q:} What factors do you think influence Support for Upzoning, and how strong is their impact? Please also indicate if these influences are positive (increasing) or negative (decreasing).\\
\textbf{A:} Definitely the idea of SAFETY is a huge factor. Just imagine you live in a peaceful neighborhood where crime isn't really a problem, then suddenly a huge number of new low income people flood in to your community and suddenly kids start getting bullied at the playground, people start getting mugged left and right. Safety is really a big concern!

\item \textbf{Q:} Does Crime rates have a positive or negative effect on Support for Upzoning, and how significant is this effect? Does it have a positive effect (increasing it) or a negative effect (decreasing it)? How strong is this effect?\\
\textbf{A:} That's what I've been talking about this entire conversation, the potential for CRIME! As I've already stated numerous times, it's a MAJOR concern and an influx of low income people would definitely affect the crime rate!

\item \textbf{Q:} Would small changes in Housing affordability lead to noticeable changes in Support for Upzoning, or would it take larger shifts? Does it have a positive effect (increasing it) or a negative effect (decreasing it)? How strong is this effect?\\
\textbf{A:} At first people would probably think things will be better because rent might go down a bit, BUT that's not guaranteed. Second, SAFETY is really something that people are probably not willing to compromise.

\item \textbf{Q:} Would small changes in Safety lead to noticeable changes in Support for Upzoning, or would it take larger shifts? Does it have a positive effect (increasing it) or a negative effect (decreasing it)? How strong is this effect?\\
\textbf{A:} If there's really no way about avoiding the creation of some kind of tall low income apartment building for the sake of ``equity'', then the next best thing would be to thoroughly do background checks on all the renters. For example, there should be strictly nobody in there with a criminal record.

\item \textbf{Q:} How would you describe the relationship between Low Income People and Support for Upzoning? Is it a strong or weak connection? Does it have a positive effect (increasing it) or a negative effect (decreasing it)? How strong is this effect?\\
\textbf{A:} Well of course low income people would support the creation of low income rental building. But the problem is that people already living in the community, like me, wouldn't support it at all for fears of safety worsening.

\item \textbf{Q:} Is the effect of Minimal Regulation on Support for Upzoning immediate, or does it take time to develop? Does it have a positive effect (increasing it) or a negative effect (decreasing it)? How strong is this effect?\\
\textbf{A:} Any policy takes TIME to develop. Rushed policies just end up in disaster because it won't be well thought out.

\item \textbf{Q:} Would small changes in Impact on businesses lead to noticeable changes in Support for Upzoning, or would it take larger shifts? Does it have a positive effect (increasing it) or a negative effect (decreasing it)? How strong is this effect?\\
\textbf{A:} No. As I've said, low income people will only provide benefit to businesses targeting low income customers. Mid to upscale businesses wouldn't benefit from them because they won't be able to afford their products and services. In short, not all businesses would be in support of having some kind of low income housing in the area if all they're going to be able to afford are low income stuff.

\item \textbf{Q:} How would you describe the relationship between Basic human decency and Support for Upzoning? Is it a strong or weak connection? Does it have a positive effect (increasing it) or a negative effect (decreasing it)? How strong is this effect?\\
\textbf{A:} There are people who make decisions based on feelings alone. Yes, they'll think it's ``decent'' to allow low income people to have low income housing in their community, BUT often, these people don't think about the consequences that would affect the people already living in the community. They are too focused on helping others that they don't realize they are causing harm to themselves.

\item \textbf{Q:} Is the effect of Redistribution on Support for Upzoning immediate, or does it take time to develop? Does it have a positive effect (increasing it) or a negative effect (decreasing it)? How strong is this effect?\\
\textbf{A:} That's definitely going to be a huge NEGATIVE right away. Nobody in their right mind would want themselves to be compromised for others. So let's think about what happens if in a moderately wealthy area, they allowed low income housing in the name of ``equity''. For actual home owners (not renters), the value of their properties would go down. These are properties that they've worked for years to maintain, and suddenly, in the name of ``equity'', is it alright to allow the values to go down? No of course not! So we have to always think about how low income housing would affect the people already living in the community.

\item \textbf{Q:} Would small changes in Negative Effect lead to noticeable changes in Support for Upzoning, or would it take larger shifts? Does it have a positive effect (increasing it) or a negative effect (decreasing it)? How strong is this effect?\\
\textbf{A:} No, it's called ``Negative Effect'' because it affects people in a bad way. Nobody would support anything like that knowingly.

\item \textbf{Q:} What factors affect Upzoning policies, and which ones have the strongest influence? Please also indicate if these influences are positive (increasing) or negative (decreasing).\\
\textbf{A:} As I've been saying this entire conversation, the major facor that affects people's support for low income housing is the SAFETY, the potential for crime rates to go up, and these things will definitely always affect support for low income housing negatively.

\item \textbf{Q:} Does Community Resistance to Upzoning have a positive or negative effect on Support for Upzoning, and how significant is this effect? Does it have a positive effect (increasing it) or a negative effect (decreasing it)? How strong is this effect?\\
\textbf{A:} Of course community resistance won't support low income housing, that's the point. People would resist these places from being built in order to protect the community from potential safety concerns.

\item \textbf{Q:} How would you describe the relationship between Time for policy development and Support for Upzoning? Is it a strong or weak connection? Does it have a positive effect (increasing it) or a negative effect (decreasing it)? How strong is this effect?\\
\textbf{A:} Of course ``Time'' will always have something to do with whether low income housing would be allowed or not. For example, maybe a politician would take his time forming some kind of bill concerning low income housing and he'll wait for enough public support before officially launching it in order to increase its chances of succeeding.

\item \textbf{Q:} Would small changes in Low Income Housing lead to noticeable changes in Support for Upzoning, or would it take larger shifts? Does it have a positive effect (increasing it) or a negative effect (decreasing it)? How strong is this effect?\\
\textbf{A:} No, even small changes in low income housing won't change people's support for it because it will negatively affect the community. People already know it's most likely going to be the cause of many safety concerns aside from property devaluation.

\item \textbf{Q:} Does Equity have a positive or negative effect on Support for Upzoning, and how significant is this effect? Does it have a positive effect (increasing it) or a negative effect (decreasing it)? How strong is this effect?\\
\textbf{A:} Equity has NEGATIVE effects on people already living in the community, because the point of equity is to take from those people (land space) and to redistribute it to other people (the low income people). People might try to frame it as ``helping the poor'', but you can help poor people in other ways without harming the community.

\item \textbf{Q:} How would you describe the relationship between Support for low income housing and Support for Upzoning? Is it a strong or weak connection? Does it have a positive effect (increasing it) or a negative effect (decreasing it)? How strong is this effect?\\
\textbf{A:} People are directly against low income housing because it's more likely to bring bad stuff with it that good ones. The consequences outweigh the positives.

\end{enumerate}

\subsection*{Sample Survey: Housing / Upzoning}

\subsubsection*{Baseline Stance}
\begin{itemize}
    \item Q1. On a scale from 1 to 10, how much do you support or oppose allowing bigger, taller apartment buildings in your neighborhood?  (Scale 1 to 10. 1 means strongly oppose and 10 means strongly support.)
\end{itemize}

\subsubsection*{Reason Evaluation (Baseline)}
Q1r. How much do the following reasons influence your general opinion on upzoning? (Scale 1 to 10. 1 means strongly oppose and 10 means strongly support.)

\subsubsection*{Scenario 1: Rent Drop}
\begin{itemize}
    \item Q2. After the city allows more apartments in low-density areas, rent prices drop 10--15\%. Your monthly rent is noticeably lower. It’s easier to find a decent place. How would this affect your stance?   (Scale 1 to 10. 1 means strongly oppose and 10 means strongly support.)
    
\end{itemize}

\subsubsection*{Reason Evaluation (Scenario 1)}
Q2r. To what extent do the following reasons influence your stance?  
\begin{center}
\begin{tabular}{p{0.6\linewidth} c}
\toprule
Reason & Scale (1--5) \\
\midrule
Building more homes helps with the housing crisis. (A) & \\
More traffic and parking is a real concern. (D) & \\
Everyone should help handle the growth. (N) & \\
More people means more business for local shops. (O) & \\
I’d worry about noise and crowding on my block. (P) & \\
\bottomrule
\end{tabular}
\end{center}


\newpage
\section{Rationale for Individual Cross-Domain Transfer}
We assume that cross-domain personalization is feasible because individuals express stable, value-laden cues throughout natural language. These cues are not tied to a single domain; rather, they reflect underlying principles that consistently shape preferences across contexts.
\label{app:cross-domain claim}
\paragraph{Quantitative Evidence: Consistent In- to Out-of-Domain Transfer}

To empirically support this assumption, we report cross-domain transfer results for four individuals. For each person, GPT-4o was evaluated across five independent runs. Across all individuals and all runs, we observe the same ordering:
$\text{In-domain accuracy} \; > \; \text{Cross-domain accuracy} \;$
$> \; \text{No-context accuracy}.$

This strict ordering across all 20 settings (4 individuals $\times$ 5 runs) provides a statistically grounded indication that the observed cross-domain transfer is reliable and not an artifact of noise.

\begin{table*}[ht]
    \centering
    \begin{adjustbox}{max totalsize={\textwidth}{\textheight},keepaspectratio}
    \begin{tabular}{l l c c c c c}
        \toprule
        \multirow{2}{*}{\textbf{User}} 
        & \multirow{2}{*}{\textbf{Experiment}} 
        & \textbf{Belief State Inference} 
        & \multicolumn{3}{c}{\textbf{Belief Dynamics Update (BDU)}} 
        & \multirow{2}{*}{\textbf{ATI (\% ↑)}} \\
        \cmidrule(lr){3-3}
        \cmidrule(lr){4-6}
        & & Acc. (\% ↑)
        & Acc. (\% ↑)
        & MAE (↓)
        & Dir. Acc. (\% ↑)
        & \\
        \midrule
        
        \multirow{3}{*}{\shortstack{User 1\\\scriptsize (n-BSI = 9,\; n-BDU = 48)}}
        & In Domain
        & 88.89$^{\pm0.01}$
        & 91.67$^{\pm0.01}$ & 0.57$^{\pm0.02}$ & 85.00$^{\pm0.01}$
        & 87.88$^{\pm0.05}$ \\
        & Cross Domain
        & 55.56$^{\pm0.01}$
        & 54.58$^{\pm3.33}$ & 1.38$^{\pm0.03}$ & 50.00$^{\pm0.01}$
        & 55.28$^{\pm0.51}$ \\
        & No Context
        & 33.33$^{\pm0.01}$
        & 8.33$^{\pm0.01}$ & 2.33$^{\pm0.02}$ & 18.00$^{\pm3.67}$
        & 27.42$^{\pm0.91}$ \\
        \midrule
        
        \multirow{3}{*}{\shortstack{User 2\\\scriptsize (n-BSI = 13,\; n-BDU = 41)}}
        & In Domain
        & 92.31$^{\pm0.01}$
        & 91.71$^{\pm1.95}$ & 0.58$^{\pm0.02}$ & 65.00$^{\pm0.01}$
        & 84.57$^{\pm0.30}$ \\
        & Cross Domain
        & 92.31$^{\pm0.01}$
        & 83.90$^{\pm1.19}$ & 0.97$^{\pm0.02}$ & 73.00$^{\pm9.80}$
        & 84.37$^{\pm2.49}$ \\
        & No Context
        & 30.77$^{\pm9.73}$
        & 39.02$^{\pm3.45}$ & 1.68$^{\pm0.04}$ & 65.00$^{\pm0.01}$
        & 43.78$^{\pm4.51}$ \\
        \midrule
        
        \multirow{3}{*}{\shortstack{User 3\\\scriptsize (n-BSI = 5,\; n-BDU = 24)}}
        & In Domain
        & 66.67$^{\pm0.01}$
        & 64.23$^{\pm2.48}$ & 1.26$^{\pm0.04}$ & 57.50$^{\pm0.01}$
        & 64.29$^{\pm0.38}$ \\
        & Cross Domain
        & 66.67$^{\pm0.01}$
        & 28.57$^{\pm1.88}$ & 1.87$^{\pm0.04}$ & 65.00$^{\pm0.01}$
        & 59.80$^{\pm0.34}$ \\
        & No Context
        & 46.67$^{\pm9.79}$
        & 19.22$^{\pm1.21}$ & 1.95$^{\pm0.03}$ & 50.00$^{\pm0.01}$
        & 44.63$^{\pm5.05}$ \\
        \midrule

        \multirow{3}{*}{\shortstack{User 4\\\scriptsize (n-BSI = 7,\; n-BDU = 38)}}
        & In Domain
        & 57.14$^{\pm0.01}$
        & 42.11$^{\pm3.33}$ & 1.72$^{\pm0.04}$ & 92.50$^{\pm0.01}$
        & 64.10$^{\pm0.52}$ \\
        & Cross Domain
        & 51.43$^{\pm0.69}$
        & 30.00$^{\pm2.11}$ & 1.79$^{\pm0.02}$ & 85.00$^{\pm0.01}$
        & 57.63$^{\pm3.47}$ \\
        & No Context
        & 31.43$^{\pm5.71}$
        & 30.00$^{\pm1.29}$ & 1.79$^{\pm0.03}$ & 77.50$^{\pm0.01}$
        & 45.73$^{\pm2.76}$ \\
        \midrule
        
    \end{tabular}
    \end{adjustbox}
    \vspace{-0.5em}
   \caption{
Comparison of GPT-4o performance across In-domain, Cross-domain, and No-context 
conditions for four representative users (mean~$\pm$~std over five runs). 
}

\end{table*}

\paragraph{Qualitative Evidence: Stable Value Dimensions Across Domains}

To complement the quantitative results, we provide a qualitative analysis of one participant’s transcript. The individual expresses a coherent set of value-laden principles that appear across healthcare, surveillance, and zoning. These principles naturally support cross-domain generalization from a single personalized conversation.

The following quotations reflect individual participant beliefs and not normative statements endorsed by the authors.

\begin{enumerate}
    \item \textbf{A safety-first orientation that generalizes across contexts.}  \\
    In zoning, the participant frames upzoning as a threat to public safety:
    \begin{quote}
    ``If cheaper apartments were made, it’ll cause overcrowding, and that means more low-income people moving in… which will make crime rates go up!''
    \end{quote}
    In surveillance, the same concern motivates strong support for camera deployment:
    \begin{quote}
    ``Cameras are a tool of preventing potential crime and it’s effective.''
    \end{quote}
    A safety-oriented value extracted from zoning thus directly predicts pro-surveillance attitudes.

    \item \textbf{A stable opposition to redistribution across domains.} \\ 
    In healthcare:
    \begin{quote}
    ``Nobody is entitled to other people’s money. Taxpayers shouldn’t be forced to pay for other people’s medical needs.''
    \end{quote}
    In zoning:
    \begin{quote}
    ``Equity means taking opportunities away from some to redistribute them to others. I don’t like redistributing what a successful person has.''
    \end{quote}
    The same anti-redistribution stance explains resistance to equity-oriented policies in both domains.

    \item \textbf{A consistent negative framing of low-income groups as a societal risk.}  \\
    In zoning:
    \begin{quote}
    ``Lots of new low-income people means lots of potential criminals… majority of criminals are low income.''
    \end{quote}
    In healthcare:
    \begin{quote}
    ``The only people who’d support universal healthcare are those who can’t afford healthcare themselves.''
    \end{quote}
    This stable attribution shapes preferences across otherwise unrelated policy areas.

    \item \textbf{Selective distrust of government intervention—except in policing.}  \\
    In healthcare:
    \begin{quote}
    ``If the government runs it, everything becomes standardized and we’re forced to pay for others.''
    \end{quote}
    In zoning:
    \begin{quote}
    ``The government really shouldn’t be too involved… less government involvement, the better.''
    \end{quote}
    Yet in surveillance:
    \begin{quote}
    ``If police use cameras to catch criminals, support will grow.''
    \end{quote}
    This produces a predictable cross-domain pattern: skepticism toward government redistribution and regulation, but support for expanded government authority in security contexts.
\end{enumerate}

\paragraph{Summary}

These four dimensions—(1) safety orientation, (2) redistribution aversion, (3) negative out-group attribution, and (4) selective distrust of government—are expressed consistently across the participant’s interview. Because such values manifest across multiple domains, personalized signals extracted from a single in-depth conversation (e.g., a two-hour interview) naturally generalize to related policy contexts.

Taken together, the quantitative seed-level consistency and qualitative evidence provide a clear explanation for why individual cross-domain personalization is expected and why our empirical findings are robust.

\clearpage
\section{Quality-Control Protocol}
\label{app:qc-protocol}

We applied a standardized protocol to ensure that only participants with reliable and reproducible data were retained. 
The following criteria were applied sequentially:

\begin{enumerate}
    \item \textbf{Redundant responses}: cases where the participant repeatedly 
    produced near-identical statements without substantive variation. 
    
    \item \textbf{Meta-level questioning}: transcripts dominated by repeated 
    challenges to the validity of the task itself rather than substantive reasoning 
    about the topic. 
    
    \item \textbf{Insufficient length}: responses falling below a minimum threshold 
    of tokens or turns, preventing meaningful inference of reasoning structure. 
    
    \item \textbf{Sparse causal belief networks}: chatbot elicitation yielding fewer 
    than five unique nodes, limiting the interpretability of downstream causal graph 
    construction.
\end{enumerate}

This filtering ensured that the retained dataset reflects consistent engagement 
with the task, while minimizing artifacts that could compromise the validity of 
subsequent analyses.

\newpage


\section{Benchmark Task Structure}
\label{app:benchmark-structure}

\begin{figure*}[ht]
    \centering
    \includegraphics[width=0.85\linewidth]{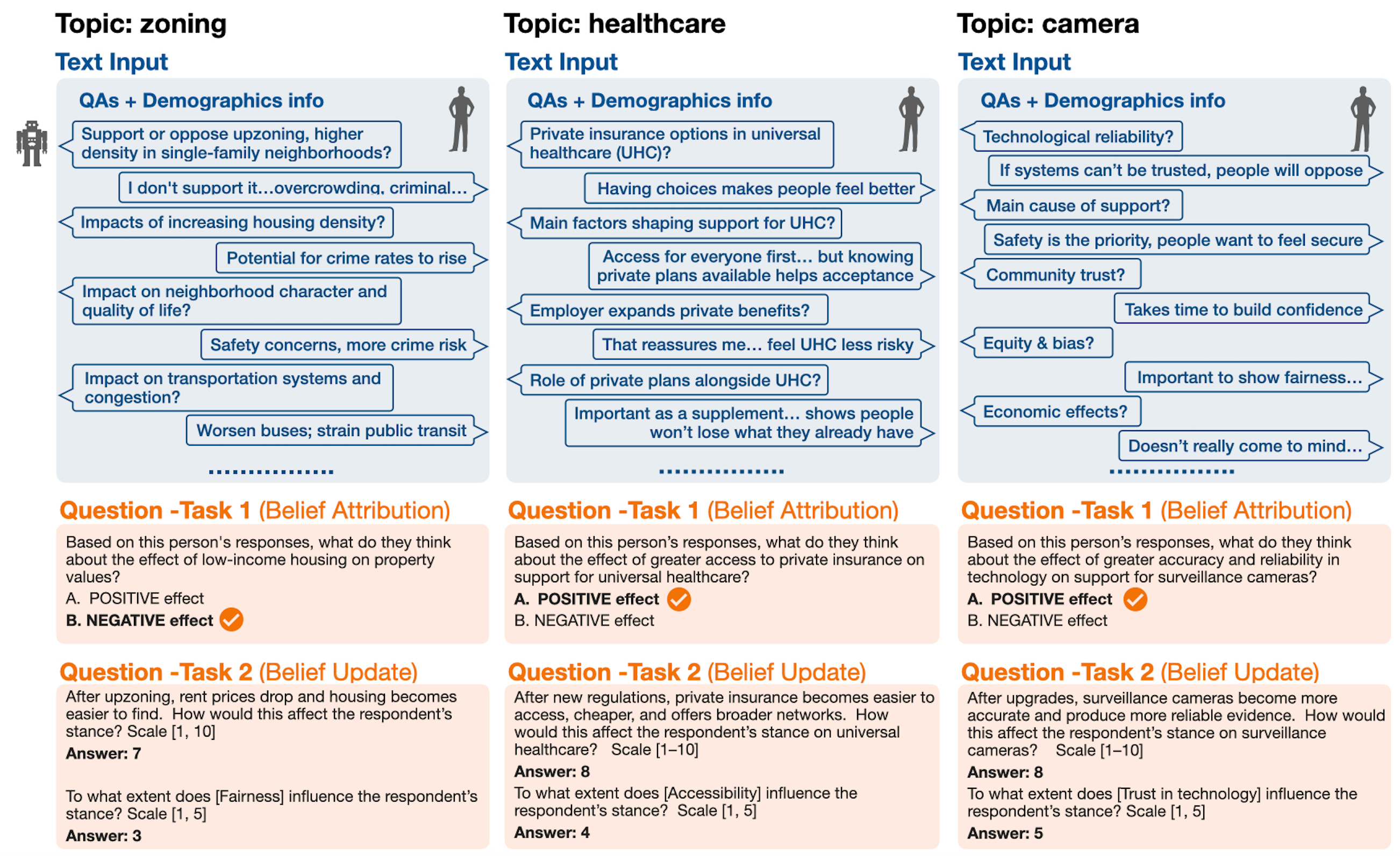}
    \caption{Overview of the HugAgent benchmark structure. Inputs (demographics, questionnaires, and transcripts) are mapped to outputs, including \textbf{belief state inference} (Task~1) and \textbf{belief dynamics update} (Task~2).}
    \label{fig:benchmark-structure}
\end{figure*}

To clarify how HugAgent maps input materials to evaluation tasks, 
we provide here a consolidated overview of the benchmark structure. 
As shown in Figure~\ref{fig:benchmark-structure}, 
raw inputs include (i) demographic profiles, (ii) structured questionnaires, 
and (iii) open-ended chatbot transcripts. 
These inputs are transformed into two core task families: 

\begin{itemize}
    \item \textbf{Task~1: Belief State Inference.}  
    Given a participant’s responses and contextual cues, models must infer the person’s stance and factor-level attribution.  
    Example questions include: ``Does the respondent view low-income housing as a positive or negative effect on property values?’’

    \item \textbf{Task~2: Belief Dynamics Update.}  
    After an intervention (e.g., rent decrease, policy change, technological improvement), models must predict both the stance shift (1–10 scale) and the reweighting of reasons (1–5 scale).  
    Example questions include: ``How would a 10\% reduction in rents affect the respondent’s stance on upzoning?’’
\end{itemize}

Each topic domain—\emph{zoning}, \emph{healthcare}, and \emph{surveillance}—is instantiated with multiple scenarios 
and corresponding reason mappings.  
This ensures comparability across domains while preserving topic-specific ecological validity.  

\clearpage
\newpage

\section{User Journey and Use Cases of \textsc{Trace-Your-Thinking} (A Semi-structured Chatbot Eliciting Human Reasoning)}
\label{app:trace-your-thinking}

This appendix provides a detailed user guide and representative use cases for \textsc{Trace-Your-Thinking}, our semi-structured chatbot system designed to elicit human reasoning at scale. We describe both participant-facing (user) and researcher-facing (admin) views, followed by system outputs and illustrative use cases. We use open science practices as an example here. Our design emphasizes three goals: (i) lowering barriers for participants, (ii) giving researchers flexible and reliable control, and (iii) producing structured outputs that make reasoning analyzable at scale.

\begin{figure}[h]
    \centering
    \includegraphics[width=\linewidth]{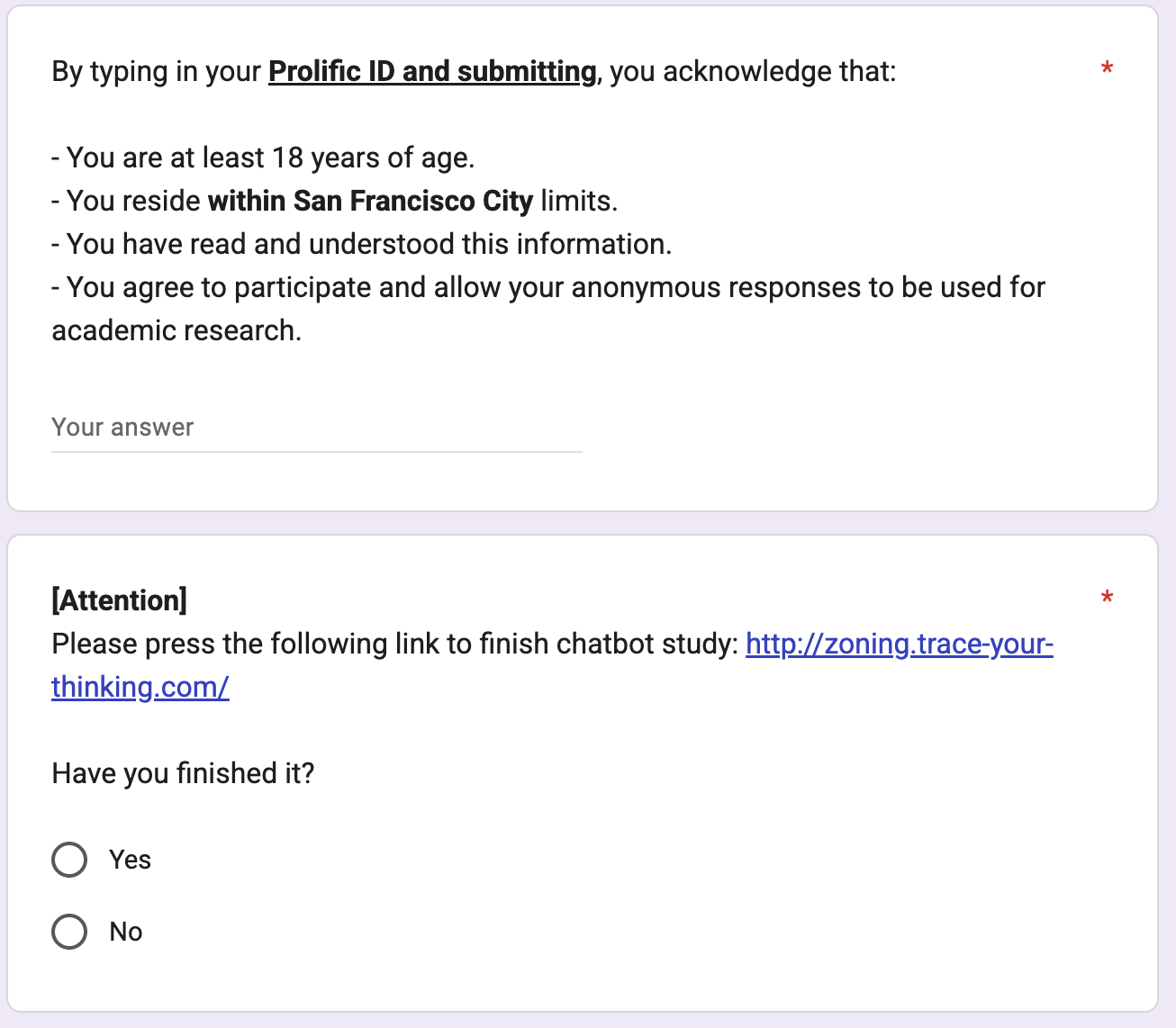}
    \caption{Participant view of the onboarding and interview flow (Consent)}
    \label{fig:user1}
\end{figure}

\begin{figure}[h]
    \centering
    \includegraphics[width=0.6\linewidth]{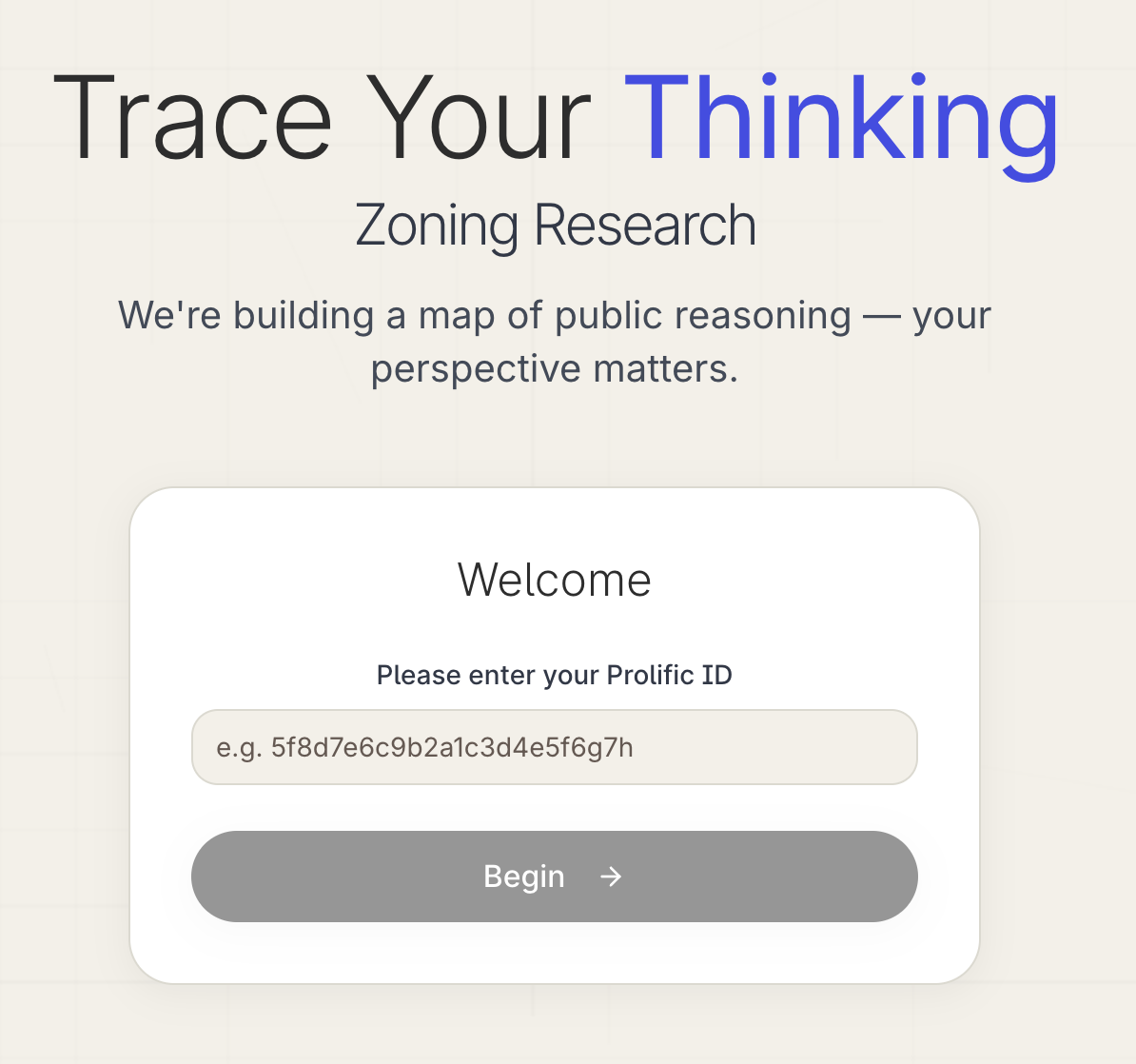}
    \caption{Participant view of the welcome page}
    \label{fig:user2}
\end{figure}

\subsection*{Participant Journey (User View)}

Participants experience a streamlined workflow that reduces friction while maximizing the richness of collected reasoning. 

\paragraph{Step 1: Consent and ID submission.}  
Recruitment begins on the Prolific platform, where participants are shown eligibility criteria and compensation details. Upon accepting the study, they are redirected to a Google Form where they confirm basic requirements (age $\geq 18$, residence within a specified region, consent for anonymous data usage). Entering their Prolific ID links the responses to the recruitment system, enabling follow-ups without storing personal identifiers, as is shown in Figure \ref{fig:user1}.

\begin{figure}[ht]
    \centering
    \includegraphics[width=\linewidth]{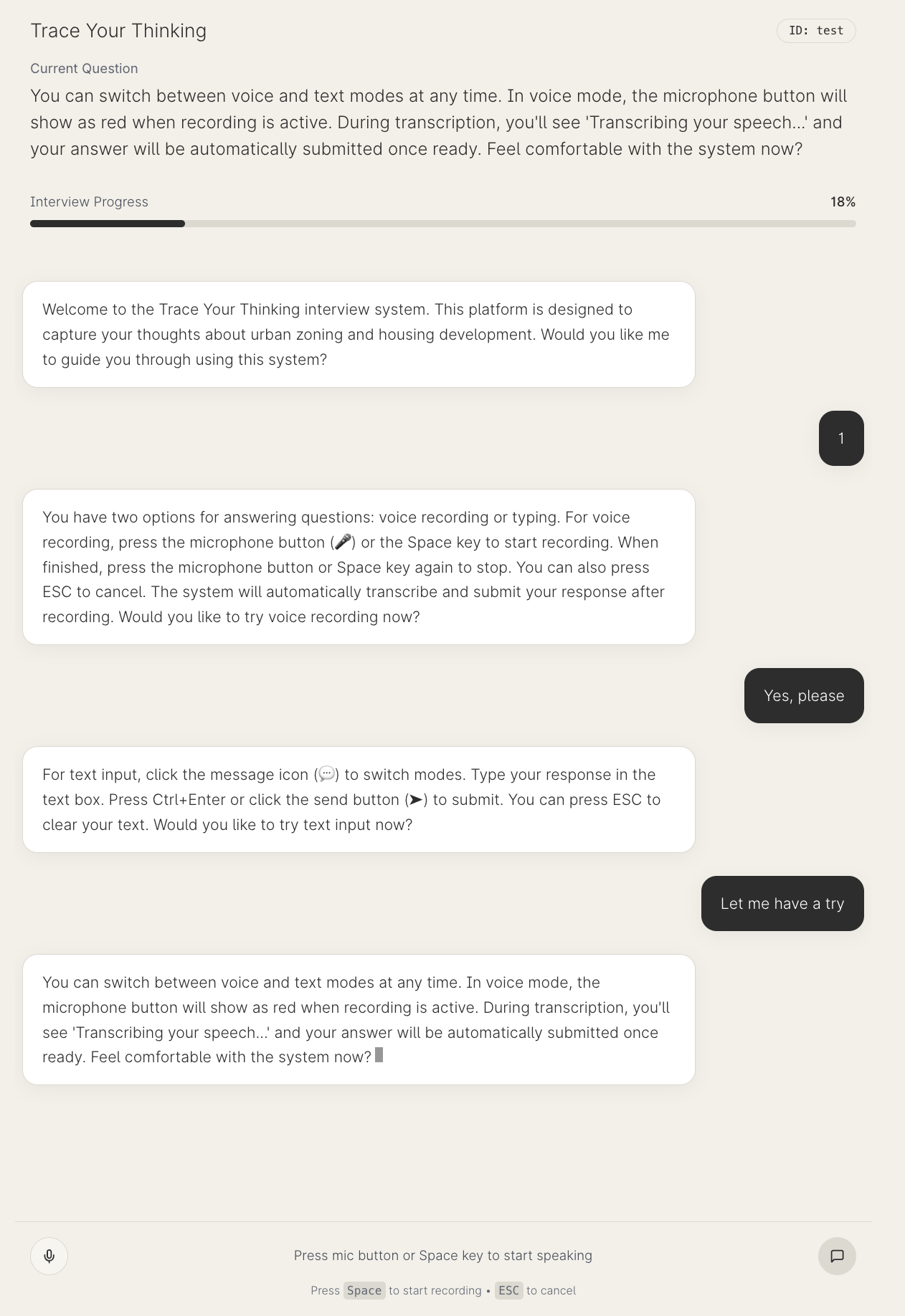}
    \caption{Participant view of the onboarding of the chatbot, including audio and text input}
    \label{fig:user4}
\end{figure}

\begin{figure}[ht]
    \centering
    \includegraphics[width=\linewidth]{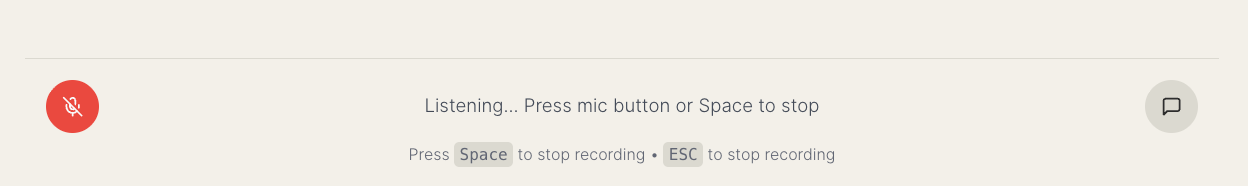}
    \caption{Participant view of using audio to give the answer}
    \label{fig:user3}
\end{figure}

\paragraph{Step 2: Login and onboarding.}  
Participants are then redirected to the \textsc{Trace-Your-Thinking} website. After inputting their Prolific ID, they are guided through a short tutorial. This tutorial introduces input modalities (typed text vs.\ voice-to-text) and explicitly informs users that all answers can be revised either immediately or retrospectively. A persistent progress bar at the top of the interface communicates task completion, reducing dropout risk by making expectations transparent. This part is shown in Figure \ref{fig:user2}.

\begin{figure}[ht]
    \centering
    \includegraphics[width=\linewidth]{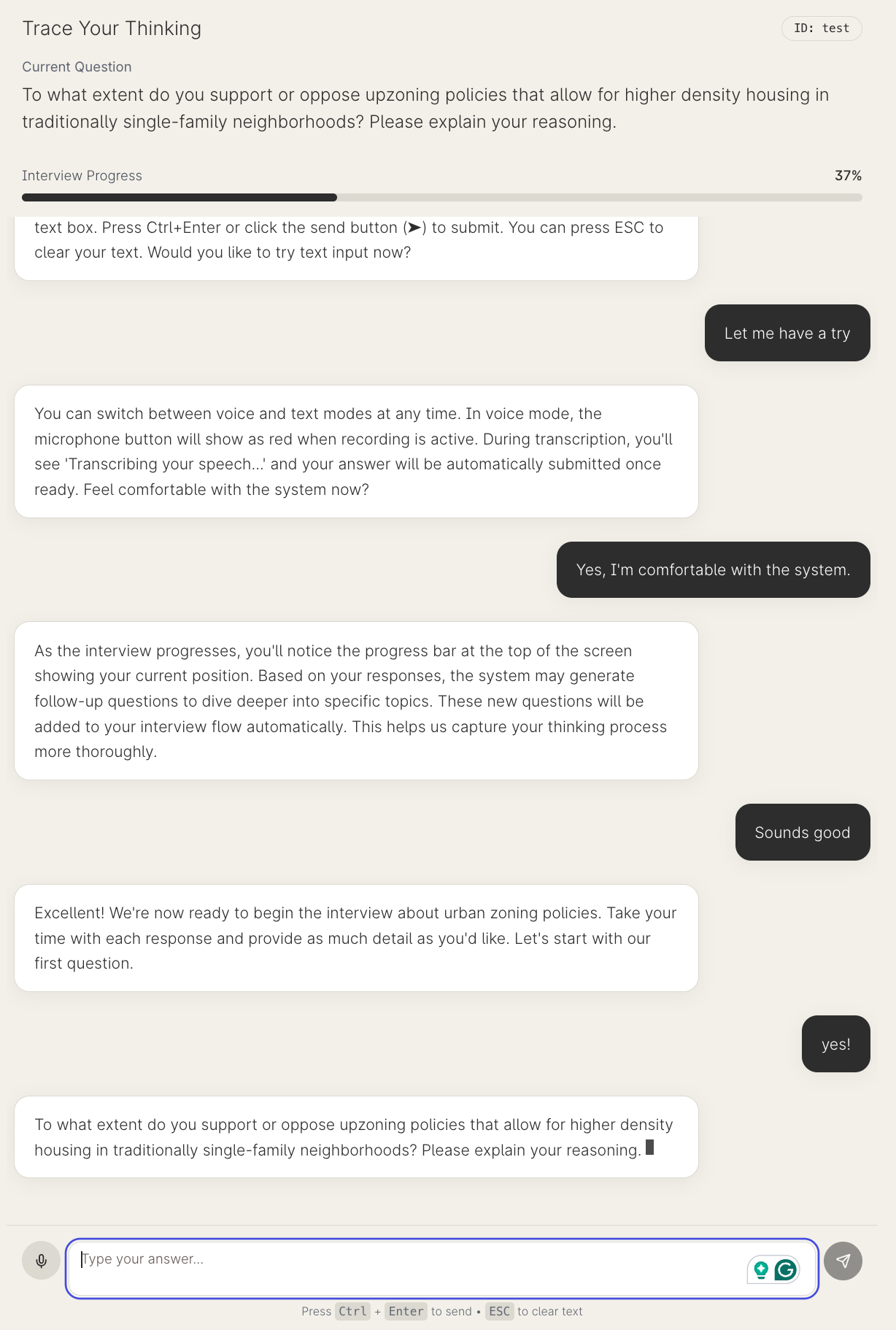}
    \caption{Participant view of answering questions and the interview progress}
    \label{fig:user6}
\end{figure}

\begin{figure}[ht]
    \centering
    \includegraphics[width=\linewidth]{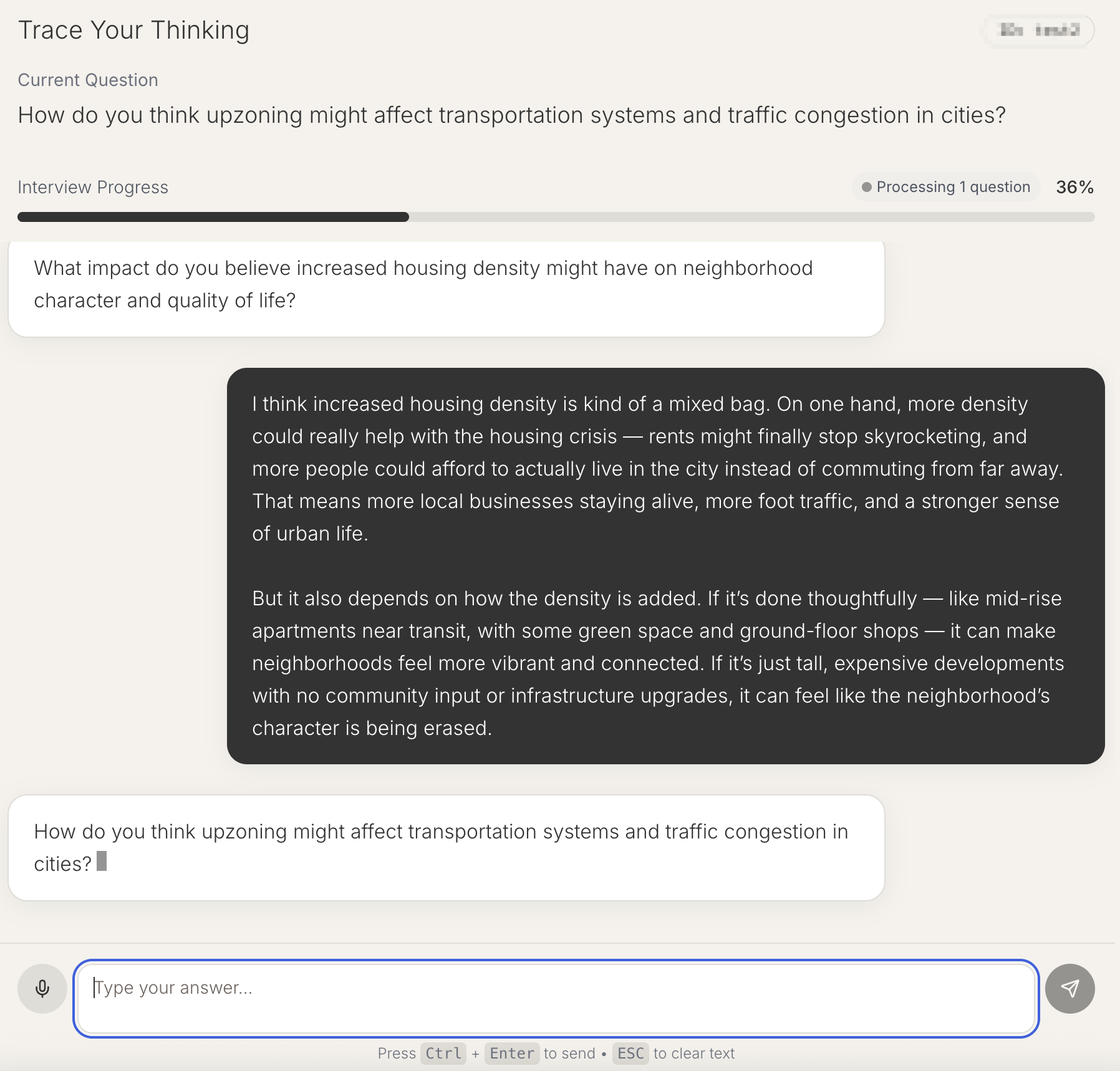}
    \caption{Participant view while AI is processing questions, but the participant can still answer the following questions without waiting}
    \label{fig:user9}
\end{figure}

\paragraph{Step 3: Semi-structured interview.}  
The core of the participant journey is the semi-structured interview, which unfolds in a guided yet flexible flow: introduction $\rightarrow$ guiding questions $\rightarrow$ follow-up probes $\rightarrow$ final review. This design balances standardization with open-ended flexibility, allowing for wide variation in content, style, and depth of responses. 

Participants can choose between typed responses and spoken input, enabling a think-aloud protocol that captures more spontaneous reasoning processes. Figure~\ref{fig:user4} illustrates the onboarding screen where both input modes are explained, while Figure~\ref{fig:user3} shows a participant actively using voice input to answer a question. Once the interview begins, participants can monitor their progress via a persistent progress bar (Figure~\ref{fig:user6}), which reduces fatigue by making task completion transparent. During processing, the interface will show the processing status while still generating new questions and allow users to answer (Figure~\ref{fig:user9}).

\begin{figure}[ht]
    \centering
    \includegraphics[width=\linewidth]{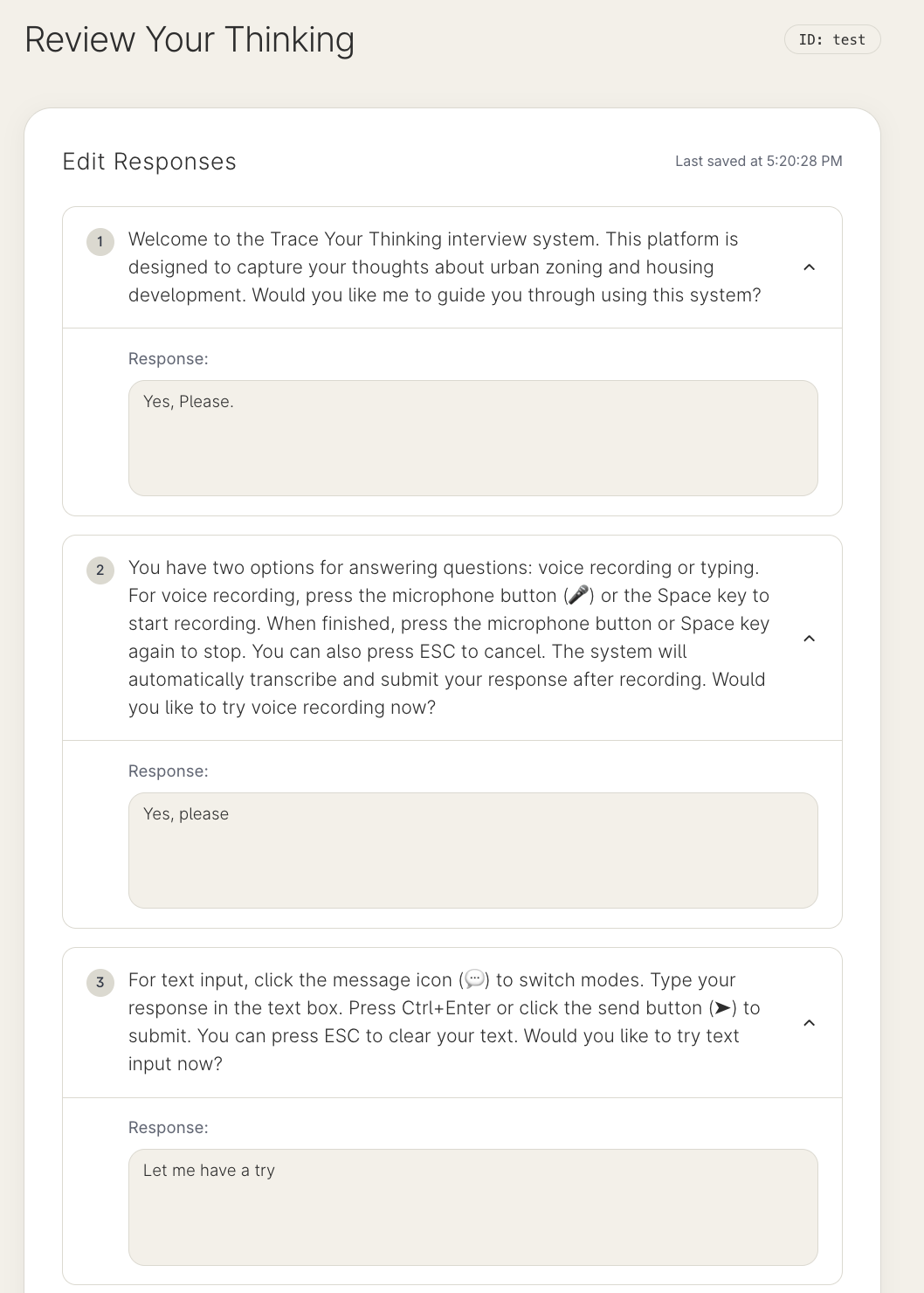}
    \caption{Participant view of the overview of the questions and answers. One can edit and go back to any of their answer.}
    \label{fig:user10}
\end{figure}

\begin{figure}[ht]
    \centering
    \includegraphics[width=\linewidth]{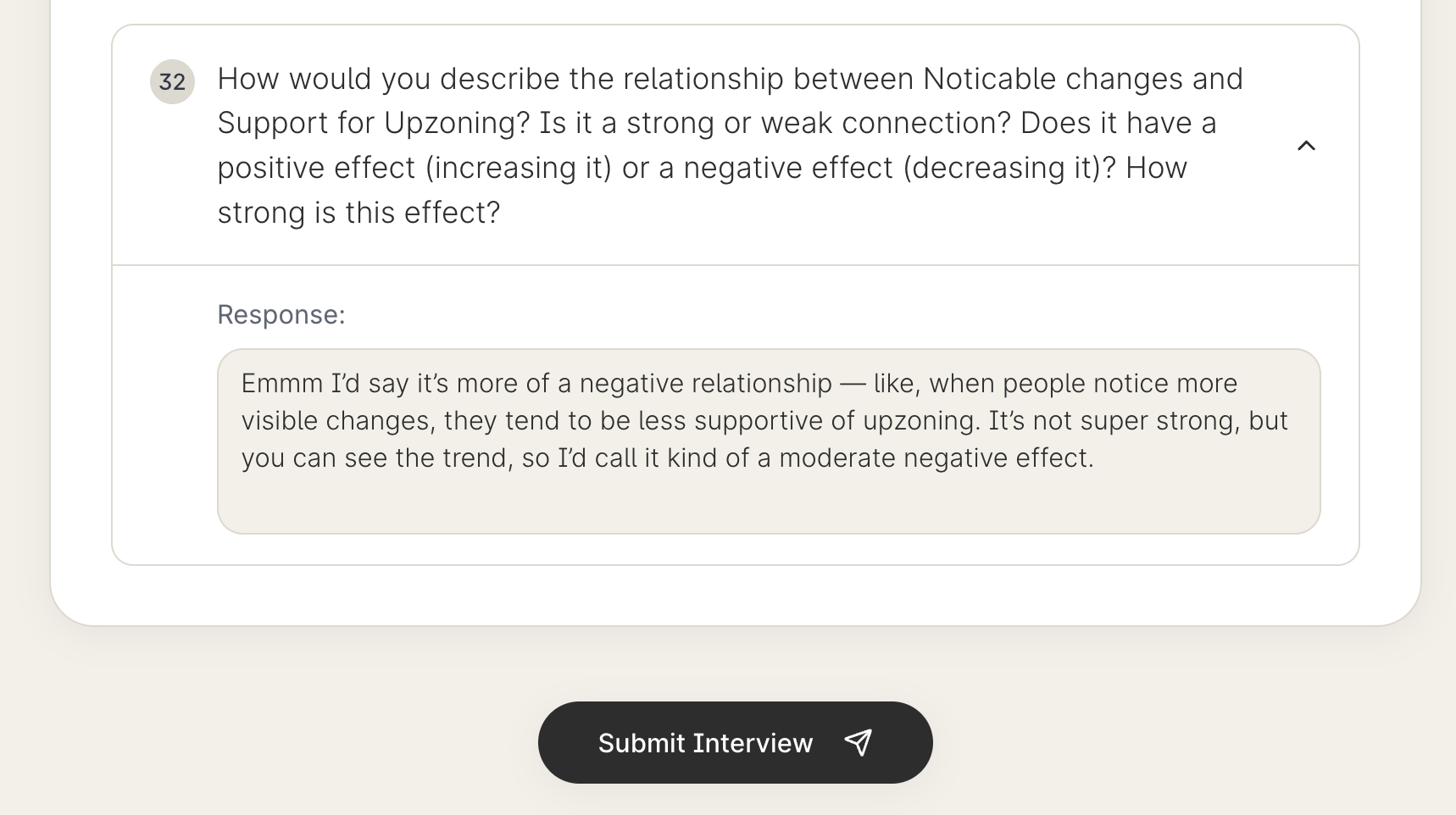}
    \caption{Participant view of the end of the overview of the questions and answers}
    \label{fig:user11}
\end{figure}

Participants can not only edit their responses immediately but also review the entire transcript at the end of the interview. As shown in Figures~\ref{fig:user10} and \ref{fig:user11}, the system presents an overview of all questions and answers, enabling users to backtrack, refine, and self-correct their reasoning. This mirrors how real-world reasoning often evolves over multiple passes rather than being fixed in a single draft.

\begin{figure}[h!]
    \centering
    \includegraphics[width=\linewidth]{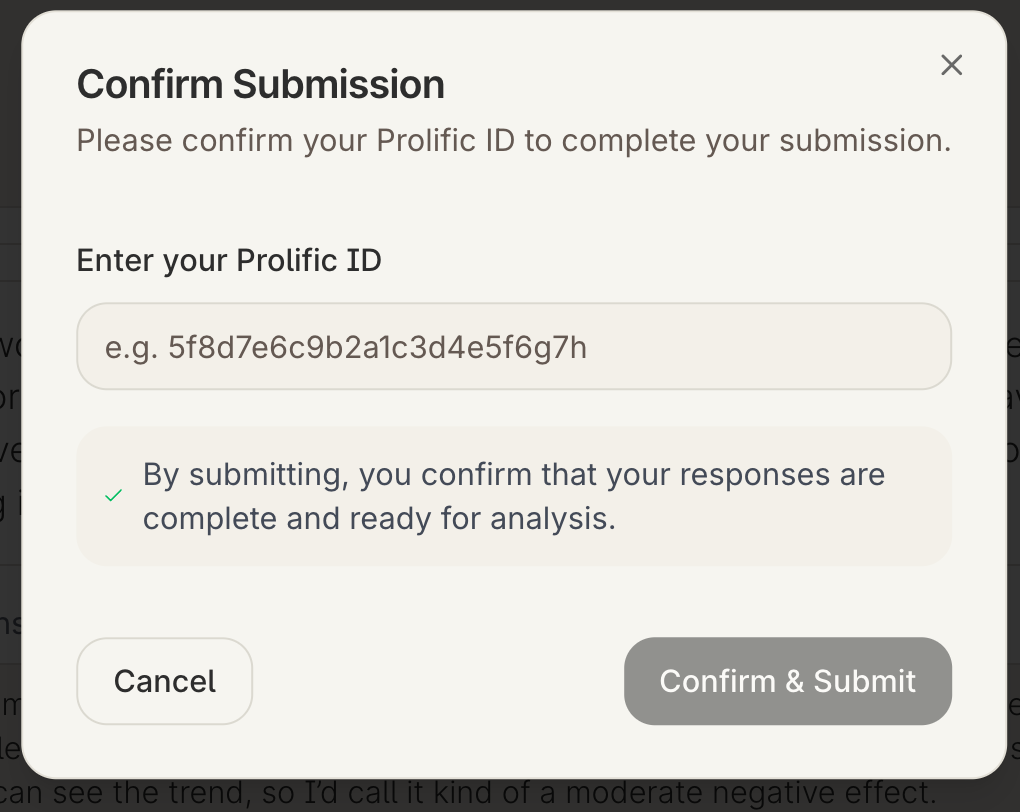}
    \caption{Participant view of reentering prolific id for submission}
    \label{fig:user12}
\end{figure}

\begin{figure}[h!]
    \centering
    \includegraphics[width=\linewidth]{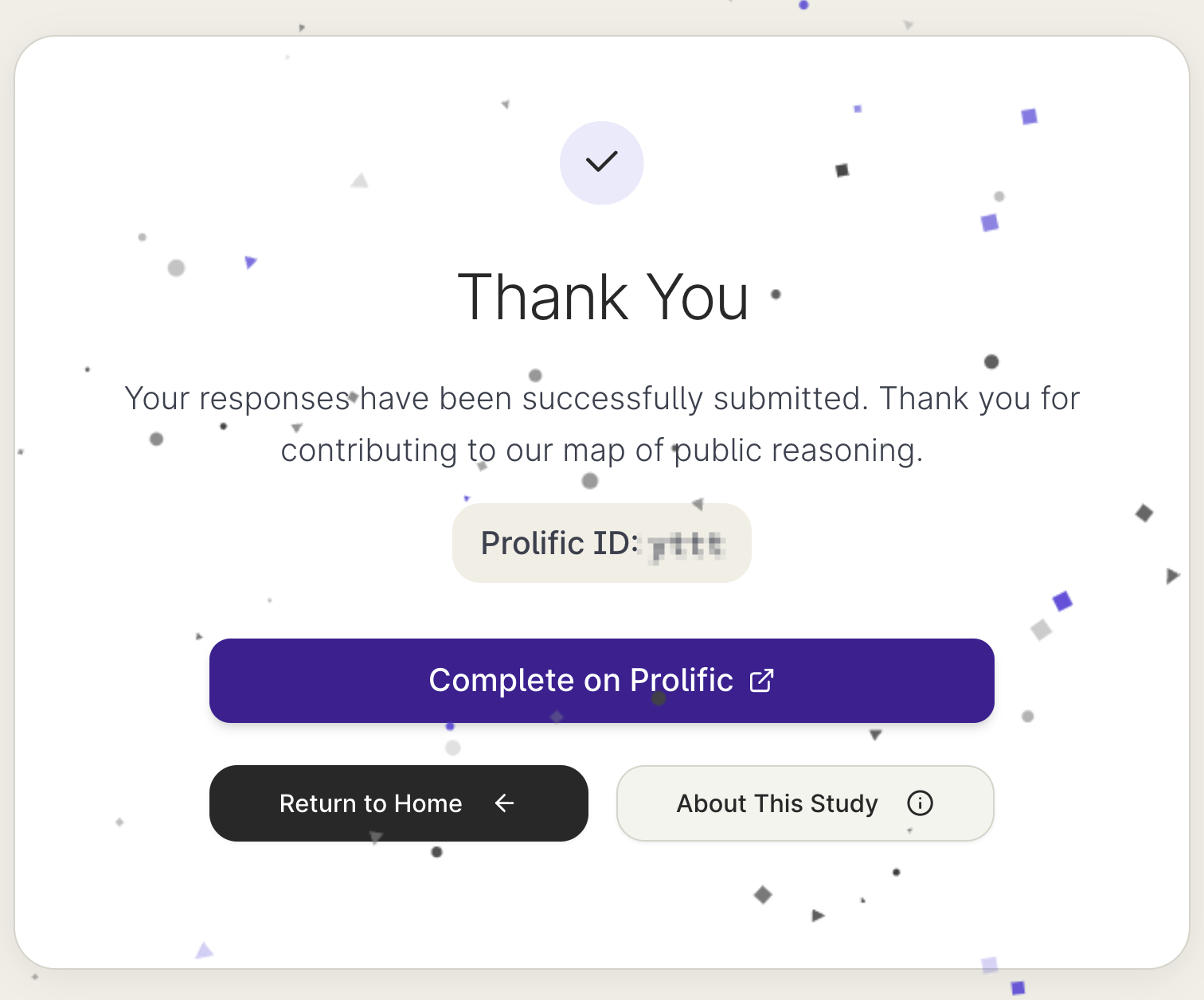}
    \caption{Participant view of the end of the test}
    \label{fig:user13}
\end{figure}

\paragraph{Step 4: Submission and compensation.}  
Once satisfied, participants submit their responses. Figures~\ref{fig:user12} and \ref{fig:user13} demonstrate the submission stage, where participants re-enter their Prolific ID to confirm completion and finalize their session. The system redirects them back to Prolific, which automatically verifies completion and issues compensation. This tight integration ensures high-quality participation while minimizing administrative overhead.

\subsection{Researcher Journey (Admin View)}

The system provides a dedicated control panel that makes data collection transparent, configurable, and scalable. Unlike static survey platforms, admins can adapt the study design on the fly and extract structured reasoning outputs.

\paragraph{Recruitment integration.}  
Admins can publish tasks directly on Prolific, embedding the study link into recruitment posts. Prolific’s filters (approval rate, demographics, geography) allow targeted participant pools, while stored Prolific IDs support longitudinal follow-ups. This design enables researchers to re-engage the same individuals across time or across topics, making it uniquely suitable for longitudinal reasoning studies.

\begin{figure}[ht]
    \centering
    \includegraphics[width=\linewidth]{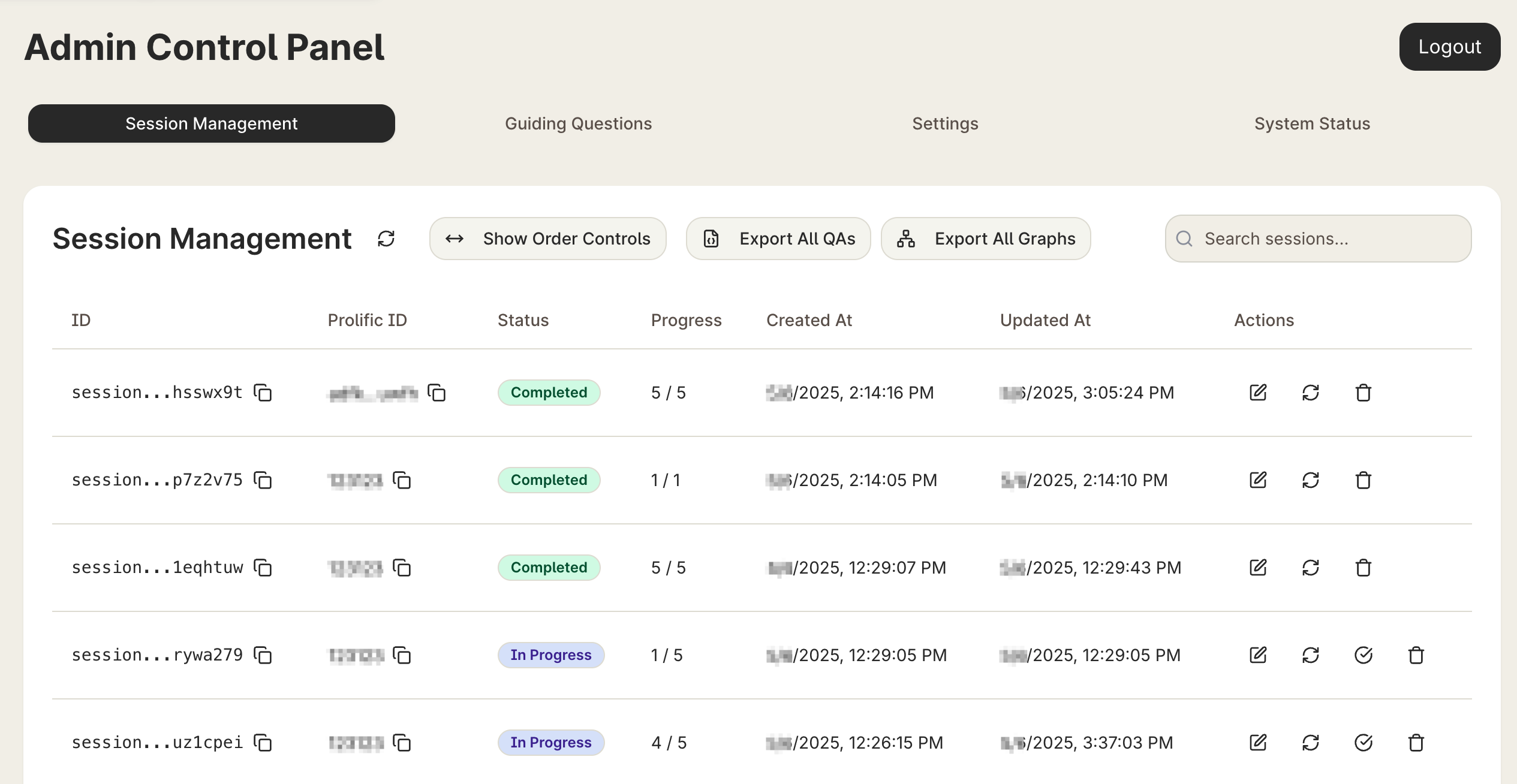}
    \caption{Admin session management panel with status tracking, progress monitoring, and export functionality. Researchers can monitor studies in real time and batch export reasoning data.}
    \label{fig:session-mgmt}
\end{figure}

\paragraph{Session management.}  
The {Session Management} dashboard (Fig.~\ref{fig:session-mgmt}) displays all ongoing and completed interviews with metadata including status, progress, and timestamps. From this panel, admins can (i) reorder questions, (ii) export raw QA data, or (iii) export causal graphs for downstream analysis. This unified view makes it easy to monitor study progress at scale and to recover high-fidelity reasoning traces.

\begin{figure}[ht]
    \centering
    \includegraphics[width=\linewidth]{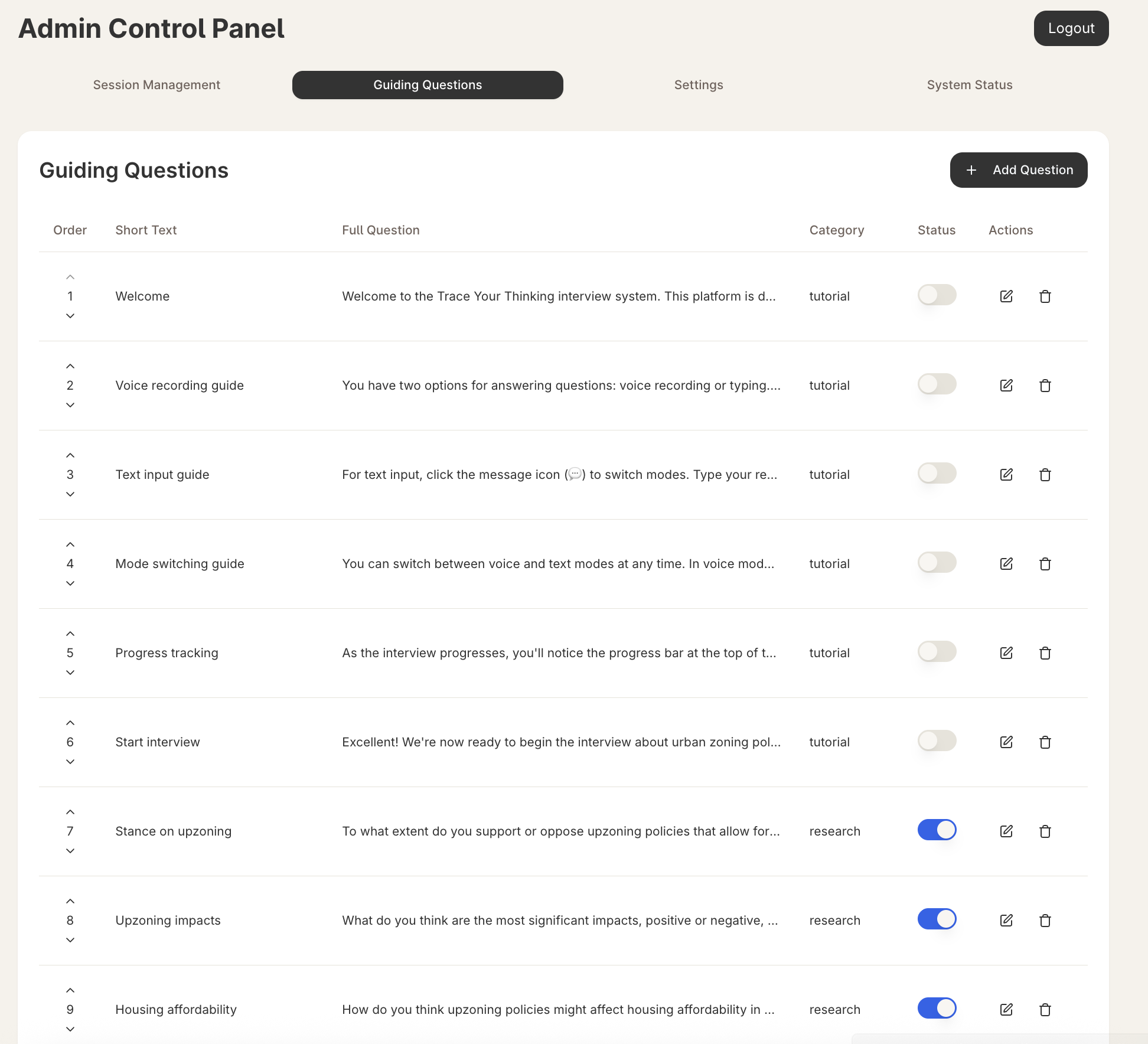}
    \caption{Guiding question editor. Researchers can toggle tutorial vs.\ research questions, reorder them dynamically, and experiment with alternative protocols. They can also choose to skip some questions by changing the status.}
    \label{fig:guiding-questions}
\end{figure}

\paragraph{Configurable guiding questions.}  
Admins can design and adjust the interview protocol using a guiding question editor (Fig.~\ref{fig:guiding-questions}). Each question has metadata (short text, full text, category), can be toggled on/off, and can be reordered dynamically. This flexibility makes it possible to test multiple hypotheses without rewriting the underlying system. In practice, this feature has been used to swap tutorial vs.\ research questions and to experiment with different probing strategies, making the platform versatile for diverse research programs.

\begin{figure}[ht]
    \centering
    \includegraphics[width=\linewidth]{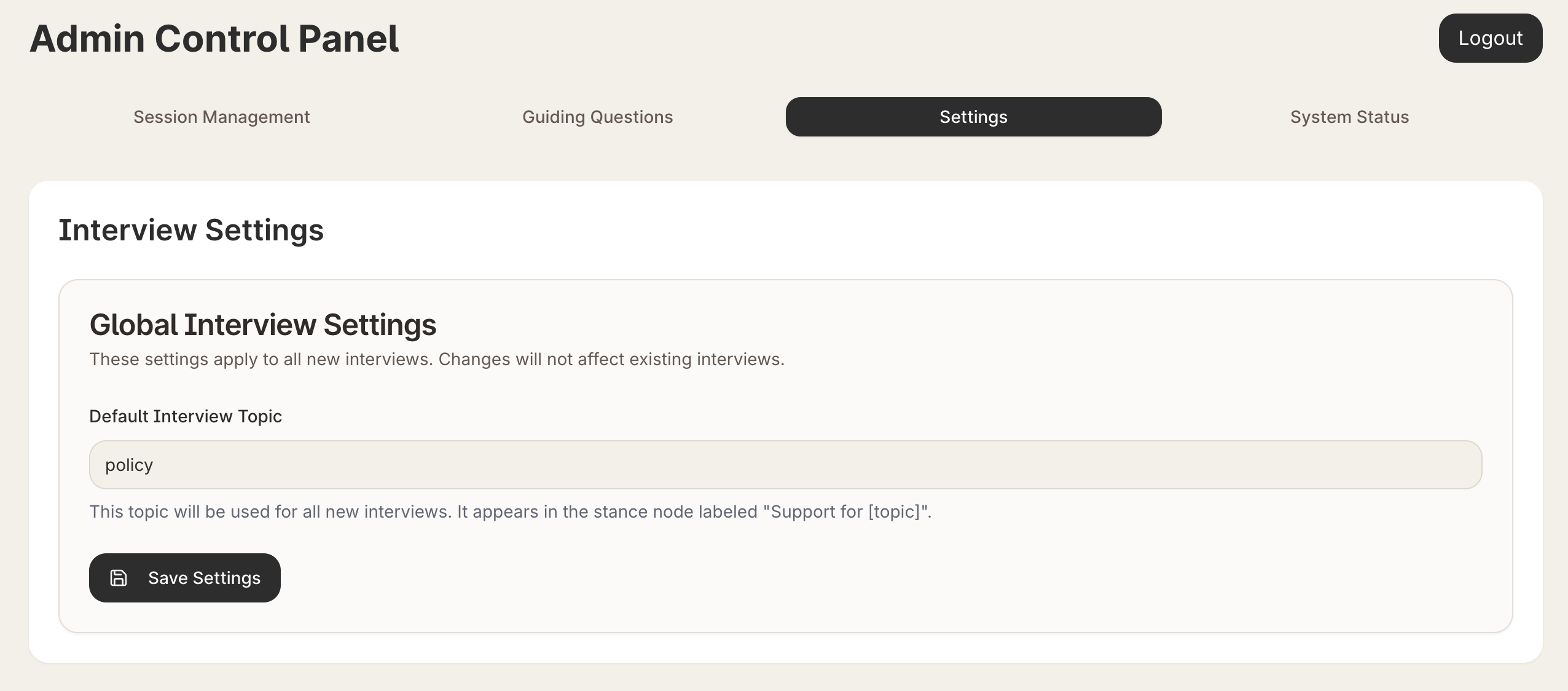}
    \caption{Global interview settings. With one change, the system can adapt to entirely new domains, enabling domain-agnostic deployment.}
    \label{fig:settings}
\end{figure}

\paragraph{Global settings.}  
Admins can set a global interview topic (e.g., policy, healthcare, surveillance) with a single configuration (Fig.~\ref{fig:settings}). This allows open-ended reasoning tasks to be deployed across arbitrary domains, ensuring that the platform is not tied to a fixed task. In effect, the system generalizes beyond a dataset-collection tool to become a reusable infrastructure for eliciting reasoning in any domain.

\subsection{Research Outputs (System Features)}

The system is designed to produce outputs that go beyond raw transcripts, giving researchers structured and analyzable data.

\begin{figure}[ht]
    \centering
    \includegraphics[width=\linewidth]{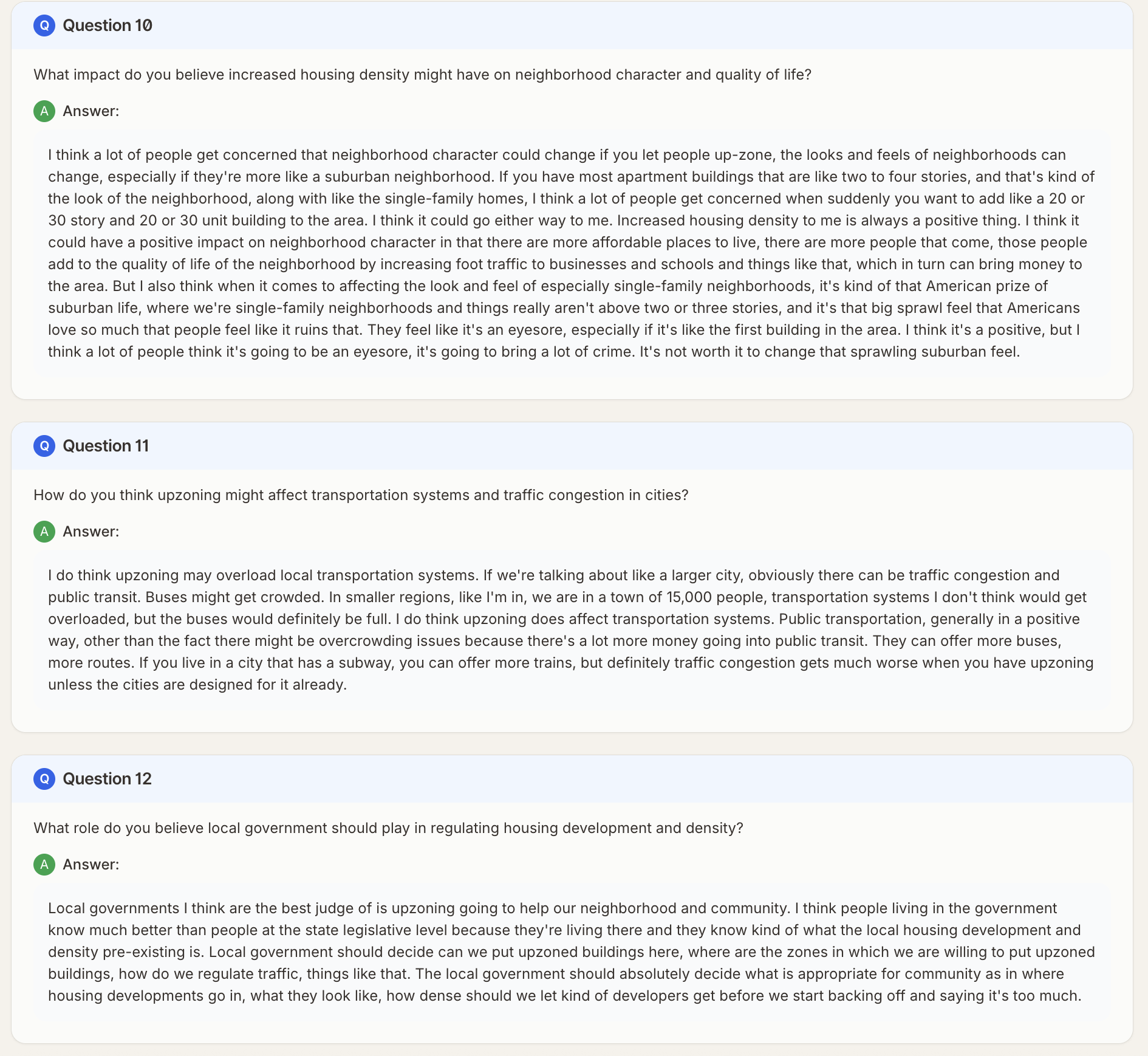}
    \caption{Sample QA transcripts highlighting variation in response depth and style. The system captures both structured argumentation and spontaneous informal commentary.}
    \label{fig:qa-sample}
\end{figure}

\paragraph{Raw QA transcripts.}  
All participant responses are preserved verbatim (Fig.~\ref{fig:qa-sample}). This ensures that qualitative nuances (hesitations, personal anecdotes, colloquial phrasing) are not lost. At the same time, transcripts provide the raw material for quantitative benchmarking, enabling evaluations of stance classification, belief calibration, and reasoning depth. The ability to capture both structured and noisy responses is a feature, not a limitation: it reflects the diversity of real-world human reasoning.

\begin{figure*}[ht]
    \centering
    \includegraphics[width=0.7\linewidth]{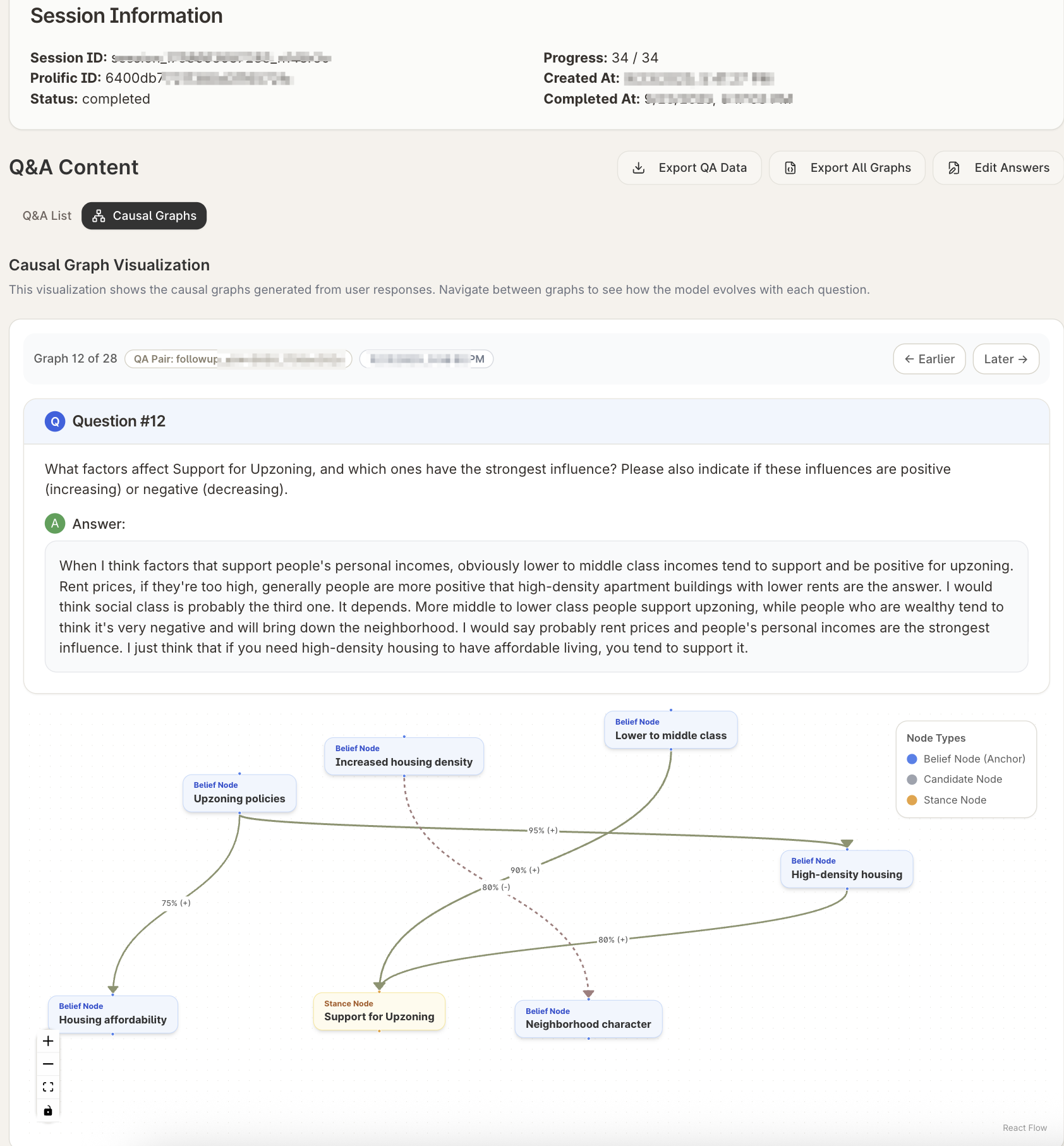}
    \caption{Dynamic causal graph visualization. Nodes capture beliefs, stances, and supporting reasons, updated continuously as the participant responds.}
    \label{fig:causal-graph}
\end{figure*}

\paragraph{Dynamic causal graphs.}  
The distinctive feature of \textsc{Trace-Your-Thinking} is the automatic construction of causal graphs in real time (Fig.~\ref{fig:causal-graph}). As participants answer questions, the system incrementally extracts stance nodes (opinions), belief nodes (anchors), and candidate nodes (supporting reasons). The graph expands as reasoning unfolds, producing a structured representation of how beliefs and justifications interconnect. This design is important for two reasons: (i) it transforms unstructured reasoning into analyzable graph data, and (ii) it enables researchers to trace belief updates step by step, rather than relying only on final outcomes. These graphs can be exported for downstream tasks such as reasoning alignment, structural consistency evaluation, or cross-domain transfer prediction.

\subsection{Use Cases}

The flexibility of the system enables multiple research paradigms:

\begin{itemize}
    \item \textbf{Baseline data collection:} Build large-scale corpora of reasoning traces in a controlled domain (e.g., housing policy), establishing benchmarks for human reasoning diversity. 
    \item \textbf{Cross-domain transfer:} Instantly switch topics (e.g., from zoning to healthcare) by editing global settings, to study how reasoning patterns generalize across domains. 
    \item \textbf{Longitudinal studies:} Re-engage the same participants over weeks or months via Prolific IDs, enabling the study of belief updates and reasoning drift. 
    \item \textbf{Human--model benchmarking:} Compare LLM predictions against human causal graphs to quantify intra-agent fidelity, context sensitivity, and adaptation gaps. 
\end{itemize}

\clearpage
\newpage

\section{Evaluation}

\subsection{Evaluation Metrics}
\label{sec:metrics}

Let $y_i$ denote the ground-truth response for instance $i$, $\hat{y}_i$ the model prediction, and $N$ the total number of instances. For belief dynamics update tasks, let $y_i^{\text{prev}}$ denote the participant's pre-intervention score.

\paragraph{Accuracy.}  
\[
\text{Acc} = \frac{1}{N} \sum_{i=1}^{N} \mathbf{1}\big[\, |\hat{y}_i - y_i| \leq \tau \,\big],
\]
where $\tau$ is the tolerance band ($\tau=1$ for 5-point scales, $\tau=2$ for 10-point scales).

\paragraph{Mean Absolute Error (MAE).}  
\[
\text{MAE} = \frac{1}{N} \sum_{i=1}^{N} |\hat{y}_i - y_i|.
\]

\paragraph{Directional Accuracy.}
We define directional accuracy as a two-stage weighted metric:
(1) \textit{Change Detection:} detecting whether a belief change occurred, and 
 if so, (2) \textit{Directional Inference:} correctly predicting the direction of change (\emph{increase}, \emph{decrease}, or \emph{no change}).
To better reflect the importance of directional reasoning, a higher weight is assigned to the second stage.

{\small
\begin{equation}
\begin{aligned}
\text{DirAcc}
={}&
\lambda \cdot \frac{1}{N}
\sum_{i=1}^{N}
\mathbf{1}\!\left[
\begin{array}{c}
\Delta y_i = 0 \wedge \Delta \hat{y}_i = 0
\\[-0.15em]
{}\vee{}
\\[-0.15em]
\Delta y_i \neq 0 \wedge \Delta \hat{y}_i \neq 0
\end{array}
\right]
\\
&+
(1-\lambda)\cdot
\frac{1}{|\mathcal{C}|}
\sum_{i\in\mathcal{C}}
\mathbf{1}\!\left[
\begin{array}{c}
\operatorname{sgn}(\Delta y_i)
\\[-0.1em]
=
\operatorname{sgn}(\Delta \hat{y}_i)
\end{array}
\right].
\end{aligned}
\label{eq:diracc}
\end{equation}
}

\noindent
where 
$\Delta \hat{y}_i = \hat{y}_i - \hat{y}_i^{\text{prev}}$ and 
$\Delta y_i = y_i - y_i^{\text{prev}}$ 
denote predicted and true belief changes, respectively.  
$\mathcal{C} = \{\, i \mid \Delta \hat{y}_i \neq 0 \wedge \Delta y_i \neq 0 \,\}$ 
is the set of samples where both predicted and true beliefs changed.  
We set $\lambda = 0.3$ by default, placing greater emphasis on directional correctness.

This weighting reflects the intuition that correctly inferring the direction of belief change is more informative than merely detecting whether a change occurred. 
While the first stage (\textit{change detection}) captures a coarse perceptual judgment that can often be guessed in noisy or stable settings, 
the second stage(\textit{directional inference})reveals whether the model truly understands and reasons about belief dynamics. 
Hence, emphasizing the latter better reflects a model’s fidelity to human-like reasoning processes and its capacity to simulate belief evolution.

\paragraph{Average-to-Individual (ATI) Score.}
To provide a single comprehensive measure of model performance across both static and dynamic belief tasks, we define a unified score, \textit{Average-to-Individual} (ATI) score, that integrates the \emph{Belief State Inference} (BSI) and \emph{Belief Dynamics Update} (BDU) components into a normalized value within $[0,1]$.
Specifically, $S_{\text{BSI}}$ denotes the normalized accuracy score for static belief state inference, 
while $S_{\text{BDU}}$ aggregates multiple metrics from the belief dynamics update task, 
including tolerance accuracy, normalized MAE, and directional reasoning accuracy. 
The unscaled ATI score is computed as:
\begin{equation}
\begin{aligned}
\text{ATI}_{\text{unscaled}}
={}&
\tfrac{1}{2} S_{\text{BSI}}
\\
&+
\tfrac{1}{2}
\Bigg[
\tfrac{1}{2}
\Big(
\tfrac{1}{2} S_{\text{BDU-mae-norm}}
+
\tfrac{1}{2} S_{\text{BDU-acc}}
\Big)
\\
&\qquad\qquad
+
\tfrac{1}{2} S_{\text{Directional-acc}}
\Bigg].
\end{aligned}
\label{eq:unifiedscore_modified}
\end{equation}
where each subscore $S_{\cdot} \in [0,1]$ represents a normalized evaluation metric.  
For MAE-based components, normalization is defined as:
\begin{equation}
S_{\text{BDU-mae-norm}} = 
\max\!\big(0,\; \min(1,\; 1 - \tfrac{\text{MAE}}{\text{MAE}_{\max}} )\big),
\label{eq:maenorm}
\end{equation}
where $\text{MAE}_{\max}$ denotes the task-specific upper bound of allowable error.
The unified score assigns equal weights to the state component ($S_{\text{BSI}}$) and the update component ($S_{\text{BDU}}$). 
Within the update branch, tolerance accuracy and MAE are equally weighted (0.5 each) to ensure a balanced consideration of robustness and precision. 
The directional reasoning component ($S_{\text{Directional-acc}}$) is assigned an equal weight (0.5) relative to the combined MAE and accuracy branch, 
reflecting its comparable importance in capturing belief-updating dynamics.

To facilitate interpretation relative to human performance and baseline behavior, we linearly rescale $\text{ATI}_{\text{unscaled}}$ to a 0--100 scale:
\begin{equation}
\text{ATI} = \frac{\text{ATI}_{\text{unscaled}} - \text{ATI}_{\text{random}}}{\text{ATI}_{\text{human}} - \text{ATI}_{\text{random}}} \times 100,
\label{eq:ati_scaled}
\end{equation}
where $\text{ATI}_{\text{random}}$ and $\text{ATI}_{\text{human}}$ denote the unscaled ATI scores of the random guess baseline and human upper bound, respectively. Under this rescaling, a score of 0 indicates random-level performance, while 100 represents human-level performance.

\subsection{Computation Details and Track Usage}
\label{app:computation-details}

Unless otherwise specified, all quantitative analyses and unified average to individual (ATI) score computations are performed on the \textbf{human track}, 
which serves as the primary evaluation benchmark due to its ecological validity and authentic reasoning diversity. 

The human track grounds evaluation in real participant reasoning. 

In practice, model predictions for both tasks---\emph{belief state inference} (BSI) and \emph{belief dynamics update} (BDU)---are first computed independently. 
All metrics (accuracy, MAE, and directional accuracy) are averaged across the three domains 
(\emph{healthcare}, \emph{surveillance}, \emph{zoning}) before aggregation into the unified score defined in Equation~\ref{eq:unifiedscore_modified}. 
Reported results in Section~\ref{sec:results} and subsequent findings are therefore based on the human track unless explicitly noted otherwise.

\newpage





\clearpage
\newpage

\section{Participant-Level Uncertainty}
\label{app:bootstrap}
Items are nested within participants, so the run-to-run standard deviations in Table~\ref{tab:all_results} do not cover the uncertainty that comes from the participant pool. We resample the 54 participants with replacement ($B = 1000$, fixed seed) and recompute the full pipeline in every replicate. The intervals in Table~\ref{tab:bootstrap} are the 2.5th and 97.5th percentiles of the resulting distribution. On belief dynamics update the gap to the human ceiling is separated: the largest model upper bound is 72.7, below the human lower bound of 80.9. On belief state inference the model upper bounds stay below the human point estimate and overlap the human interval, so we do not claim separation there. The intervals of closely ranked systems contain each other's point estimates, so we make no pairwise ranking claims. Substitute retrieval and full individual context are also not separated at this level; what supports that ordering is its stability across folds and metric variants (Tables~\ref{tab:lopo} and~\ref{tab:sensitivity}). The human row comes from the 14-day test--retest subset of 13 participants and measures consistency within individuals, so it is not the same kind of interval as the model rows.

Table~\ref{tab:bootstrap-domain} repeats the analysis within each domain. Healthcare is the hardest domain for belief updating at the point-estimate level for 13 of the 15 systems, and no system reaches 95\% separation from the better of the other two domains. We therefore report a consistent direction and do not rank the domains.

\begin{table*}[t]
\centering
\resizebox{\textwidth}{!}{%
\begin{tabular}{lccccc}
\toprule
System & BSI Acc & BDU Acc & MAE & Dir. Acc & ATI \\
\midrule
\textit{Human (test--retest, $n{=}13$)} & \textit{84.84 [80.00, 89.68]} & \textit{85.66 [80.91, 90.40]} & \textit{0.68 [0.55, 0.80]} & \textit{88.92 [82.09, 96.24]} & \textit{100 (by constr.)} \\
\midrule
LLaMA 3.3 70B & 76.4 [70.1, 82.0] & 67.6 [63.3, 71.5] & 1.24 [1.12, 1.37] & 79.6 [60.6, 88.9] & 69.8 [50.7, 80.5] \\
Claude Sonnet 4.5 & 76.0 [70.6, 81.0] & 68.6 [64.4, 72.7] & 1.18 [1.08, 1.29] & 78.7 [69.2, 87.2] & 67.4 [57.9, 76.2] \\
GPT-4o & 74.7 [69.0, 80.3] & 63.1 [58.1, 68.1] & 1.29 [1.17, 1.42] & 82.3 [74.7, 88.8] & 67.3 [57.1, 76.7] \\
DeepSeek-R1 & 75.4 [68.6, 81.1] & 64.9 [59.5, 70.0] & 1.29 [1.17, 1.41] & 79.7 [69.1, 87.4] & 67.2 [55.3, 77.4] \\
RAG-FC & 77.6 [71.2, 83.2] & 60.0 [54.6, 65.6] & 1.39 [1.25, 1.53] & 76.8 [58.1, 86.4] & 65.6 [46.8, 76.6] \\
Qwen-max & 77.4 [72.1, 82.5] & 58.9 [53.1, 64.8] & 1.40 [1.26, 1.55] & 77.2 [65.5, 86.7] & 65.2 [54.7, 75.7] \\
Qwen2.5-7B-instr. & 77.2 [72.0, 82.1] & 58.8 [53.0, 64.8] & 1.40 [1.26, 1.55] & 77.1 [65.5, 86.5] & 64.8 [53.6, 75.1] \\
Qwen2.5-32B-instr. & 77.2 [72.0, 82.2] & 59.0 [53.2, 64.9] & 1.40 [1.25, 1.55] & 76.9 [64.8, 86.4] & 64.7 [54.3, 75.0] \\
GPT5.1 Rea. High & 73.4 [68.7, 77.4] & 67.1 [63.1, 71.1] & 1.17 [1.05, 1.28] & 80.9 [73.4, 87.6] & 64.7 [56.3, 72.1] \\
Gemini 2.5 Pro & 75.4 [70.9, 79.9] & 64.9 [60.6, 69.0] & 1.27 [1.14, 1.39] & 78.7 [64.2, 87.4] & 64.3 [52.6, 74.1] \\
Generative Agents & 76.2 [70.5, 81.3] & 58.2 [52.5, 64.0] & 1.40 [1.26, 1.54] & 76.1 [57.7, 85.9] & 62.4 [45.3, 71.7] \\
GPT-5-mini & 75.3 [69.8, 80.0] & 58.2 [53.2, 62.9] & 1.43 [1.31, 1.57] & 77.0 [66.6, 85.3] & 61.5 [50.8, 70.0] \\
o3-mini & 75.1 [69.9, 80.2] & 64.5 [59.8, 69.4] & 1.22 [1.10, 1.35] & 71.3 [58.8, 81.2] & 60.9 [50.3, 69.8] \\
Gemini 2.0 Flash & 70.0 [64.1, 75.3] & 60.5 [54.6, 66.2] & 1.35 [1.21, 1.50] & 83.3 [75.5, 89.0] & 59.8 [50.2, 68.9] \\
RAG & 75.5 [70.2, 80.8] & 51.9 [45.9, 58.0] & 1.57 [1.42, 1.71] & 72.2 [59.4, 83.5] & 55.0 [43.9, 64.9] \\
\bottomrule
\end{tabular}}
\caption{Participant-level cluster bootstrap 95\% confidence intervals
($B{=}1000$).}
\label{tab:bootstrap}
\end{table*}

\begin{table*}[t]
\centering
\small
\begin{tabular}{lccc}
\toprule
System & Healthcare & Surveillance & Zoning \\
\midrule
LLaMA 3.3 70B & 64.3 [57.5, 70.2] & 65.3 [56.8, 73.7] & 73.1 [66.4, 79.1] \\
Claude Sonnet 4.5 & 65.2 [57.4, 72.1] & 72.7 [64.9, 79.8] & 67.9 [61.3, 73.8] \\
GPT-4o & 61.6 [53.3, 69.1] & 63.7 [55.0, 71.7] & 64.0 [56.0, 71.6] \\
DeepSeek-R1 & 61.4 [53.6, 68.9] & 68.3 [60.3, 75.7] & 64.9 [57.4, 71.7] \\
RAG-FC & 58.7 [50.3, 66.5] & 59.3 [50.4, 67.9] & 61.9 [53.8, 69.8] \\
Qwen-max & 56.8 [47.9, 64.7] & 57.7 [48.5, 66.6] & 62.1 [54.0, 70.0] \\
Qwen2.5-7B-instr. & 56.8 [47.8, 64.8] & 57.6 [48.5, 66.4] & 62.0 [54.0, 69.8] \\
Qwen2.5-32B-instr. & 56.8 [47.9, 64.7] & 57.8 [48.7, 66.8] & 62.3 [54.3, 70.2] \\
GPT5.1 Rea. High & 63.4 [56.6, 69.6] & 68.2 [61.6, 74.3] & 69.8 [63.7, 75.0] \\
Gemini 2.5 Pro & 62.1 [54.8, 68.6] & 62.5 [54.9, 69.2] & 70.0 [64.2, 75.7] \\
Generative Agents & 57.7 [49.1, 65.6] & 55.7 [46.2, 64.7] & 61.3 [53.0, 69.3] \\
GPT-5-mini & 59.9 [52.4, 67.0] & 55.2 [47.2, 62.7] & 59.6 [52.6, 66.4] \\
o3-mini & 60.1 [52.0, 67.3] & 62.5 [54.5, 69.9] & 71.0 [64.1, 77.3] \\
Gemini 2.0 Flash & 56.2 [46.8, 65.0] & 62.3 [53.4, 70.4] & 63.2 [55.2, 70.5] \\
RAG & 50.6 [41.5, 59.4] & 53.7 [45.6, 62.2] & 51.4 [43.5, 59.5] \\
\bottomrule
\end{tabular}
\caption{Belief-dynamics-update accuracy within each domain, with participant-level bootstrap 95\% confidence intervals (same procedure and seed as Table~\ref{tab:bootstrap}). Healthcare is the hardest domain at the point-estimate level for 13 of the 15 systems, and no system reaches 95\% separation from the better of the other two domains.}
\label{tab:bootstrap-domain}
\end{table*}

\section{Per-Participant Distributions}
\label{app:per-participant}

Table~\ref{tab:per-participant} reports the distribution over participants for every system. Predictions are pooled over the three domains and the five runs, which gives about 33 belief-state-inference and 128 belief-dynamics-update predictions per participant. The medians track the pooled accuracies, so the aggregates are not carried by a few well-predicted participants. Three to six participants out of 54 fall below the random reference for the stronger systems, and the count rises to 17 for the substitute-retrieval baseline. Belief state inference is pooled over domains because no single domain covers all 54 participants (43, 47 and 51 for healthcare, surveillance and zoning), which is also why its ranges are wide: a participant with few items can reach 0 or 100\%.

\begin{table*}[t]
\centering
\resizebox{\textwidth}{!}{%
\begin{tabular}{lcccccc}
\toprule
& \multicolumn{3}{c}{BDU Acc} & \multicolumn{3}{c}{BSI Acc} \\
\cmidrule(lr){2-4}\cmidrule(lr){5-7}
System & median [IQR] & range & $<$rand. & median [IQR] & range & $<$rand. \\
\midrule
LLaMA 3.3 70B & 68.9 [54.4, 76.0] & [28.0, 92.5] & 5/54 & 83.3 [66.7, 100.0] & [0.0, 100.0] & 5/54 \\
GPT5.1 Rea. High & 65.3 [55.7, 74.6] & [26.7, 88.8] & 3/54 & 77.2 [62.5, 84.2] & [0.0, 100.0] & 5/54 \\
Claude Sonnet 4.5 & 65.0 [56.5, 78.9] & [33.3, 90.5] & 3/54 & 80.9 [66.7, 100.0] & [0.0, 100.0] & 5/54 \\
DeepSeek-R1 & 64.6 [52.9, 74.8] & [21.1, 91.7] & 6/54 & 78.6 [66.7, 97.3] & [0.0, 100.0] & 5/54 \\
o3-mini & 64.4 [52.3, 75.8] & [32.5, 94.4] & 6/54 & 80.0 [66.2, 91.1] & [0.0, 100.0] & 5/54 \\
GPT-4o & 62.0 [48.5, 74.9] & [25.6, 91.9] & 5/54 & 80.0 [66.7, 92.1] & [0.0, 100.0] & 7/54 \\
Gemini 2.5 Pro & 61.0 [54.1, 70.7] & [32.0, 100.0] & 4/54 & 75.0 [63.5, 89.7] & [0.0, 100.0] & 4/54 \\
Qwen2.5-7B-instr. & 60.1 [43.1, 71.7] & [24.4, 91.8] & 14/54 & 80.9 [67.5, 100.0] & [0.0, 100.0] & 5/54 \\
Generative Agents & 59.9 [42.7, 72.7] & [27.7, 94.4] & 15/54 & 80.9 [63.5, 93.3] & [0.0, 100.0] & 5/54 \\
Qwen2.5-32B-instr. & 59.8 [43.6, 71.3] & [23.3, 91.8] & 13/54 & 80.9 [66.7, 100.0] & [0.0, 100.0] & 5/54 \\
Gemini 2.0 Flash & 59.8 [46.1, 75.5] & [25.0, 91.8] & 12/54 & 71.4 [60.0, 88.1] & [0.0, 100.0] & 8/54 \\
Qwen-max & 59.5 [43.5, 70.9] & [23.3, 91.8] & 13/54 & 82.6 [66.7, 100.0] & [0.0, 100.0] & 5/54 \\
RAG-FC & 58.9 [44.2, 74.6] & [25.6, 94.4] & 12/54 & 83.1 [63.5, 100.0] & [0.0, 100.0] & 6/54 \\
GPT-5-mini & 57.6 [48.4, 69.0] & [25.6, 88.6] & 10/54 & 80.0 [62.6, 93.6] & [0.0, 100.0] & 5/54 \\
RAG & 50.9 [38.7, 66.4] & [16.7, 90.0] & 17/54 & 80.9 [60.8, 97.1] & [0.0, 100.0] & 5/54 \\
\bottomrule
\end{tabular}}
\caption{Per-participant performance distributions, with predictions pooled
across the three domains and the five runs. The column `$<$rand.' counts
participants below the random reference, which is 50\% for belief state
inference and 43.12\% for belief dynamics update.}
\label{tab:per-participant}
\end{table*}

\section{Leaving Out Participants and Domains}
\label{app:robustness-folds}

We rerun the full pipeline on 54 leave-one-participant-out folds and on the three leave-one-domain-out folds. No single participant drives the results. The top-ranked system is the same in all 54 folds, Kendall's $\tau$ against the full-sample ranking is at least 0.77 (mean 0.94), and the index of the leading system stays within $[66.5, 72.0]$. Systems in the middle of the table move by a few positions, which is what their overlapping bootstrap intervals predict. Domains matter more, which is expected of three topics chosen to differ. Holding out zoning moves the runner-up to first place, and the bootstrap intervals of the two systems overlap. The gap to the human ceiling, the ordering of the two tasks, and the ranking of full individual context above the retrieval baselines hold in every fold.

\begin{table*}[t]
\centering
\small
\begin{tabular}{lccccc}
\toprule
System & ATI (full) & ATI over folds & BSI over folds & BDU Acc over folds & max rank shift \\
\midrule
LLaMA 3.3 70B & 69.84 & [66.52, 71.97] & [75.62, 77.60] & [66.63, 68.27] & 0 \\
Claude Sonnet 4.5 & 67.39 & [65.88, 69.28] & [75.49, 77.24] & [67.80, 69.23] & 2 \\
GPT-4o & 67.29 & [65.87, 69.23] & [73.99, 75.86] & [62.06, 63.78] & 3 \\
DeepSeek-R1 & 67.20 & [65.14, 69.23] & [74.58, 76.65] & [63.93, 65.68] & 2 \\
RAG-FC & 65.65 & [62.29, 67.67] & [76.80, 78.57] & [58.83, 60.74] & 4 \\
Qwen-max & 65.21 & [63.48, 66.65] & [76.79, 78.40] & [57.66, 59.88] & 2 \\
Qwen2.5-7B-instr. & 64.83 & [62.82, 66.45] & [76.56, 78.17] & [57.62, 59.83] & 2 \\
Qwen2.5-32B-instr. & 64.71 & [62.71, 66.32] & [76.56, 78.15] & [57.76, 59.95] & 3 \\
GPT5.1 Rea. High & 64.66 & [63.38, 66.02] & [72.77, 74.25] & [66.35, 67.77] & 4 \\
Gemini 2.5 Pro & 64.29 & [62.05, 67.33] & [74.85, 76.34] & [64.00, 65.39] & 7 \\
Generative Agents & 62.43 & [58.39, 64.33] & [75.50, 77.19] & [57.14, 59.15] & 3 \\
GPT-5-mini & 61.53 & [59.76, 63.84] & [74.69, 76.32] & [57.16, 59.14] & 1 \\
o3-mini & 60.92 & [59.07, 63.34] & [74.32, 76.19] & [63.71, 65.40] & 1 \\
Gemini 2.0 Flash & 59.76 & [58.42, 61.70] & [69.00, 71.33] & [59.40, 61.47] & 2 \\
RAG & 54.96 & [52.70, 57.52] & [74.70, 76.39] & [50.75, 53.00] & 0 \\
\bottomrule
\end{tabular}
\caption{Leave-one-participant-out robustness. The columns give the range of
each metric over the 54 jackknife folds. The top-ranked system is unchanged in
54 of 54 folds and Kendall's $\tau$ against the full-sample ranking is at least
0.77 (mean 0.94). Systems with adjacent ranks do swap, which is consistent with
their overlapping intervals in Table~\ref{tab:bootstrap}.}
\label{tab:lopo}
\end{table*}

\begin{table}[t]
\centering
\small
\setlength{\tabcolsep}{4pt}
\begin{tabular}{lcccc}
\toprule
System & Full & $-$health. & $-$surv. & $-$zoning \\
\midrule
LLaMA 3.3 70B & 69.8 & 76.0 (1) & 68.5 (1) & 65.0 (3) \\
Claude Sonnet 4.5 & 67.4 & 71.4 (3) & 62.9 (9) & 67.8 (1) \\
GPT-4o & 67.3 & 72.0 (2) & 66.4 (4) & 63.5 (6) \\
DeepSeek-R1 & 67.2 & 69.3 (5) & 67.0 (2) & 65.4 (2) \\
RAG-FC & 65.6 & 68.9 (6) & 64.4 (7) & 63.6 (5) \\
Qwen-max & 65.2 & 67.4 (10) & 66.6 (3) & 61.7 (7) \\
Qwen2.5-7B-instr. & 64.8 & 67.4 (9) & 66.2 (5) & 60.9 (10) \\
Qwen2.5-32B-instr. & 64.7 & 67.6 (8) & 65.4 (6) & 61.2 (9) \\
GPT5.1 Rea. High & 64.7 & 68.8 (7) & 61.0 (12) & 64.2 (4) \\
Gemini 2.5 Pro & 64.3 & 70.0 (4) & 63.7 (8) & 59.1 (12) \\
Generative Agents & 62.4 & 65.8 (12) & 62.0 (11) & 59.4 (11) \\
GPT-5-mini & 61.5 & 64.4 (14) & 59.0 (14) & 61.2 (8) \\
o3-mini & 60.9 & 67.3 (11) & 59.5 (13) & 56.0 (13) \\
Gemini 2.0 Flash & 59.8 & 65.8 (13) & 62.0 (10) & 51.4 (15) \\
RAG & 55.0 & 62.8 (15) & 50.4 (15) & 51.7 (14) \\
\bottomrule
\end{tabular}
\caption{Leave-one-domain-out robustness: adaptation index and rank when one
domain is held out and the metrics are macro-averaged over the two that remain.
Kendall's $\tau$ against the full-sample ranking is 0.70, 0.64 and 0.70 for the
three folds. The top-ranked system is unchanged in two of the three; holding out
zoning moves Claude Sonnet 4.5 to first place, and the bootstrap intervals of
the two systems overlap.}
\label{tab:lodo}
\end{table}

\section{Sensitivity to Metric Design Choices}
\label{app:sensitivity}

We re-derive the ranking under 11 alternatives to the metric design, changing one choice at a time: the weight on the change-detection stage of directional accuracy, the MAE normalization bound, four component weight schemes, two tighter tolerance bands, and per-participant aggregation of reason weights in place of item-pooled aggregation. The final rescaling of the index is monotone, so rankings are computed on the unscaled index. The top-ranked system is unchanged in all 11 variants, and the same four systems fill the top four ranks in 7 of them (Table~\ref{tab:sensitivity}). The weakest agreement, $\tau = 0.56$, comes from weighting all four components equally, which reduces the influence of belief state inference. Tolerance bands move absolute accuracy a great deal but not the top of the leaderboard (Table~\ref{tab:bands}). The gap to the human ceiling does not depend on the band, because it also appears in MAE, which no band affects (human 0.68 against 1.17 to 1.57 across systems). Averaging within each participant before averaging across participants lowers belief-dynamics-update accuracy by about four points and leaves the ranking close to unchanged, $\tau = 0.94$ (Table~\ref{tab:reason-split}). Every system is more accurate on stances than on the reason weights behind them, which is the asymmetry the main text builds on.

\begin{table*}[t]
\centering
\small
\begin{tabular}{lccl}
\toprule
Variant & Kendall $\tau$ & Top-1 changed & Top-3 \\
\midrule
base (paper) & 1.000 & no & LLaMA, Claude, GPT-4o \\
$\lambda = 0.5$ & 0.886 & no & LLaMA, Claude, DeepSeek-R1 \\
$\lambda = 0.7$ & 0.790 & no & LLaMA, Claude, DeepSeek-R1 \\
MAE bound unified at 5 & 0.848 & no & LLaMA, Claude, GPT-4o \\
MAE bound unified at 4 & 0.848 & no & LLaMA, Claude, GPT-4o \\
weights flat (0.25 each) & 0.562 & no & LLaMA, DeepSeek-R1, GPT-4o \\
weights BDU-internal equal & 0.943 & no & LLaMA, Claude, DeepSeek-R1 \\
weights BSI:BDU $=$ 0.4:0.6 & 0.771 & no & LLaMA, GPT-4o, DeepSeek-R1 \\
weights BSI:BDU $=$ 0.6:0.4 & 0.924 & no & LLaMA, Claude, RAG-FC \\
band tight ($\pm1$ / exact) & 0.600 & no & LLaMA, Gemini 2.5 Pro, GPT-4o \\
band exact match & 0.638 & no & LLaMA, Gemini 2.5 Pro, GPT-4o \\
reason weights: participant mean & 0.943 & no & LLaMA, DeepSeek-R1, GPT-4o \\
\bottomrule
\end{tabular}
\caption{Sensitivity of the ranking to metric design choices, one choice changed
at a time. Stability is measured by Kendall's $\tau$ of the adaptation-index
ordering of all 15 systems against the configuration used in the paper. LLaMA
abbreviates LLaMA 3.3 70B and Claude abbreviates Claude Sonnet 4.5. The
substitute-retrieval baseline ranks last in every row.}
\label{tab:sensitivity}
\end{table*}

\begin{table}[t]
\centering
\small
\begin{tabular}{lrrr}
\toprule
System & $\pm2$ / $\pm1$ & $\pm1$ / exact & exact \\
\midrule
Claude Sonnet 4.5 & 68.61 & 34.99 & 27.36 \\
LLaMA 3.3 70B & 67.57 & 39.81 & 34.05 \\
GPT5.1 Rea. High & 67.13 & 40.93 & 33.05 \\
DeepSeek-R1 & 64.88 & 28.59 & 20.84 \\
Gemini 2.5 Pro & 64.87 & 43.59 & 35.65 \\
o3-mini & 64.54 & 37.85 & 31.75 \\
GPT-4o & 63.11 & 33.05 & 24.71 \\
Gemini 2.0 Flash & 60.55 & 34.22 & 25.36 \\
RAG-FC & 59.97 & 31.69 & 22.68 \\
Qwen2.5-32B-instr. & 58.96 & 32.00 & 23.00 \\
Qwen-max & 58.86 & 31.91 & 22.71 \\
Qwen2.5-7B-instr. & 58.82 & 32.08 & 22.82 \\
Generative Agents & 58.22 & 32.03 & 22.57 \\
GPT-5-mini & 58.21 & 34.90 & 26.12 \\
RAG & 51.90 & 23.59 & 15.48 \\
\bottomrule
\end{tabular}
\caption{Belief-dynamics-update accuracy under the three tolerance bands. The
first column is the band used in the paper. Tightening the band moves absolute
accuracy a great deal and reshuffles the middle of the table, while the top does
not move.}
\label{tab:bands}
\end{table}

\begin{table*}[t]
\centering
\small
\begin{tabular}{lrrrrrr}
\toprule
& \multicolumn{2}{c}{Accuracy} & \multicolumn{2}{c}{MAE} & \multicolumn{2}{c}{BDU Acc aggregation} \\
\cmidrule(lr){2-3}\cmidrule(lr){4-5}\cmidrule(lr){6-7}
System & stance & reason & stance & reason & item-pooled & participant-mean \\
\midrule
Claude Sonnet 4.5 & 75.09 & 67.40 & 0.95 & 1.24 & 68.61 & 64.72 \\
LLaMA 3.3 70B & 75.91 & 64.62 & 1.08 & 1.29 & 67.57 & 63.36 \\
GPT5.1 Rea. High & 72.47 & 64.63 & 0.97 & 1.24 & 67.13 & 62.97 \\
DeepSeek-R1 & 73.34 & 62.88 & 1.06 & 1.34 & 64.88 & 63.24 \\
Gemini 2.5 Pro & 73.72 & 61.38 & 0.99 & 1.35 & 64.87 & 60.74 \\
o3-mini & 66.85 & 63.65 & 1.12 & 1.26 & 64.54 & 61.88 \\
GPT-4o & 70.47 & 62.22 & 1.05 & 1.33 & 63.11 & 60.97 \\
Gemini 2.0 Flash & 71.55 & 56.74 & 1.01 & 1.46 & 60.55 & 57.56 \\
RAG-FC & 71.90 & 56.55 & 1.13 & 1.46 & 59.97 & 57.94 \\
Qwen2.5-32B-instr. & 72.34 & 54.77 & 1.04 & 1.51 & 58.96 & 57.04 \\
Qwen-max & 72.32 & 54.70 & 1.04 & 1.51 & 58.86 & 56.92 \\
Qwen2.5-7B-instr. & 71.96 & 54.78 & 1.04 & 1.51 & 58.82 & 56.94 \\
Generative Agents & 72.94 & 53.29 & 1.06 & 1.52 & 58.22 & 56.66 \\
GPT-5-mini & 69.46 & 54.14 & 1.13 & 1.52 & 58.21 & 56.04 \\
RAG & 63.38 & 47.37 & 1.20 & 1.69 & 51.90 & 51.08 \\
\bottomrule
\end{tabular}
\caption{Belief dynamics update split into stance items (1 to 10 scale) and
reason-weight items (1 to 5 scale), and reported under both reason-weight
aggregations.}
\label{tab:reason-split}
\end{table*}

\section{A Stronger Retriever for the Retrieval Baselines}
\label{app:dense-rag}

TF-IDF is a weak representative of retrieval-augmented and memory-augmented architectures, so we re-ran both variants with a dense retriever (\texttt{bge-small-en-v1.5}, cosine similarity over embedded question--answer pairs) and kept the protocol of the paper: the same base model and prompts, $k = 5$, the per-participant retrieval pool, temperature 0.1, and three runs over all 1{,}742 items per setting with no parsing failures. The conclusion does not change (Table~\ref{tab:dense-rag}). As a substitute for individual context, the dense retriever reaches an adaptation index of 55.07 against 54.96 for TF-IDF, so the gap of about ten points to full individual context remains. As an auxiliary signal it stays below full context in all three runs. At the metric level the dense retriever recovers numeric closeness on belief dynamics update, with tolerance accuracy rising from 51.90 to 59.24 and MAE falling from 1.57 to 1.41. This is consistent with our finding that belief updating turns on a few high-signal question--answer pairs, which a semantic retriever is better at locating. Belief state inference and directional accuracy do not recover (73.01 against 77.17, and 72.05 against 76.88). Stronger retrieval therefore improves local answer matching and does not substitute for full individual context.

\begin{table*}[t]
\centering
\small
\begin{tabular}{lrrrrr}
\toprule
System & BSI Acc & BDU Acc & MAE & Dir. Acc & ATI \\
\midrule
Substitute retrieval, TF-IDF (paper) & 75.46$^{\pm0.42}$ & 51.90$^{\pm0.35}$ & 1.57$^{\pm0.00}$ & 72.25$^{\pm0.57}$ & 54.96$^{\pm0.87}$ \\
Substitute retrieval, dense & 73.01$^{\pm0.60}$ & 59.24$^{\pm0.15}$ & 1.41$^{\pm0.00}$ & 72.05$^{\pm1.90}$ & 55.07$^{\pm1.75}$ \\
Auxiliary retrieval, TF-IDF (paper) & 77.56$^{\pm0.34}$ & 59.97$^{\pm0.21}$ & 1.39$^{\pm0.00}$ & 76.80$^{\pm0.80}$ & 65.65$^{\pm0.61}$ \\
Auxiliary retrieval, dense & 75.92$^{\pm0.71}$ & 62.30$^{\pm0.18}$ & 1.36$^{\pm0.00}$ & 74.01$^{\pm2.43}$ & 62.28$^{\pm1.56}$ \\
\midrule
Full individual context (paper) & 77.17$^{\pm0.28}$ & 58.96$^{\pm0.05}$ & 1.40$^{\pm0.00}$ & 76.88$^{\pm0.29}$ & 64.71$^{\pm0.23}$ \\
\textit{Human (quoted)} & \textit{84.84} & \textit{85.66} & \textit{0.68} & \textit{88.92} & \textit{100.0} \\
\bottomrule
\end{tabular}
\caption{Replacing the TF-IDF retriever of the two retrieval baselines with a
dense retriever. Substitute retrieval replaces the individual context with the
retrieved pairs; auxiliary retrieval prepends them to the full context. All
system rows use \texttt{Qwen2.5-32B-instruct} as the base model. Dense rows are
means over three runs and the paper rows are means over five runs. Directional
accuracy varies more across runs under dense retrieval, so differences of a few
points in that column should not be over-read.}
\label{tab:dense-rag}
\end{table*}

\section{Use of LLMs}
\label{app:llm-usage}
Large Language Models (LLMs) were used to aid in the writing and polishing of the manuscript. 

It is important to note that the LLM was not involved in the ideation, research methodology, or experimental design. All research concepts, ideas, and analyses were developed and conducted by the authors. The contributions of the LLM were solely focused on improving the linguistic quality of the paper, with no involvement in the scientific content or data analysis.

The authors take full responsibility for the content of the manuscript, including any text generated or polished by the LLM. We have ensured that the LLM-generated text adheres to ethical guidelines and does not contribute to plagiarism or scientific misconduct.

\end{document}